\documentclass[11pt]{article} 

\usepackage{textcomp, amssymb}  
\usepackage{anysize}            
\usepackage{amsfonts}
\usepackage{amsmath}
\usepackage{dsfont}
\usepackage{MnSymbol}
\usepackage{mathtools}
\usepackage{graphicx}
\usepackage{epsfig}
\usepackage{epstopdf}
\usepackage{tikz}
\usepackage{amsthm}
\usepackage{ifpdf}
\usepackage{thm-restate}
\usepackage{stmaryrd}
\usepackage{cleveref}
\usepackage{subcaption}
\usepackage{longtable}
\usepackage{adjustbox}

\usetikzlibrary{calc,decorations.pathmorphing,shapes,graphs,positioning,automata,angles,cd}

\newcounter{sarrow}

\newcounter{sarrow1}
\newcommand\xnrsquigarrow[1]{%
\stepcounter{sarrow1}%
\mathrel{\begin{tikzpicture}[baseline= {( $ (current bounding box.south) + (0,-0.5ex) $ )}]
\node[inner sep=.5ex] (\thesarrow) {$\scriptstyle #1$};
\path[draw,<-,decorate,
  decoration={zigzag,amplitude=0.7pt,segment length=1.2mm,pre=lineto,pre length=4pt}]
    (\thesarrow1.south east) -- (\thesarrow1.south west);
    $\slashedarrowfill@\relbar\relbar/$
    \end{tikzpicture}}%
}

\makeatletter
\def\slashedarrowfill@#1#2#3#4#5{%
  $\m@th\thickmuskip0mu\medmuskip\thickmuskip\thinmuskip\thickmuskip
   \relax#5#1\mkern-7mu%
   \cleaders\hbox{$#5\mkern-2mu#2\mkern-2mu$}\hfill
   \mathclap{#3}\mathclap{#2}%
   \cleaders\hbox{$#5\mkern-2mu#2\mkern-2mu$}\hfill
   \mkern-7mu#4$%
}
\def\rightslashedarrowfillb@{%
  \slashedarrowfill@\relbar\relbar/\rightarrow}
\newcommand\xnrightarrow[2][]{%
  \ext@arrow 0055{\rightslashedarrowfillb@}{#1}{#2}}

\def\rightslashedarrowfille@{%
  \slashedarrowfill@\relbar\relbar/\twoheadrightarrow}
\newcommand\xntworightarrow[2][]{%
  \ext@arrow 0055{\rightslashedarrowfille@}{#1}{#2}}

\def\rightslashedarrowfillg@{%
  \slashedarrowfill@\relbar\relbar{\raisebox{.12em}{}}\twoheadrightarrow}
\newcommand\xtworightarrow[2][]{%
  \ext@arrow 0055{\rightslashedarrowfillg@}{#1}{#2}}

\def\rightslashedarrowfillx@{%
  \slashedarrowfill@\Relbar\Relbar/\rightrightarrows}
\newcommand\xnTworightarrow[2][]{%
  \ext@arrow 0055{\rightslashedarrowfillx@}{#1}{#2}}

\def\rightslashedarrowfilly@{%
  \slashedarrowfill@\Relbar\Relbar{\raisebox{.12em}{}}\rightrightarrows}
\newcommand\xTworightarrow[2][]{%
  \ext@arrow 0055{\rightslashedarrowfilly@}{#1}{#2}}

\pgfdeclareshape{slash underlined}
{
  \inheritsavedanchors[from=rectangle] 
  \inheritanchorborder[from=rectangle]
  \inheritanchor[from=rectangle]{north}
  \inheritanchor[from=rectangle]{north west}
  \inheritanchor[from=rectangle]{north east}
  \inheritanchor[from=rectangle]{center}
  \inheritanchor[from=rectangle]{west}
  \inheritanchor[from=rectangle]{east}
  \inheritanchor[from=rectangle]{mid}
  \inheritanchor[from=rectangle]{mid west}
  \inheritanchor[from=rectangle]{mid east}
  \inheritanchor[from=rectangle]{base}
  \inheritanchor[from=rectangle]{base west}
  \inheritanchor[from=rectangle]{base east}
  \inheritanchor[from=rectangle]{south}
  \inheritanchor[from=rectangle]{south west}
  \inheritanchor[from=rectangle]{south east}
  \inheritanchorborder[from=rectangle]
  \foregroundpath{
    \southwest \pgf@xa=\pgf@x \pgf@ya=\pgf@y
    \northeast \pgf@xb=\pgf@x \pgf@yb=\pgf@y
    \pgf@xc=\pgf@xa
    \advance\pgf@xc by .5\pgf@xb
    \pgf@yc=\pgf@ya 
    \advance\pgf@xc by -1.3pt
    \advance\pgf@yc by -1.8pt
    \pgfpathmoveto{\pgfqpoint{\pgf@xc}{\pgf@yc}}
    \advance\pgf@xc by  2.6pt
    \advance\pgf@yc by  3.6pt
    \pgfpathlineto{\pgfqpoint{\pgf@xc}{\pgf@yc}}
    \pgfpathmoveto{\pgfqpoint{\pgf@xa}{\pgf@ya}}
    \pgfpathlineto{\pgfqpoint{\pgf@xb}{\pgf@ya}}
 }
}
\tikzset{nomorepostaction/.code=\let\tikz@postactions\pgfutil@empty}

\newlength{\doublefracgap}
\DeclareRobustCommand{\doublefrac}[2]{%
  \mathinner{\mathpalette\doublefrac@{{#1}{#2}}}%
}
\newcommand{\doublefrac@}[2]{\doublefrac@@#1#2}
\newcommand{\doublefrac@@}[3]{%
  \ooalign{%
    \raisebox{\doublefracgap}{$\m@th#1\frac{#2}{\phantom{#3}}$}\cr
    \raisebox{-\doublefracgap}{$\m@th#1\frac{\phantom{#2}}{#3}$}\cr
  }%
}

\newcommand*{\rmbrace}{|\mskip-4mu\}}
\newcommand*{\lmbrace}{\{\mskip-4mu|}
\newcommand*{\mset}[1]{\lmbrace#1\rmbrace}
\newcommand*{\rsbrace}{|\mskip-4mu\rangle}
\newcommand*{\lsbrace}{\langle\mskip-4mu|}
\newcommand*{\step}[1]{\lsbrace#1\rsbrace}
\newtheorem{theorem}{Theorem}[section]
\newtheorem{definition}[theorem]{Definition}
\newtheorem{proposition}[theorem]{Proposition}
\newtheorem{lemma}[theorem]{Lemma}

\newcommand{\runhead}[1]{\begin{center}\par\medskip\noindent\textbf{#1}\end{center}}

\begin{document}

\begin{titlepage}
\thispagestyle{empty}

\hrule
\begin{center}
{\bf\LARGE Categorical Foundations of AI Agents\\}

\vspace{0.5cm}
--- Yong Wang ---

\vspace{2cm}

\end{center}
\end{titlepage}

\newpage 

\setcounter{page}{1}\pagenumbering{roman}

\tableofcontents

\newpage
\setcounter{page}{1}\pagenumbering{arabic}

        \section{Introduction}\label{intro}

AI agents bing task and software automation into reality. The Model Context Protocol (MCP) \cite{MCP} like protocols provide interfaces to discover and invoke external tools. There are many tool-augmented AI agents paradigms, such as ReAct \cite{ReAct}, Toolformer \cite{Toolformer}, ToolLLM \cite{Toolllm}, Gorilla \cite{Gorilla}, HuggingGPT \cite{HuggingGPT}, ART \cite{Art}, et al. In the multi AI agent framework, such as LangChain \cite{LangChain}, AutoGen \cite{AutoGen}, MetaGPT \cite{MetaGPT}, ChatDev \cite{ChatDev}, TaskWeaver \cite{TaskWeaver}, the reasoning with LLM and the execution are mixed together. Recently, \cite{Separating} introduces a framework \cite{Separating} to separate the reasoning and the execution: during the intelligence phase, the user consults with the LLM to produce the orchestration and choreography of AI agent workflow; while during the execution phase, the AI agent workflow is executed.

We follow this idea of separating intelligence from execution and give the categorical foundations of AI agents: the transformers, the agents, and the orchestrations and choreographies of AI agents are categorized, and which also include their coalgebras and algebras.

\subsection{Pomset}

\begin{definition}[Labelled poset]
Let $\Sigma$ be a fixed set of labels. A labelled poset is a tuple $\mathbf{u}=\langle S, \le, \ell\rangle$, where $S$ is the carrier set, $\le$ is a partial order on $S$ and $\ell$ is a labelling function $\ell:S\to\Sigma$.

For a labelled poset $\mathbf{u}$, $S_{\mathbf{u}}$, $\le_{\mathbf{u}}$ and $\ell_{\mathbf{u}}$ denote the carrier, the partial order and the labelling of $\mathbf{u}$ respectively. The set of labelled posets is denoted $\mathsf{LP}$ and the empty labelled poset is $1$.
\end{definition}

\begin{definition}[Poset morphism]
For posets $\langle S,\le\rangle$ and $\langle S',\le'\rangle$ and function $\varphi:S\to S'$, $f$ is called a poset morphism if for $s_0,s_1\in S$ with $s_0\le s_1$, then $\varphi(s_0)\le' \varphi(s_1)$ holds.
\end{definition}

\begin{definition}[Labelled poset isomorphism]
Let $\mathbf{u}=\langle S_1,\le_1,\ell_1\rangle$ and $\mathbf{v}=\langle S_2,\le_2,\ell_2\rangle$ be labelled posets. A labelled poset morphism $\varphi$ from $\mathbf{u}=\langle S_1,\le_1,\ell_1\rangle$ to $\mathbf{v}=\langle S_2,\le_2,\ell_2\rangle$ is a poset morphism from $\langle S_1,\le_1\rangle$ and $\langle S_2,\le_2\rangle$ with $\ell_2\circ \varphi=\ell_1$. Moreover, $\varphi$ is a labelled poset isomorphism if it is a bijection with $\varphi^{-1}$ is a poset isomorphism from $\langle S_2,\le_2,\ell_2\rangle$ to $\langle S_1,\le_1,\ell_1\rangle$. We say that $\mathbf{u}=\langle S_1,\le_1,\ell_1\rangle$ is isomorphic to $\mathbf{v}=\langle S_2,\le_2,\ell_2\rangle$ denoted $\langle S_1,\le_1,\ell_1\rangle\cong\langle S_2,\le_2,\ell_2\rangle$, if there exists a poset isomorphism $\varphi$ between $\langle S_1,\le_1,\ell_1\rangle$ and $\langle S_2,\le_2,\ell_2\rangle$.
\end{definition}

\begin{definition}[Multiset]
A multiset is a kind of set of objects which may be repetitive denoted $\mset{-}$, such that $\mset{a,b,b}$ is significantly distinguishable from $\mset{a,b}$.
\end{definition}

\begin{definition}[Labelled multiset]
Let $\Sigma$ be a fixed set of labels. A labelled multiset is a tuple $\langle S, \ell\rangle$, where $S$ is a multiset, and $\ell$ is a labelling function $\ell:S\to\Sigma$. The set of occurrences / instances of a labelled multiset is denoted lmset.
\end{definition}

\begin{definition}[Pomset]
A partially ordered multiset, pomset, is a $\cong$-equivalence class of posets. The set of pomsets is denoted $\mathsf{Pom}$; the empty pomset is denoted $\mathbf{1}$ and the $\cong$-equivalence class of $1$ is also denoted by $1$. If there does not exist partial orders between any two objects in a pomset, such pomset is called a step denoted $\step{-}$. The set of steps is denoted $\mathsf{Stp}$.
\end{definition}

We assume that the partial order $\le$ can be divided into two kinds: execution order $\le^{e}$ and communication $\le^{c}$. In the same parallel branch, the partial orders usually execution orders and communication usually exists among different parallel branches. Of course, parallel branches can be nested. Then, we can get the following definitions naturally.

\begin{definition}[Labelled poset with communications]
A labelled poset with communications is a tuple $\mathbf{u}=\langle S, \le^{e}, \le^{c}, \ell\rangle$, where $S$ is the carrier set, $\le^{e}$ is an execution order on $S$, $\le^{c}$ is a communication on $S$, and $\ell$ is a labelling function $\ell:S\to\Sigma$. We usually use $\mathbf{u},\mathbf{v}$ to denote labelled posets with communications. And the set of labelled posets with communications is denoted $\mathsf{LPC}$, and the empty labelled poset with communications is $1$.
\end{definition}

\begin{definition}[Labelled poset isomorphism]
Let $\mathbf{u}=\langle S_1,\leq^{e}_1,\leq^{c}_1,\lambda_1\rangle$ and $\mathbf{v}=\langle S_2,\leq^{e}_2, \leq^{c}_2,\lambda_2\rangle$ be labelled posets. A labelled poset morphism $h$ from $\langle S_1,\leq^{e}_1,\leq^{c}_1,\lambda_1\rangle$ to $\langle S_2,\leq^{e}_2,\leq^{c}_2,\lambda_2\rangle$ is a poset morphism from $\langle S_1,\leq^{e}_1,\leq^{c}_1\rangle$ and $\langle S_2,\leq^{e}_2,\leq^{c}_2\rangle$ with $\lambda_2\circ h=\lambda_1$. Moreover, $h$ is a labelled poset isomorphism if it is a bijection with $h^{-1}$ is a poset isomorphism from $\langle S_2,\leq^{e}_2,\leq^{c}_2,\lambda_2\rangle$ to $\langle S_1,\leq^{e}_1,\leq^{c}_1,\lambda_1\rangle$. We say that $\mathbf{u}=\langle S_1,\leq^{e}_1,\leq^{c}_1,\lambda_1\rangle$ is isomorphic to $\mathbf{v}=\langle S_2,\leq^{e}_2,\leq^{c}_2,\lambda_2\rangle$ denoted $\langle S_1,\leq^{e}_1,\leq^{c}_1,\lambda_1\rangle\sim\langle S_2,\leq^{e}_2,\leq^{c}_2,\lambda_2\rangle$, if there exists a poset isomorphism $h$ between $\langle S_1,\leq^{e}_1,\leq^{c}_1,\lambda_1\rangle$ and $\langle S_2,\leq^{e}_2,\leq^{c}_2,\lambda_2\rangle$.
\end{definition}

It is easy to see that $\cong$ is an equivalence and can be used to abstract from the carriers.

\begin{definition}[Pomset with communications]
A partially ordered multiset with communications, pomsetc, is a $\cong$-equivalence class of labelled posets with communications $\mathbf{u}$, written as $[\mathbf{u}]$, i.e., $[\mathbf{u}]=\{\mathbf{v}\in \mathsf{LPC}:\mathbf{u}\cong\mathbf{v}\}$. The set of pomsetcs is also denoted $\mathsf{Pom}$; the empty labelled poset with communications is denoted $1$ and the $\cong$-equivalence class of $1$ is denoted by $1$; the pomsetc containing exactly one action $a\in\Sigma$ is called primitive.
\end{definition}

Concurrency includes parallelism and communication, then, we can get the following definitions of Pomsetc compositions.

\begin{definition}[Pomsetc composition in parallel]
Let $U,V\in\mathsf{Pomc}$ with $U=[\mathbf{u}]$ and $V=[\mathbf{v}]$. We write $U\parallel V$ for the parallel composition of $U$ and $V$, which is the pomsetc represented by $\mathbf{u}\parallel\mathbf{v}$, where

\[S_{\mathbf{u}\parallel\mathbf{v}}=S_{\mathbf{u}}\cup S_{\mathbf{v}}
\quad\quad\le^{e}_{\mathbf{u}\parallel\mathbf{v}}=\le^{e}_{\mathbf{u}}\cup\le^{e}_{\mathbf{v}}
\quad\quad\le^{c}_{\mathbf{u}\parallel\mathbf{v}}=\le^{c}_{\mathbf{u}}\cup\le^{c}_{\mathbf{v}}
\quad\quad\ell_{\mathbf{u}\parallel\mathbf{v}}(x)=\begin{cases}
\ell_{\mathbf{u}},& x\in S_{\mathbf{u}};\\
\ell_{\mathbf{v}},& x\in S_{\mathbf{v}}.
\end{cases}\]
\end{definition}

\begin{definition}[Pomsetc composition in communication]
Let $U,V\in\mathsf{Pomc}$ with $U=[\mathbf{u}]$ and $V=[\mathbf{v}]$. We write $U\mid V$ for the communicative composition of $U$ and $V$, which is the pomsetc represented by $\mathbf{u}\mid\mathbf{v}$, where

\[S_{\mathbf{u}\mid\mathbf{v}}=S_{\mathbf{u}}\cup S_{\mathbf{v}}
\quad\quad\le^{e}_{\mathbf{u}\mid\mathbf{v}}=\le^{e}_{\mathbf{u}}\cup\le^{e}_{\mathbf{v}}
\quad\quad\le^{c}_{\mathbf{u}\mid\mathbf{v}}=\le^{c}_{\mathbf{u}}\cup\le^{c}_{\mathbf{v}}\cup(S_{\mathbf{u}}\times S_{\mathbf{v}})
\quad\quad\ell_{\mathbf{u}\mid\mathbf{v}}(x)=\begin{cases}
\ell_{\mathbf{u}},& x\in S_{\mathbf{u}};\\
\ell_{\mathbf{v}},& x\in S_{\mathbf{v}}.
\end{cases}\]
\end{definition}

\begin{definition}[Pomsetc composition in concurrency]
Let $U,V\in\mathsf{Pomc}$ with $U=[\mathbf{u}]$ and $V=[\mathbf{v}]$. We write $U\between V$ for the concurrent composition of $U$ and $V$, which is the pomsetc represented by $\mathbf{u}\between\mathbf{v}$, where

\[S_{\mathbf{u}\between\mathbf{v}}=S_{\mathbf{u}}\cup S_{\mathbf{v}}
\quad\le^{e}_{\mathbf{u}\between\mathbf{v}}=\le^{e}_{\mathbf{u}}\cup\le^{e}_{\mathbf{v}}
\quad\le^{c}_{\mathbf{u}\between\mathbf{v}}=\le^{c}_{\mathbf{u}}\cup\le^{c}_{\mathbf{v}}
\mbox{ or }\le^{c}_{\mathbf{u}\between\mathbf{v}}=\le^{c}_{\mathbf{u}}\cup\le^{c}_{\mathbf{v}}\cup(S_{\mathbf{u}}\times S_{\mathbf{v}})
\quad\ell_{\mathbf{u}\between\mathbf{v}}(x)=\begin{cases}
\ell_{\mathbf{u}},& x\in S_{\mathbf{u}};\\
\ell_{\mathbf{v}},& x\in S_{\mathbf{v}}.
\end{cases}\]
\end{definition}

\begin{definition}[Pomsetc composition in sequence]
Let $U,V\in\mathsf{Pomc}$ with $U=[\mathbf{u}]$ and $V=[\mathbf{v}]$. We write $U\cdot V$ for the sequential composition of $U$ and $V$, which is the pomsetc represented by $\mathbf{u}\cdot\mathbf{v}$, where

\[S_{\mathbf{u}\cdot\mathbf{v}}=S_{\mathbf{u}}\cup S_{\mathbf{v}}
\quad\quad\le^{e}_{\mathbf{u}\cdot\mathbf{v}}=\le^{e}_{\mathbf{u}}\cup\le^{e}_{\mathbf{v}}\cup(S_{\mathbf{u}}\times S_{\mathbf{v}})
\quad\quad\le^{c}_{\mathbf{u}\cdot\mathbf{v}}=\le^{c}_{\mathbf{u}}\cup\le^{c}_{\mathbf{v}}
\quad\quad\ell_{\mathbf{u}\cdot\mathbf{v}}(x)=\begin{cases}
\ell_{\mathbf{u}},& x\in S_{\mathbf{u}};\\
\ell_{\mathbf{v}},& x\in S_{\mathbf{v}}.
\end{cases}\]
\end{definition}

The following definitions and conclusions are coming from \cite{CKA7}, we retype them.

\begin{definition}[Pomset types]
Let $U\in\mathsf{Pom}$, $U$ is sequential (resp. parallel) if there exist non-empty pomsets $U_1$ and $U_2$ such that $U=U_1\cdot U_2$ (resp. $U=U_1\parallel U_2$).
\end{definition}

\begin{definition}[Factorization]
Let $U\in\mathsf{Pom}$. (1) When $U=U_1\cdot\cdots \cdot U_i\cdot\cdots \cdot U_n$ with each $U_i$ non-sequential and non-empty, the sequence $U_1,\cdots,U_i,\cdots,U_n$ is called a sequential factorization of $U$. (2) When $U=U_1\parallel\cdots \parallel U_i\parallel\cdots \parallel U_n$ with each $U_i$ non-parallel and non-empty, the multiset $\mset{U_1,\cdots,U_i,\cdots,U_n}$ is called a parallel factorization of $U$.
\end{definition}

\begin{lemma}[Factorization]\label{LemmaFactorization}
Sequential and parallel factorizations exist uniquely.
\end{lemma}

\begin{lemma}
For $U\in\mathsf{Pomc}$, then the following two conclusions hold:

\begin{enumerate}
  \item $U$ is either sequential or parallel, and there are not other types in $U$.
  \item Sequential and parallel factorizations exist in $U$ uniquely.
\end{enumerate}
\end{lemma}

\begin{definition}[Series-parallel pomset]
The set of series-parallel pomset, or sp-pomsets denoted $\mathsf{SP}$, is the smallest set satisfying the following rules:

\[\frac{}{1\in\mathsf{SP}}
\quad\frac{a\in\Sigma}{a\in\mathsf{SP}}
\quad\frac{U,V\in\mathsf{SP}}{U\cdot V\in\mathsf{SP}}
\quad\frac{U,V\in\mathsf{SP}}{U\parallel V\in\mathsf{SP}}\]
\end{definition}

\begin{definition}[Series-communication-parallel pomsetc]
The set of series-communication-parallel pomsetcs, or scp-pomsetcs denoted $\mathsf{SCP}$, is the smallest set satisfying the following rules:

\[\frac{}{1\in\mathsf{SCP}}
\quad\frac{a\in\Sigma}{a\in\mathsf{SCP}}
\quad\frac{U,V\in\mathsf{SCP}}{U\cdot V\in\mathsf{SCP}}
\quad\frac{U,V\in\mathsf{SCP}}{U\parallel V\in\mathsf{SCP}}
\quad\frac{U,V\in\mathsf{SCP}}{U\mid V\in\mathsf{SCP}}
\quad\frac{U,V\in\mathsf{SCP}}{U\between V\in\mathsf{SCP}}\]
\end{definition}

\begin{definition}[N-shape1]\label{nshape1s}
Let $U=[\mathbf{u}]$ be a pomset. An N-shape1 in $U$ is a quadruple $u_0,u_1,u_2,u_3\in S_{\mathbf{u}}$ of distinct points such that $u_0\le_{\mathbf{u}}u_1$, $u_2\le_{\mathbf{u}}u_3$ and $u_0\le_{\mathbf{u}}u_3$ and their exists no other relations among them. A pomset $U$ is N-free if it has no N-shape1s.
\end{definition}

\begin{definition}[N-shape2]\label{nshape2s}
Let $U=[\mathbf{u}]$ be a pomsetc. An N-shape2 in $U$ is a quadruple $u_0,u_1,u_2,u_3\in S_{\mathbf{u}}$ of distinct points such that $u_0\le^{e}_{\mathbf{u}}u_1$, $u_2\le^{e}_{\mathbf{u}}u_3$ and $u_0\le^{c}_{\mathbf{u}}u_3$ and their exists no other relations among them.
\end{definition}

The definition of N-shape2 in Definition \ref{nshape2s} is based on the assumption that partial orders (causalities) among different parallel branches are all communications.

\begin{theorem}[N-shape1]
A pomset is series-parallel if and only if it is N-shape1-free in Definition \ref{nshape1s}.
\end{theorem}

\begin{theorem}[N-shape2]
A pomsetc is series-communication-parallel if and only if it only contains N-shape2s in Definition \ref{nshape2s}.
\end{theorem}

\begin{theorem}[Series-communication-parallelism to series-parallelism]
A series-communication-parallel pomsetc $U$ can be translated into a series-parallel pomset $U'$ if all the communications are all synchronous, i.e., for all $u_i\le^{c} u_j$ in $U$, $u_i,u_j$ can merge into a single $\gamma(u_i,u_j)$ in $U'$, where $\gamma(u_i,u_j)$ is the communication function between $u_i$ and $u_j$.
\end{theorem}

So, in the following chapters, assume that all pomsets are series-parallel and all pomsetcs are series-communication-parallel. With a little of abuse of notions, we use $\mathsf{Pom}$, $\mathsf{Pomc}$, $\mathsf{SP}$ and $\mathsf{SCP}$ without distinctions.

\subsection{Category}

\begin{definition}[Category]
A category $\mathbb{C}$ consists of a collection $\mathbf{Obj}(\mathbb{C})$ of objects and a collection $\mathbf{Arr}(\mathbb{C})$ of arrows (also called maps or morphisms). We write $X\in\mathbb{C}$ for $X\in\mathbf{Obj}(\mathbb{C})$. Each arrow in $\mathbb{C}$ written as $X\xrightarrow{f}Y$ or as $f:X\to Y$, has a domain object $X\in\mathbb{C}$ and a codomain object $Y\in\mathbb{C}$. And also a category $\mathbb{C}$ satisfies:

\begin{enumerate}
  \item A composition operator $\circ$ assigns to each pair of arrows $f:X\to Y$ and $g:Y\to Z$ a composite arrow $g\circ f:X\to Z$, which satisfying the following associative law:
  
  For any arrows $f:X\to Y$, $g:Y\to Z$ and $h:Z\to W$, it holds that $h\circ(g\circ f)=(h\circ g)\circ f$.
  \item For each object $X\in\mathbb{C}$, an identity arrow $\mathrm{id}_{X}:X\to X$ satisfies the following indentity law:
  
  For any arrow $f:X\to Y$, they hold that $\mathrm{id}_{Y}\circ f=f$ and $f\circ\mathrm{id}_{X}=f$.
\end{enumerate} 
\end{definition} 

\begin{definition}[Commutative diagram]
A diagram in a category $\mathbb{C}$ is a collection of vertices and directed edges, consistently labelled with objects and arrows in $\mathbb{C}$. A diagram in a category $\mathbb{C}$ is said to commute if for every pair of vertices $X$ and $Y$, all the paths in the diagram from $X$ to $Y$ are equal.

\begin{center}
\begin{tikzcd}[row sep=huge, column sep=huge]
W \arrow[r,"f"]  & Y \\
X \arrow[r,"f'"'] \arrow[u,"g'"]& Z  \arrow[u,"g"']
\end{tikzcd}
\end{center}

To say that the above diagram commutes is equal to say that $f\circ g'=g\circ f'$.

Especially in the diagram, the uniqueness of a morphism $f:X\to Y$ is expressed by a dashed arrow $f:X\dashrightarrow Y$.
\end{definition}

\begin{proposition}[Labelled multiset category $\mathbf{LMSets}$]
An object of $\mathbf{LMSets}$ is a pair $\langle X,\ell_{X}\rangle$ which is called an lmset, where

\begin{itemize}
  \item $X$ is a set (the set of occurrences / instances of the multiset).
  \item $\ell_{X}:X\to \Sigma_{X}$ is a function, called the labelling map, and $\Sigma_{X}$ is some set of the label alphabet.
\end{itemize}

We write such an object as $\langle X,\ell_{X}:X\to \Sigma_{X}\rangle$. This is a labelled multiset: each occurrence $x\in X$ carries a label $\ell_{X}(x)\in\Sigma_{X}$. Different objects may have different label alphabets.

Let $\langle X,\ell_{X}:X\to \Sigma_{X}\rangle$, $\langle Y,\ell_{Y}:Y\to \Sigma_{Y}\rangle$ be two objects in $\mathbf{LMSets}$. A morphism $\langle X, \ell_{X}\rangle \to \langle Y,\ell_{Y}\rangle$ is a pair $\langle f,\sigma\rangle : \langle X,\ell_{X}:X\to\Sigma_{X}\rangle \to \langle Y,\ell_{Y}:Y\to\Sigma_{Y}\rangle$, where

\begin{enumerate}
  \item $f: X\to Y$ is a function on occurrences.
  \item $\sigma: \Sigma_{X}\to \Sigma_{Y}$ is a function on labels (the uniform renaming map: every occurrence of a label $a\in\Sigma_{X}$ is renamed to the same $\sigma(a)\in\Sigma_{Y}$).
\end{enumerate}

such that the following diagram commutes:

\[\begin{tikzcd}[row sep=huge, column sep=huge]
\Sigma_{X} \arrow[r, "\sigma"] & \Sigma_{Y}\\
X \arrow[r, "f"'] \arrow[u, "\ell_{X}"] & Y \arrow[u, "\ell_{Y}"'] 
\end{tikzcd}\]

Formally, the required equation is $\ell_{Y} \circ f = \sigma \circ \ell_{X}$. For every $x\in X$, this says $\ell_{Y}(f(x)) = \sigma(\ell_{X}(x))$.

Suppose we have two composable morphisms: $\langle X,\ell_{X}\rangle\xrightarrow{\langle f,\sigma\rangle } \langle Y, \ell_{Y}\rangle$ and $\langle Y,\ell_{Y}\rangle\xrightarrow{\langle g,\xi\rangle} \langle Z,\ell_{Z}\rangle$, their composition is defined by $\langle g,\xi\rangle \circ \langle f,\sigma\rangle  \triangleq \langle g\circ f,\; \xi\circ \sigma\rangle$. Here $g\circ f:X\to Z$ is ordinary function composition on occurrences, and $\xi\circ\sigma:\Sigma_{X}\to\Sigma_{Z}$ is ordinary function composition on labels.

For any object $\langle X,\ell_{X}:X\to\Sigma_{X}\rangle$, its identity morphism is $\mathrm{id}_X \triangleq \langle \mathrm{id}_X, \mathrm{id}_{\Sigma_{X}}\rangle$, where $\mathrm{id}_X:X\to X$ is the identity function on occurrences, and $\mathrm{id}_{\Sigma_{X}}:\Sigma_{X}\to\Sigma_{X}$ is the identity function on labels.

$\mathbf{LMSets}$ forms a category.
\end{proposition}

\begin{definition}[Equalizer]
An arrow $e:X\to Y$ is an equalizer of a pair of arrows $f:Y\to Z$ and $g:Y\to Z$, if:

\begin{enumerate}
  \item $f\circ e=g\circ e$.
  \item Whenever $e':X'\to Y$ satisfies $f\circ e'=g\circ e'$, there is a unique arrow $k:X'\to X$ such that $e\circ k=e'$.
\end{enumerate}

\[\begin{tikzcd}[row sep=huge, column sep=huge]
X \arrow[r, "e"] & Y \arrow[r,shift left, "f"] \arrow[r,shift right,"g"']& Z\\
X' \arrow[dashed,u, "k"] \arrow[ru, "e'"'] &&
\end{tikzcd}\]
\end{definition}

\begin{definition}[Functor]
Assume two categories $\mathbb{C}$ and $\mathbb{D}$, a functor $F:\mathbb{C}\to\mathbb{D}$ consists of two mappings $\mathbf{Obj}(\mathbb{C})\to\mathbf{Obj}(\mathbb{D})$, i.e., mapping each $\mathbb{C}$-object $X\in\mathbb{C}$ to a $\mathbb{D}$-object $F(X)\in\mathbb{D}$, and $\mathbf{Arr}(\mathbb{C})\to\mathbf{Arr}(\mathbb{D})$, i.e., mapping each $\mathbb{C}$-arrow $f:X\to Y$ to a $\mathbb{D}$-arrow $F(f):F(X)\to F(Y)$, such that:

\begin{enumerate}
  \item $F$ preserves domains and codomains: if $f:X\to Y$ in $\mathbb{C}$, then $F(f):F(X)\to F(Y)$ in $\mathbb{D}$.
  \item $F$ preserves identities: $F(\mathrm{id}_{X})=\mathrm{id}_{F(X)}$, for each $X\in\mathbb{C}$.
  \item $F$ preserves compositions: $F(g\circ f)=F(g)\circ F(f)$, for all maps $f:X\to Y$ and $g:Y\to Z$ in $\mathbb{C}$.
\end{enumerate}
\end{definition}

Endofunctors are functors $\mathbb{C}\to\mathbb{C}$ for some category $\mathbb{C}$. 

\begin{definition}[Coalgebra]
Let $\mathbb{C}$ be an arbitrary category equipped with an endofunctor $F:\mathbb{C}\to\mathbb{C}$.

\begin{enumerate}
  \item An $F$-coalgebra or just called a coalgebra consists of an object $X\in\mathbb{C}$ together with a morphism $c:X\to F(X)$, as before, $X$ is called the state space and $c$ is called the transition or coalgebra structure.
  \item A homomorphism of coalgebras from one coalgebra $c:X\to F(X)$ to another coalgebra $d:Y\to F(Y)$ consists of a morphism $f:X\to Y$ in $\mathbb{C}$, which makes the following diagram commute:
      
      \begin{center}
      \begin{tikzcd}[row sep=huge, column sep=huge]
        F(X) \arrow[r,"F(f)"] & F(Y)  \\
        X \arrow[r,"f"'] \arrow[u,"c"] & Y \arrow[u,"d"']
      \end{tikzcd}
      \end{center}
  \item $F$-coalgebra with the above homomorphisms between them form a category, denoted $\mathbf{CoAlg}(F)$, with a forgetful functor $\mathbf{CoAlg}(F)\to\mathbb{C}$, mapping a coalgebra $X\to F(X)$ to its state space $X$ and a coalgebra homomorphism $f$ to $f$.
\end{enumerate}
\end{definition} 

\begin{definition}[Algebra]
Let $\mathbb{C}$ be an arbitrary category with an endofunctor $F:\mathbb{C}\to\mathbb{C}$.

\begin{enumerate}
  \item An $F$-algebra or just called an algebra consists of an object $X\in\mathbb{C}$ together with a morphism $a:F(X)\to X$, as often, it is called the constructor or operation.
  \item A homomorphism of algebras from one algebra $a:F(X)\to X$ to another algebra $b:F(Y)\to Y$ consists of a morphism $f:X\to Y$ in $\mathbb{C}$, which makes the following diagram commute:
      
      \begin{center}
      \begin{tikzcd}[row sep=huge, column sep=huge]
        F(X) \arrow[r,"F(f)"]\arrow[d,"a"'] & F(Y) \arrow[d,"b"] \\
        X \arrow[r,"f"']  & Y 
      \end{tikzcd}
      \end{center}
  \item $F$-algebra with the above homomorphisms between them form a category, denoted $\mathbf{Alg}(F)$.
\end{enumerate}
\end{definition}

\begin{definition}[Bialgebra]
A bialgebra consists of an algebra-coalgebra pair $F(X)\to X\to G(X)$ on a common state space $X$, where $F,G:\mathbb{C}\to\mathbb{C}$ are two endofunctors on the same category $\mathbb{C}$.
\end{definition} 
\newpage\section{Intelligence Phase}\label{ip} 

In this section, we introduce the intelligence phase. During the intelligence phase, the user consults with the LLM to produce the orchestration and choreography of AI agent workflow. The main structure of an LLM is based on transformers \cite{AIAYN}. The transformer structure is decomposed into atomic actions and then assembled together \cite{TTCB}, as \cref{DoT} illustrates.

\begin{center}
    \begin{longtable}{|c|c|c|}
      \caption{Decomposition of Transformers}\\
      \hline Names &Functions& Types\\
      \hline
      \endfirsthead
      \multicolumn{3}{c}{\bfseries continuing: Decomposition of Transformers} \\
      \hline Names &Functions& Types\\
      \hline
      \endhead
      \hline
      \multicolumn{3}{r}{to be continued\ldots} \\
      \endfoot
      \hline
      \multicolumn{3}{r}{end of Decomposition of Transformers} \\
      \endlastfoot
      $a_{1}$ & word embedding & $\Sigma\rightarrow \mathbb{R}^{d}$\\
      $a_{2}$ & position embedding & $\mathbb{N}\rightarrow\mathbb{R}^{d}$\\
      $a_{3}$ & identity & $\mathbb{R}^{d}\rightarrow\mathbb{R}^{d}$\\
      $a_{4}$ & min/max & $\mathbb{R}^{2}\rightarrow\mathbb{R}$\\
      $a_{5}$ & add/subtract & $\mathbb{R}^{2}\rightarrow\mathbb{R}$\\
      $a_{6}$ & multiply by $c$ & $\mathbb{R}\rightarrow\mathbb{R}$\\
      $a_{7}$ & multiply $xy$ & $\mathbb{R}^{2}\rightarrow\mathbb{R}$\\
      $a_{8}$ & comparators & $\mathbb{R}\rightarrow\{0,1\}$\\
      $a_{9}$ & boolean functions & $\{0,1\}^{m}\rightarrow\{0,1\}$\\
      $a_{10}$ & continuous piecewise linear functions & $\mathbb{R}^{d}\rightarrow\mathbb{R}$\\
      $a_{11}$ & cancel residual & $\mathbb{R}^{d}\rightarrow\mathbb{R}^{d}$\\
      $a_{12}$ & identity & $(\mathbb{R}^{d})^{+}\rightarrow(\mathbb{R}^{d})^{+}$\\
      $a_{13}$ & uniform average & $(\mathbb{R}^{d})^{+}\rightarrow(\mathbb{R}^{d})^{+}$\\
      $a_{14}$ & first & $(\mathbb{R}^{d})^{+}\rightarrow(\mathbb{R}^{d})^{+}$\\
      $a_{15}$ & index lookup & $(\mathbb{R}^{d})^{+}\rightarrow\mathbb{R}^{+}$\\
      $a_{16}$ & predecessor & $(\mathbb{R}^{d})^{+}\rightarrow(\mathbb{R}^{d})^{+}$\\
      $a_{17}$ & tie-breaking & $--$\\
      $a_{18}$ & multi-head & $--$\\
      $a_{19}$ & unembedding & $--$\\
      $a_{20}$ & layer normalization & $\mathbb{R}^{d}\rightarrow\mathbb{R}^{d}$\\
      \label{DoT}
    \end{longtable}
\end{center}

Then the category of transformers $\mathbf{Trans}$ can be deemed as $\mathbf{LMSets}$ equipped with an execution partial order which is N-free. We then introduce the coalgebra of $\mathbf{Trans}$, and also the algebra of transformers which is introduced in \cite{AT} firstly.

\subsection{Category of Transformers $\mathbf{Trans}$}

\begin{definition}[Category of transformers $\mathbf{Trans}$]
An object in $\mathbf{Obj}(\mathbf{Trans})$ is a triple $\langle X,\le^e_X,\ell_X:X\to \Sigma_X\rangle$ where:

\begin{itemize}
  \item $X$ is a set (the set of occurrences / instances of the multiset).
  \item $\le^e_X$ is an execution partial order on $X$ (reflexive, antisymmetric, transitive), and $\langle X,\le^e_X\rangle$ is an N-free poset.
  \item $\Sigma_X$ is a set (label alphabet of $X$).
  \item $\ell_X: X\to \Sigma_X$ is a function (labelling map).
\end{itemize}

Given $\langle X,\le^e_X,\ell_X:X\to\Sigma_X\rangle$, $\langle Y,\le^e_Y,\ell_Y:Y\to\Sigma_Y\rangle$, a morphism $\langle f,\sigma\rangle : \langle X,\le^e_X,\ell_X:X\to\Sigma_X\rangle\to \langle Y,\le^e_Y,\ell_Y:Y\to\Sigma_Y\rangle$ is a pair $(f,\sigma)$ with $f: X\to Y,\sigma: \Sigma_X\to \Sigma_Y$ satisfying:

\begin{enumerate}
  \item Commutativity: $\ell_Y\circ f = \sigma \circ \ell_X$.
  \item Monotonicity: $\forall x_1,x_2\in X.x_1\le^e_X x_2 \Leftrightarrow f(x_1)\le^e_Y f(x_2)$.
\end{enumerate}

If $\langle f,\sigma\rangle: \langle X,\le^e_X,\ell_X:X\to\Sigma_X\rangle\to \langle Y,\le^e_Y,\ell_Y:Y\to\Sigma_Y\rangle$, $\langle g,\xi\rangle: \langle X,\le^e_Y,\ell_Y:Y\to\Sigma_Y\rangle\to \langle Z,\le^e_Z,\ell_Z:Z\to\Sigma_Z\rangle$, define $\langle g,\xi\rangle \circ \langle f,\sigma\rangle \triangleq \langle g\circ f, \xi\circ \sigma\rangle$.

For $A=\langle X,\le^e_X,\ell_X:X\to\Sigma_X\rangle$, $\mathrm{id}_A \triangleq \langle \mathrm{id}_X,\mathrm{id}_{\Sigma_X}\rangle$.

$\mathbf{Trans}$ forms a category.
\end{definition}

\begin{proof}
We assume $\langle f,\sigma\rangle: \langle X,\le^e_X,\ell_X:X\to\Sigma_X\rangle\to \langle Y,\le^e_Y,\ell_Y:Y\to\Sigma_Y\rangle, \langle g,\xi\rangle: \langle Y,\le^e_Y,\ell_Y:Y\to\Sigma_Y\rangle\to \langle Z,\le^e_Z,\ell_Z:Z\to\Sigma_Z\rangle$ are morphisms, and prove $\langle g\circ f,\xi\circ\sigma\rangle$ satisfies the two morphism conditions.

\begin{enumerate}
  \item Commutativity:
\[\begin{aligned}
\ell_C \circ (g\circ f)
&= (\ell_C \circ g)\circ f \\
&= (\xi \circ \ell_Y)\circ f \\
&= \xi \circ (\ell_Y \circ f) \\
&= \xi \circ (\sigma \circ \ell_X) \\
&= (\xi\circ \sigma)\circ \ell_X.
\end{aligned}\]

So $\ell_C \circ (g\circ f) = (\xi\circ\sigma)\circ \ell_X$.
  \item Monotonicity: take $x_1,x_2\in X$ with $x_1\le^e_X x_2$. Since $\langle f,\sigma\rangle$ is monotone, $f(x_1)\le^e_Y f(x_2)$. Since $\langle g,\xi\rangle$ is monotone, $g(f(x_1)) \le_Z^e g(f(x_2))$. Thus $g\circ f$ is monotone.
\end{enumerate}

Therefore $\langle g\circ f,\xi\circ\sigma\rangle$ is indeed a morphism $\langle X,\le^e_X,\ell_X:X\to\Sigma_X\rangle\to \langle Z,\le^e_Z,\ell_Z:Z\to\Sigma_Z\rangle$.

For $A=\langle X,\le^e_X,\ell_X:X\to\Sigma_X\rangle$, $\mathrm{id}_A = \langle \mathrm{id}_X,\mathrm{id}_{\Sigma_X}\rangle$. Check the two conditions:

\begin{enumerate}
  \item Commutativity: $\ell_X \circ \mathrm{id}_X= \ell_X= \mathrm{id}_{\Sigma_X}\circ \ell_X$.
  \item Monotonicity: if $x_1\le^e_X x_2$, then $\mathrm{id}_X(x_1)=x_1\le^e_X x_2=\mathrm{id}_X(x_2)$.
\end{enumerate}

Hence $\mathrm{id}_A : A\to A$ is a morphism.

Then, we prove the associativity of composition.

Take three composable morphisms: $\langle f,\sigma\rangle: \langle X,\le^e_X,\ell_X:X\to\Sigma_X\rangle\to \langle Y,\le^e_Y,\ell_Y:Y\to\Sigma_Y\rangle$,$\langle g,\xi\rangle: \langle Y,\le^e_Y,\ell_Y:Y\to\Sigma_Y\rangle\to \langle Z,\le^e_Z,\ell_Z:Z\to\Sigma_Z\rangle$,$\langle h,\zeta\rangle: \langle Z,\le^e_Z,\ell_Z:Z\to\Sigma_Z\rangle\to \langle W,\le^e_W,\ell_W:W\to\Sigma_W\rangle$. By definition:

\[\begin{aligned}
(\langle h,\zeta\rangle \circ \langle g,\xi\rangle)\circ \langle f,\sigma\rangle&=\langle h\circ g,\zeta\circ\xi\rangle\circ \langle f,\sigma\rangle \\
&=\langle (h\circ g)\circ f, (\zeta\circ\xi)\circ \sigma\rangle
\end{aligned}\]

and

\[\begin{aligned}
\langle h,\zeta\rangle\circ(\langle g,\xi\rangle \circ \langle f,\sigma\rangle)&=\langle h,\zeta\rangle\circ\langle g\circ f,\xi\circ\sigma\rangle \\
&=\langle h\circ(g\circ f), \zeta\circ(\xi\circ\sigma)\rangle
\end{aligned}\]

Function composition on sets is associative: $(h\circ g)\circ f = h\circ(g\circ f),(\zeta\circ\xi)\circ\sigma = \zeta\circ(\xi\circ\sigma)$. Therefore the two composite morphisms are equal: $(\langle h,\zeta\rangle \circ \langle g,\xi\rangle)\circ \langle f,\sigma\rangle=\langle h,\zeta\rangle\circ(\langle g,\xi\rangle \circ \langle f,\sigma\rangle)$.

Lastly, we prove the identity laws.

Let $\langle f,\sigma\rangle: A=\langle X,\le^e_X,\ell_X:X\to\Sigma_X\rangle\to B=\langle Y,\le^e_Y,\ell_X:Y\to\Sigma_Y\rangle$.

\[\begin{aligned}
\langle f,\sigma\rangle \circ \mathrm{id}_A&=\langle f,\sigma\rangle\circ\langle \mathrm{id}_X,\mathrm{id}_{\Sigma_X}\rangle \\
&=\langle f\circ \mathrm{id}_X, \sigma\circ \mathrm{id}_{\Sigma_X}\rangle
&=\langle f,\sigma\rangle
\end{aligned}\]

\[\begin{aligned}
\mathrm{id}_B \circ \langle f,\sigma\rangle&=\langle \mathrm{id}_Y,\mathrm{id}_{\Sigma_Y}\rangle\circ\langle f,\sigma\rangle \\
&=\langle \mathrm{id}_Y \circ f,\; \mathrm{id}_{\Sigma_Y}\circ \sigma\rangle
&=\langle f,\sigma\rangle
\end{aligned}\]

Therefore, $\mathbf{Trans}$ is a category.
\end{proof}

\subsubsection{Polynomial Functors of $\mathbf{Trans}$}

\runhead{Products}

For lmsets $\langle X,\le^e_X,\ell_{X}\rangle$ and $\langle Y,\le^e_Y,\ell_{Y}\rangle$, their products written as $\langle X,\le^e_X,\ell_{X}\rangle\times \langle Y,\le^e_Y,\ell_{Y}\rangle$, is defined as:

\[\langle X,\le^e_X,\ell_{X}\rangle\times \langle Y,\le^e_Y,\ell_{Y}\rangle=\{(x,y)|x\in \langle X,\le^e_X,\ell_{X}\rangle\wedge y\in \langle Y,\le^e_Y,\ell_{Y}\rangle\}\]

The product operation $\times$ is a functor $\mathbf{Trans}\times\mathbf{Trans}\to\mathbf{Trans}$.

\begin{definition}[Product]\label{definitionofproduct}
Let us take two objects $\langle X,\le^e_X,\ell_X\rangle$ and $\langle Y,\le^e_Y,\ell_Y\rangle$ in $\mathbf{Trans}$. The product of these two objects is a new object $\langle X,\le^e_X,\ell_{X}\rangle\times\langle Y,\le^e_Y,\ell_{Y}\rangle$ with two projection morphisms:

\[\begin{tikzcd}[row sep=huge, column sep=huge]
\langle X,\le^e_X,\ell_{X}\rangle & \langle X,\le^e_X,\ell_{X}\rangle\times\langle Y,\le^e_Y,\ell_{Y}\rangle \arrow[l, "\pi_1"'] \arrow[r, "\pi_2"] & \langle Y,\le^e_Y,\ell_{Y}\rangle 
\end{tikzcd}\]

For functions $f:\langle Z,\le^e_Z,\ell_{Z}\rangle\to\langle X,\le^e_X,\ell_{X}\rangle$ and $g:\langle Z,\le^e_Z,\ell_{Z}\rangle\to\langle Y,\le^e_Y,\ell_{Y}\rangle$, (where $f$ contains a pair of functions $f:\langle f_{f},f_{\sigma}\rangle : \langle Z,\le^e_Z,\ell_{Z}:Z\to\Sigma_{Z}\rangle \to \langle X,\le^e_X,\ell_{X}:X\to\Sigma_{X}\rangle$ and where $g$ also contains a pair of functions $g:\langle g_{g},g_{\xi}\rangle : \langle Z,\le^e_Z,\ell_{Z}:Z\to\Sigma_{Z}\rangle \to \langle Y,\le^e_Y,\ell_{Y}:Y\to\Sigma_{Y}\rangle$, in the following chapters, if no needs, we do not expand them and use their collapsed forms), there is a unique pairing function $\langle f,g\rangle:\langle Z,\le^e_Z,\ell_{Z}\rangle\to\langle X,\le^e_X,\ell_{X}\rangle\times\langle Y,\le^e_Y,\ell_{Y}\rangle$ with $\langle f,g\rangle(z)=(f(z),g(z))\in \langle X,\le^e_X,\ell_{X}\rangle\times\langle Y,\le^e_Y,\ell_{Y}\rangle$ for $z\in\langle Z,\le^e_Z,\ell_{Z}\rangle$, such that the following equations hold:

\[\pi_1\circ\langle f,g\rangle=f\qquad \langle \pi_1,\pi_2\rangle=\mathrm{id}_{\langle X,\le^e_X,\ell_{X}\rangle\times\langle Y,\le^e_Y,\ell_{Y}\rangle}\]
\[\pi_2\circ\langle f,g\rangle=g\qquad \langle f,g\rangle\circ h=\langle f\circ h,g\circ h\rangle\]

where $h:\langle W,\le^e_W,\ell_{W}\rangle\to\langle Z,\le^e_Z,\ell_{Z}\rangle$.

For the above morphisms $f$ and $g$, there is a unique pairing morphism $\langle f,g\rangle$ defined as above, making the following diagram commute:

\[\begin{tikzcd}[row sep=huge, column sep=huge]
\langle X,\le^e_X,\ell_{X}\rangle & \langle X,\le^e_X,\ell_{X}\rangle\times\langle Y,\le^e_Y,\ell_{Y}\rangle \arrow[l, "\pi_1"'] \arrow[r, "\pi_2"] & \langle Y,\le^e_Y,\ell_{Y}\rangle \\
&\langle Z,\le^e_Z,\ell_{Z}\rangle \arrow[lu,"f"] \arrow[dashed,u,"\ensuremath{\langle f,g\rangle}"description] \arrow[ru,"g"']&
\end{tikzcd}\]
\end{definition}

For functions $f:\langle X,\le^e_X,\ell_{X}\rangle\to\langle X',\le^e_{X'},\ell_{X'}\rangle$ and $g:\langle Y,\le^e_Y,\ell_{Y}\rangle\to\langle Y',\le^e_{Y'},\ell_{Y'}\rangle$, the product of functions $f\times g$ is defined as:

\[\begin{tikzcd}[row sep=huge, column sep=huge]
\langle X,\le^e_X,\ell_{X}\rangle\times\langle Y,\le^e_Y,\ell_{Y}\rangle \arrow[r, "f\times g"] & \langle X',\le^e_{X'},\ell_{X'}\rangle\times\langle Y',\le^e_{Y'},\ell_{Y'}\rangle 
\end{tikzcd}\] 

which is given by $(x,y)\mapsto(f(x),g(y))$, where $x\in\langle X,\le^e_X,\ell_{X}\rangle$ and $y\in\langle Y,\le^e_Y,\ell_{Y}\rangle$, such that the following equations hold:

\[f\times g=\langle f\circ\pi_1,g\circ\pi_2\rangle\qquad \mathrm{id}_{\langle X,\le^e_X,\ell_{X}\rangle}\times\mathrm{id}_{\langle Y,\le^e_Y,\ell_{Y}\rangle}=\mathrm{id}_{\langle X,\le^e_X,\ell_{X}\rangle\times\langle Y,\le^e_Y,\ell_{Y}\rangle}\]
\[(f\circ h)\times(g\circ k)=(f\times g)\circ(h\times k)\]

Two functions $\langle Z,\le^e_Z,\ell_{Z}\rangle\to\langle X,\le^e_X,\ell_{X}\rangle$ and $\langle Z,\le^e_Z,\ell_{Z}\rangle\to\langle Y,\le^e_Y,\ell_{Y}\rangle$ can be composed into $\langle Z,\le^e_Z,\ell_{Z}\rangle\to\langle X,\le^e_X,\ell_{X}\rangle\times\langle Y,\le^e_Y,\ell_{Y}\rangle$ by \cref{definitionofproduct} and $\langle Z,\le^e_Z,\ell_{Z}\rangle\to\langle X,\le^e_X,\ell_{X}\rangle\times\langle Y,\le^e_Y,\ell_{Y}\rangle$ can be decomposed into $\langle Z,\le^e_Z,\ell_{Z}\rangle\to\langle X,\le^e_X,\ell_{X}\rangle$ and $\langle Z,\le^e_Z,\ell_{Z}\rangle\to\langle Y,\le^e_Y,\ell_{Y}\rangle$ by use of projections $\pi_1$ and $\pi_2$, which can be expressed by the following doubled line rule working in two directions:

\[\doublefrac{\langle Z,\le^e_Z,\ell_{Z}\rangle\to\langle X,\le^e_X,\ell_{X}\rangle\quad \langle Z,\le^e_Z,\ell_{Z}\rangle\to\langle Y,\le^e_Y,\ell_{Y}\rangle}{\langle Z,\le^e_Z,\ell_{Z}\rangle\to\langle X,\le^e_X,\ell_{X}\rangle\times\langle Y,\le^e_Y,\ell_{Y}\rangle}\] 

The $n$-fold product called power is defined as $\langle X,\le^e_X,\ell_{X}\rangle^{n}=\langle X,\le^e_X,\ell_{X}\rangle\times\cdots\times\langle X,\le^e_X,\ell_{X}\rangle$ for a given $\langle X,\le^e_X,\ell_{X}\rangle$ and $n\in\mathbb{N}$. And the empty product $\langle X,\le^e_X,\ell_{X}\rangle^{0}$ is called a final or terminal object.

\begin{definition}[Final object]
A final object in a category $\mathbb{C}$ is an object $1\in\mathbb{C}$, such that for each object $X\in\mathbb{C}$ there is a unique morphism $!_{X}:X\to 1$ in $\mathbb{C}$.
\end{definition} 

\begin{proposition}[Final object of $\mathbf{Trans}$]
$\mathbf{Trans}$ has a final object.
\end{proposition}

\begin{proof}
Any singleton labelled occurrence multiset is final.
\end{proof}

$\mathbf{Trans}$ has finite products because it has binary products and a final object $1$. Finite products can be generalized to set-indexed products: given an index set $I$, a collection $(\langle X,\le^e_X,\ell_{X}\rangle)_{i\in I}$ of $I$-indexed objects has $I$-indexed products $\langle X,\le^e_X,\ell_{X}\rangle=\prod_{i\in I}\langle X_{i},\le^e_{X_{i}},\ell_{X_{i}}\rangle$ with projections $\pi_{i}:\langle X,\le^e_X,\ell_{X}\rangle\to\langle X_{i},\le^e_{X_{i}},\ell_{X_{i}}\rangle$ for $i\in I$, which are universal as in \cref{definitionofproduct}: for an arbitrary object $\langle Y,\le^e_Y,\ell_{Y}\rangle$ and an $I$-indexed collection $f_{i}:\langle Y,\le^e_Y,\ell_{Y}\rangle\to\langle X_{i},\le^e_{X_{i}},\ell_{X_{i}}\rangle$ of morphisms there is a unique morphism $f=\langle f_{i}\rangle:\langle Y,\le^e_Y,\ell_{Y}\rangle\to\langle X,\le^e_X,\ell_{X}\rangle$ with $\pi_{i}\circ f=f_{i}$ for $i\in I$. In the category $\mathbf{Trans}$, such products exist and may be described as:

\[\prod_{i\in I}\langle X_{i},\le^e_{X_{i}},\ell_{X_{i}}\rangle=\{t:I\to\bigcup_{i\in I}\langle X_{i},\le^e_{X_{i}},\ell_{X_{i}}\rangle|\forall i\in I.t(i)\in\langle X_{i},\le^e_{X_{i}},\ell_{X_{i}}\rangle\}\] 

\runhead{Coproducts}

For lmsets $\langle X,\le^e_X,\ell_{X}\rangle$ and $\langle Y,\le^e_Y,\ell_{Y}\rangle$, their coproducts written as $\langle X,\le^e_X,\ell_{X}\rangle+\langle Y,\le^e_Y,\ell_{Y}\rangle$, is defined as:

\[\langle X,\le^e_X,\ell_{X}\rangle+\langle Y,\le^e_Y,\ell_{Y}\rangle=\{(x,1)|x\in \langle X,\le^e_X,\ell_{X}\rangle\}\cup\{(y,2)|y\in \langle Y,\le^e_Y,\ell_{Y}\rangle\}\]

where $1$ and $2$ force the above union to be disjoint.

The coproduct operation $+$ is also a functor $\mathbf{Trans}\times\mathbf{Trans}\to\mathbf{Trans}$, actually, it is the dual of product $\times$.

\begin{definition}[Coproduct]\label{definitionofcoproduct}
Let us take two objects $\langle X,\le^e_X,\ell_X\rangle $ and $\langle Y,\le^e_Y,\ell_Y\rangle $ in $\mathbf{Trans}$. The coproduct of these two objects is a new object $\langle X,\le^e_X,\ell_{X}\rangle+\langle Y,\le^e_Y,\ell_{Y}\rangle$ with two coprojection morphisms:

\[\begin{tikzcd}[row sep=huge, column sep=huge]
\langle X,\le^e_X,\ell_{X}\rangle \arrow[r, "\kappa_1"] & \langle X,\le^e_X,\ell_{X}\rangle+\langle Y,\le^e_Y,\ell_{Y}\rangle & \langle Y,\le^e_Y,\ell_{Y}\rangle \arrow[l, "\kappa_2"']
\end{tikzcd}\]

For functions $f:\langle X,\le^e_X,\ell_{X}\rangle\to\langle Z,\le^e_Z,\ell_{Z}\rangle$ and $g:\langle Y,\le^e_Y,\ell_{Y}\rangle\to\langle Z,\le^e_Z,\ell_{Z}\rangle$, there is a unique copairing function $[f,g]:\langle X,\le^e_X,\ell_{X}\rangle+\langle Y,\le^e_Y,\ell_{Y}\rangle\to\langle Z,\le^e_Z,\ell_{Z}\rangle$ with:

\[[f,g](w)=\begin{cases}
             f(x), & \mbox{if } w=(x,1); \\
             g(y), & \mbox{if } w=(y,2).
           \end{cases}\]
           
where $x\in\langle X,\le^e_X,\ell_{X}\rangle$ and $y\in\langle Y,\le^e_Y,\ell_{Y}\rangle$.

And the following equations hold:

\[[f,g]\circ\kappa_1=f\qquad [\kappa_1,\kappa_2]=\mathrm{id}_{\langle X,\le^e_X,\ell_{X}\rangle+\langle Y,\le^e_Y,\ell_{Y}\rangle}\]
\[[f,g]\circ\kappa_2=g\qquad h\circ[f,g]=[h\circ f,h\circ g]\]

where $h:\langle Z,\le^e_Z,\ell_{Z}\rangle\to\langle W,\le^e_W,\ell_{W}\rangle$.

For the above morphisms $f$ and $g$, there is a unique pairing morphism $[f,g]$ defined as above, making the following diagram commute:

\[\begin{tikzcd}[row sep=huge, column sep=huge]
\langle X,\le^e_X,\ell_{X}\rangle \arrow[r, "\kappa_1"]\arrow[rd,"f"'] & \langle X,\le^e_X,\ell_{X}\rangle+\langle Y,\le^e_Y,\ell_{Y}\rangle \arrow[dashed,d,"\ensuremath{[f,g]}"description] & \langle Y,\le^e_Y,\ell_{Y}\rangle \arrow[l, "\kappa_2"']\arrow[ld,"g"]\\
&\langle Z,\le^e_Z,\ell_{Z}\rangle&
\end{tikzcd}\]
\end{definition}

For functions $f:\langle X,\le^e_X,\ell_{X}\rangle\to\langle X',\le^e_{X'},\ell_{X'}\rangle$ and $g:\langle Y,\le^e_Y,\ell_{Y}\rangle\to\langle Y',\le^e_{Y'},\ell_{Y'}\rangle$, the coproduct of functions $f+ g$ is defined as:

\[(f+g)(w)=\begin{cases}
             (f(x),1), & \mbox{if } w=(x,1); \\
             (g(y),2), & \mbox{if } w=(y,2).
           \end{cases}\]
           
where $x\in\langle X,\le^e_X,\ell_{X}\rangle$ and $y\in\langle Y,\le^e_Y,\ell_{Y}\rangle$, such that the following equations hold:

\[f+ g=\langle \kappa_1\circ f,\kappa_2\circ g\rangle\qquad \mathrm{id}_{\langle X,\le^e_X,\ell_{X}\rangle}+\mathrm{id}_{\langle Y,\le^e_Y,\ell_{Y}\rangle}=\mathrm{id}_{\langle X,\le^e_X,\ell_{X}\rangle+\langle Y,\le^e_Y,\ell_{Y}\rangle}\]
\[(f\circ h)+(g\circ k)=(f+ g)\circ(h+ k)\]

Two functions $\langle X,\le^e_X,\ell_{X}\rangle\to\langle Z,\le^e_Z,\ell_{Z}\rangle$ and $\langle Y,\le^e_Y,\ell_{Y}\rangle\to\langle Z,\le^e_Z,\ell_{Z}\rangle$ can be composed into $\langle X,\le^e_X,\ell_{X}\rangle+\langle Y,\le^e_Y,\ell_{Y}\rangle\to\langle Z,\le^e_Z,\ell_{Z}\rangle$ by \cref{definitionofcoproduct} and $\langle X,\le^e_X,\ell_{X}\rangle+\langle Y,\le^e_Y,\ell_{Y}\rangle\to\langle Z,\le^e_Z,\ell_{Z}\rangle$ can be decomposed into $\langle X,\le^e_X,\ell_{X}\rangle\to\langle Z,\le^e_Z,\ell_{Z}\rangle$ and $\langle Y,\le^e_Y,\ell_{Y}\rangle\to\langle Z,\le^e_Z,\ell_{Z}\rangle$ by use of coprojections $\kappa_1$ and $\kappa_2$, which can be expressed by the following doubled line rule working in two directions:

\[\doublefrac{\langle X,\le^e_X,\ell_{X}\rangle\to\langle Z,\le^e_Z,\ell_{Z}\rangle\quad \langle Y,\le^e_Y,\ell_{Y}\rangle\to\langle Z,\le^e_Z,\ell_{Z}\rangle}{\langle X,\le^e_X,\ell_{X}\rangle+\langle Y,\le^e_Y,\ell_{Y}\rangle\to\langle Z,\le^e_Z,\ell_{Z}\rangle}\] 

The $n$-fold coproduct is defined as $n\cdot\langle X,\le^e_X,\ell_{X}\rangle=\langle X,\le^e_X,\ell_{X}\rangle+\cdots+\langle X,\le^e_X,\ell_{X}\rangle$ for a given $\langle X,\le^e_X,\ell_{X}\rangle$ and $n\in\mathbb{N}$. And the empty coproduct $0\cdot\langle X,\le^e_X,\ell_{X}\rangle$ is called an initial object.

\begin{definition}[Initial object]
An initial object in a category $\mathbb{C}$ is an object $0\in\mathbb{C}$, such that for each object $X\in\mathbb{C}$ there is a unique morphism $!_{X}:0\to X$ in $\mathbb{C}$.
\end{definition} 

\begin{proposition}[Initial object of $\mathbf{Trans}$]
$\mathbf{Trans}$ has an initial object.
\end{proposition}

\begin{proof}
The empty labelled occurrence multiset $0$ is initial.
\end{proof}

$\mathbf{Trans}$ has finite coproducts because it has binary coproducts and an initial object $0$. Finite coproducts can be generalized to set-indexed coproducts: given an index set $I$, a collection $(\langle X,\le^e_X,\ell_{X}\rangle)_{i\in I}$ of $I$-indexed objects has $I$-indexed coproducts $\langle X,\le^e_X,\ell_{X}\rangle=\coprod_{i\in I}\langle X_{i},\le^e_{X_{i}},\ell_{X_{i}}\rangle$ with coprojections $\kappa_{i}:\langle X_{i},\le^e_{X_{i}},\ell_{X_{i}}\rangle\to\langle X,\le^e_X,\ell_{X}\rangle$ for $i\in I$, which are universal as in \cref{definitionofcoproduct}: for an arbitrary object $\langle Y,\le^e_Y,\ell_{Y}\rangle$ and an $I$-indexed collection $f_{i}:\langle X_{i},\le^e_{X_{i}},\ell_{X_{i}}\rangle\to\langle Y,\le^e_Y,\ell_{Y}\rangle$ of morphisms there is a unique morphism $f=[f_{i}]:\langle X,\le^e_X,\ell_{X}\rangle\to\langle Y,\le^e_Y,\ell_{Y}\rangle$ with $f\circ\kappa_{i}=f_{i}$ for $i\in I$. In the category $\mathbf{Trans}$, such coproducts are disjoint unions and may be described as:

\[\coprod_{i\in I}\langle X_{i},\le^e_{X_{i}},\ell_{X_{i}}\rangle=\{(i,x):i\in I \wedge x\in\langle X_{i},\le^e_{X_{i}},\ell_{X_{i}}\rangle\}\] 

\runhead{Exponents}

For lmsets $\langle X,\le^e_X,\ell_{X}\rangle$ and $\langle Y,\le^e_Y,\ell_{Y}\rangle$, their exponents written $\langle Y,\le^e_Y,\ell_{Y}\rangle^{\langle X,\le^e_X,\ell_{X}\rangle}$, is defined as:

\[\langle Y,\le^e_Y,\ell_{Y}\rangle^{\langle X,\le^e_X,\ell_{X}\rangle}=\{f|f\mbox{ is a total function }\langle X,\le^e_X,\ell_{X}\rangle\to \langle Y,\le^e_Y,\ell_{Y}\rangle\}\]

The exponent operation $\langle Y,\le^e_Y,\ell_{Y}\rangle^{\langle X,\le^e_X,\ell_{X}\rangle}$ is a functor $\mathbf{Trans}^{\mathrm{op}}\times\mathbf{Trans}\to\mathbf{Trans}$, where $\mathbf{Trans}^{\mathrm{op}}$ is the opposite category of $\mathbf{Trans}$.

\begin{definition}[Exponent]\label{definitionofexponent}
Let us take two objects $\langle X,\le^e_X,\ell_X\rangle$ and $\langle Y,\le^e_Y,\ell_Y\rangle$ in $\mathbf{Trans}$. The exponent of these two objects is a new object $\langle Y,\le^e_Y,\ell_{Y}\rangle^{\langle X,\le^e_X,\ell_{X}\rangle}$ with an evaluation morphism:

\[\begin{tikzcd}[row sep=huge, column sep=huge]
\langle Y,\le^e_Y,\ell_{Y}\rangle^{\langle X,\le^e_X,\ell_{X}\rangle}\times\langle X,\le^e_X,\ell_{X}\rangle \arrow[r, "\ensuremath{\mathrm{ev}}"] & \langle Y,\le^e_Y,\ell_{Y}\rangle 
\end{tikzcd}\]

For each function $f:\langle Z,\le^e_Z,\ell_{Z}\rangle\times\langle X,\le^e_X,\ell_{X}\rangle\to\langle Y,\le^e_Y,\ell_{Y}\rangle$, there is a unique abstraction morphism $\Lambda(f):\langle X,\ell_{Z}\rangle\to\langle Y,\le^e_Y,\ell_{Y}\rangle^{\langle X,\le^e_X,\ell_{X}\rangle}$, such that the following equations hold:

\[\mathrm{ev}\circ(\Lambda(f)\times\mathrm{id}_{\langle X,\le^e_X,\ell_{X}\rangle})=f\qquad \Lambda(\mathrm{ev})=\mathrm{id}_{\langle Y,\le^e_Y,\ell_{Y}\rangle^{\langle X,\le^e_X,\ell_{X}\rangle}}\]
\[\Lambda(f)\circ h=\Lambda(f\circ(h\times\mathrm{id}_{\langle X,\le^e_X,\ell_{X}\rangle}))\]

For the above morphism $f$, there is a unique abstraction morphism $\Lambda(f)$ defined as above, making the following diagram commute:

\[\begin{tikzcd}[row sep=huge, column sep=huge]
\langle Y,\le^e_Y,\ell_{Y}\rangle^{\langle X,\le^e_X,\ell_{X}\rangle}\times\langle X,\le^e_X,\ell_{X}\rangle \arrow[r, "\ensuremath{\mathrm{ev}}"]  & \langle Y,\le^e_Y,\ell_{Y}\rangle \\
\langle Z,\le^e_Z,\ell_{Z}\rangle\times\langle X,\le^e_X,\ell_{X}\rangle \arrow[dashed,u,"\ensuremath{\Lambda(f)\times\mathrm{id}_{\langle X,\le^e_X,\ell_{X}\rangle}}"description] \arrow[ru,"f"']&
\end{tikzcd}\]
\end{definition}

For two maps $k:\langle X,\le^e_X,\ell_{X}\rangle\to\langle U,\le^e_U,\ell_{U}\rangle$ in $\mathbf{Trans}^{\mathrm{op}}$ and $h:\langle Y,\le^e_Y,\ell_{Y}\rangle\to\langle V,\le^e_V,\ell_{V}\rangle$ in $\mathbf{Trans}$, the exponent of functions $h^{k}$ is defined as:

\[h^{k}(f)=h\circ f\circ k\]

such that the following equations hold:

\[(\mathrm{id}^{\mathrm{id}}(f))=\mathrm{id}\circ f\circ\mathrm{id}=f\]
\[\begin{aligned}
(h_{2}^{k_{2}}\circ h_{1}^{k_{1}})(f)&=(h_{2}^{k_{2}}(h_{1}^{k_{1}}(f))\\
&=h_{2}^{k_{2}}(h_{1}\circ f\circ k_{1})\\
&=h_{2}\circ h_{1}\circ f\circ k_{1}\circ k_{2}\\
&=(h_{2}\circ h_{1})^{k_{1}\circ k_{2}}(f)\\
&=(h_{2}\circ_{\mathbb{C}} h_{1})^{k_{1}\circ_{\mathbb{C}^{\mathrm{op}}} k_{2}}(f)
\end{aligned}\]

The exponent, also called currying, can be expressed by the following doubled line rule:

\[\doublefrac{\langle Z,\le^e_Z,\ell_{Z}\rangle\times\langle X,\le^e_X,\ell_{X}\rangle\to\langle Y,\le^e_Y,\ell_{Y}\rangle}{\langle Z,\le^e_Z,\ell_{Z}\rangle\to\langle Y,\le^e_Y,\ell_{Y}\rangle^{\langle X,\le^e_X,\ell_{X}\rangle}}\] 

A cartesian closed category, denoted $\mathbf{CCC}$, is a category with finite products and exponents. And a bicartesian closed category, denoted $\mathbf{BiCCC}$, is a $\mathbf{CCC}$ with finite coproducts, and $\mathbf{Trans}$ is a $\mathbf{BiCCC}$.

\runhead{Powersets}

For an lmset $\langle X,\le^e_X,\ell_{X}\rangle$ we write the powerset of $\langle X,\le^e_X,\ell_{X}\rangle$ as $\mathcal{P}(\langle X,\le^e_X,\ell_{X}\rangle)=\{\langle U,\le^e_U,\ell_{U}\rangle|\langle U,\le^e_U,\ell_{U}\rangle\subseteq \langle X,\le^e_X,\ell_{X}\rangle\}$ for the set of all subsets of $\langle X,\le^e_X,\ell_{X}\rangle$. The powerset $\mathcal{P}(\langle X,\le^e_X,\ell_{X}\rangle)$ is naturally ordered by set inclusion $\subseteq$ and yields a poset $\langle \mathcal{P}(\langle X,\le^e_X,\ell_{X}\rangle),\subseteq\rangle$. By defining meets by intersection $\bigcap_{i\in I}\langle U_{i},\le^e_{U_{i}},\ell_{U_{i}}\rangle=\{x\in\langle X,\le^e_X,\ell_{X}\rangle|\forall i\in I.x\in\langle U_{i},\le^e_{U_{i}},\ell_{U_{i}}\rangle\}$, joins by unions $\bigcup_{i\in I}\langle U_{i},\le^e_{U_{i}},\ell_{U_{i}}\rangle=\{x\in\langle X,\le^e_X,\ell_{X}\rangle|\exists i\in I.x\in\langle U_{i},\le^e_{U_{i}},\ell_{U_{i}}\rangle\}$ and negation by complement $\neg\langle U,\le^e_U,\ell_{U}\rangle=\{x\in\langle X,\le^e_X,\ell_{X}\rangle|x\notin\langle U,\le^e_U,\ell_{U}\rangle\}$, and truth predicate $\top_{\langle X,\le^e_X,\ell_{X}\rangle}=(\langle X,\le^e_X,\ell_{X}\rangle\subseteq\langle X,\le^e_X,\ell_{X}\rangle)$ and falsity predicate $\bot_{\langle X,\le^e_X,\ell_{X}\rangle}=(\emptyset\not\subseteq\langle X,\le^e_X,\ell_{X}\rangle)$, the poset $\langle \mathcal{P}(\langle X,\le^e_X,\ell_{X}\rangle),\subseteq\rangle$ forms a complete Boolean algebra, and thus $\{\top,\bot\}$ of Booleans is the powerset $\mathcal{P}(1)$ of the final object $1$.

A relation $R$ on lmsets $\langle X,\le^e_X,\ell_{X}\rangle$ and $\langle Y,\le^e_Y,\ell_{Y}\rangle$ is a subset $R\subseteq\langle X,\le^e_X,\ell_{X}\rangle\times\langle Y,\le^e_Y,\ell_{Y}\rangle$, i.e, $R\in\mathcal{P}(\langle X,\le^e_X,\ell_{X}\rangle\times\langle Y,\le^e_Y,\ell_{Y}\rangle)$ is a special case of predicates. Like predicates, $\langle \mathcal{P}(\langle X,\le^e_X,\ell_{X}\rangle\times\langle Y,\le^e_Y,\ell_{Y}\rangle),\subseteq\rangle$ forms again a complete Boolean algebra. Ordinarily, relations are equipped with reversal $R^{\dagger}$ for $R$ and composition $\circ$.

For endo-relations $R\subseteq\langle X,\le^e_X,\ell_{X}\rangle\times\langle X,\le^e_X,\ell_{X}\rangle$ on some lmset $\langle X,\le^e_X,\ell_{X}\rangle$, the equality relation $\mathrm{Eq}(\langle X,\le^e_X,\ell_{X}\rangle)=\{(x,y)\in \langle X,\le^e_X,\ell_{X}\rangle\times\langle X,\le^e_X,\ell_{X}\rangle|x=y\}=\{(x,x)|x\in \langle X,\le^e_X,\ell_{X}\rangle\}$. The category $\mathbf{LMSetsRel}$ contains lmsets as objects, relations as morphisms and equality relations as identity morphisms.

The powerset operation $\langle X,\le^e_X,\ell_{X}\rangle\mapsto\mathcal{P}({\langle X,\le^e_X,\ell_{X}\rangle})$ is also functorial. For a function $f:\langle X,\le^e_X,\ell_{X}\rangle\to \langle Y,\le^e_Y,\ell_{Y}\rangle$, there is a function $\mathcal{P}(f):\mathcal{P}(\langle X,\le^e_X,\ell_{X}\rangle)\to\mathcal{P}(\langle Y,\le^e_Y,\ell_{Y}\rangle)$, there are several ways to give this functor:

\begin{enumerate}
  \item The direct image way: for $\langle U,\le^e_U,\ell_{U}\rangle\subseteq \langle X,\le^e_X,\ell_{X}\rangle$,
  \[\begin{aligned}
  \mathcal{P}(f)(\langle U,\le^e_U,\ell_{U}\rangle)&=\{f(x)|x\in\langle U,\le^e_U,\ell_{U}\rangle\}\\
  &=\{y\in\langle Y,\le^e_Y,\ell_{Y}\rangle|\exists x\in\langle X,\le^e_X,\ell_{X}\rangle.f(x)=y\wedge x\in \langle U,\le^e_U,\ell_{U}\rangle\}
  \end{aligned}\]
  The direct image can also be described by $f[\langle U,\le^e_U,\ell_{U}\rangle]$ or $\coprod_{f}(\langle U,\le^e_U,\ell_{U}\rangle)$. The direct image way describe the powerset as a functor $\mathcal{P}:\mathbf{Trans}\to\mathbf{Trans}$. For finite powerset $\mathcal{P}_{\mathrm{fin}}(\langle X,\le^e_X,\ell_{X}\rangle)=\{\langle U,\le^e_U,\ell_{U}\rangle\in\mathcal{P}(\langle X,\le^e_X,\ell_{X}\rangle)|\langle U,\le^e_U,\ell_{U}\rangle\mbox{ is finite}\}$, which also forms a functor $\mathbf{Trans}\to\mathbf{Trans}$.
  \item The inverse image way: the inverse image way describes the powerset as a functor $\mathcal{P}:\mathbf{Trans}^{\mathrm{op}}\to\mathbf{Trans}$. For a function $f:\langle X,\le^e_X,\ell_{X}\rangle\to \langle Y,\le^e_Y,\ell_{Y}\rangle$, there is a function $f^{-1}:\mathcal{P}(\langle Y,\le^e_Y,\ell_{Y}\rangle)\to\mathcal{P}(\langle X,\le^e_X,\ell_{X}\rangle)$, for $\langle U,\le^e_U,\ell_{U}\rangle\subseteq \langle X,\le^e_X,\ell_{X}\rangle$,
  \[f^{-1}(\langle U,\le^e_U,\ell_{U}\rangle)=\{x|f(x)\in\langle U,\le^e_U,\ell_{U}\rangle\}\] 
\end{enumerate}

The correspondence of the direct image and inverse image can be reflected by the following rule:

\[\doublefrac{\coprod_{f}(\langle U,\le^e_U,\ell_{U}\rangle)\subseteq\langle V,\le^e_V,\ell_{V}\rangle}{\langle U,\le^e_U,\ell_{U}\rangle\subseteq f^{-1}(\langle V,\le^e_V,\ell_{V}\rangle)}\]

And there is a correspondence for powerset, which can be expressed by the following rule:

\[\doublefrac{\langle X,\le^e_X,\ell_{X}\rangle\to\mathcal{P}(\langle Y,\le^e_Y,\ell_{Y}\rangle)}{\mbox{relations}\subseteq \langle Y,\le^e_Y,\ell_{Y}\rangle\times\langle X,\le^e_X,\ell_{X}\rangle}\]

\runhead{Injections and Surjections}

For lmsets $\langle X,\le^e_X,\ell_{X}\rangle$ and $\langle Y,\le^e_Y,\ell_{Y}\rangle$, an injection (a monomorphism (mono) or a monic map) is a map such that if $f(x)=f(x')$ then $x=x'$ for $x,x'\in\langle X,\le^e_X,\ell_{X}\rangle$. For an injection, in case $\langle X,\le^e_X,\ell_{X}\rangle\subseteq\langle Y,\le^e_Y,\ell_{Y}\rangle$, we write $f:\langle X,\le^e_X,\ell_{X}\rangle\hookrightarrow\langle Y,\le^e_Y,\ell_{Y}\rangle$ or $f:\langle X,\le^e_X,\ell_{X}\rangle\rightarrowtail\langle Y,\le^e_Y,\ell_{Y}\rangle$. A surjection (an epimorphism (epi) or epic map) is a map written as $f:\langle X,\le^e_X,\ell_{X}\rangle\twoheadrightarrow\langle Y,\le^e_Y,\ell_{Y}\rangle$ such that for each $y\in\langle Y,\le^e_Y,\ell_{Y}\rangle$ there is an $x\in \langle X,\le^e_X,\ell_{X}\rangle$ with $f(x)=y$.

Every surjection $f:\langle X,\le^e_X,\ell_{X}\rangle\twoheadrightarrow\langle Y,\le^e_Y,\ell_{Y}\rangle$ has a section (splitting), a function $s:\langle Y,\le^e_Y,\ell_{Y}\rangle\to\langle X,\le^e_X,\ell_{X}\rangle$ in the reverse direction with $f\circ s=\mathrm{id}$. This $s$ thus chooses an element $s(y)$ in the non-empty set $f^{-1}(y)=\{x\in \langle X,\le^e_X,\ell_{X}\rangle|f(x)=y\}$, and such a section is an injection.

Also every injection $f:\langle X,\le^e_X,\ell_{X}\rangle\rightarrowtail\langle Y,\le^e_Y,\ell_{Y}\rangle$ with $\langle X,\le^e_X,\ell_{X}\rangle\neq\emptyset$, there is a function $g:\langle Y,\le^e_Y,\ell_{Y}\rangle\to\langle X,\le^e_X,\ell_{X}\rangle$ with $g\circ f=\mathrm{id}$, where $x_0\in \langle X,\le^e_X,\ell_{X}\rangle$:

\[g(y)=\begin{cases}
         x, & \mbox{if there is an element }x \mbox{ with } f(x)=y;\\
         x_0, & \mbox{otherwise}.
       \end{cases}\]

\begin{lemma}[Injection and surjection]
Let $F:\mathbf{Trans}\to\mathbf{Trans}$ be an arbitrary functor.

\begin{enumerate}
  \item If $f:\langle X,\le^e_X,\ell_{X}\rangle\to\langle Y,\le^e_Y,\ell_{Y}\rangle$ is surjective, then so is $F(f):F(\langle X,\le^e_X,\ell_{X}\rangle)\to F(\langle Y,\le^e_Y,\ell_{Y}\rangle)$.
  \item If $f:\langle X,\le^e_X,\ell_{X}\rangle\to\langle Y,\le^e_Y,\ell_{Y}\rangle$ is injective with $\langle X,\le^e_X,\ell_{X}\rangle\neq\emptyset$, then so is $F(f):F(\langle X,\le^e_X,\ell_{X}\rangle)\to F(\langle Y,\le^e_Y,\ell_{Y}\rangle)$.
\end{enumerate}
\end{lemma}

\begin{definition}[Collections of polynomial functors]
We define three collections of functors $\mathbf{Trans}\to\mathbf{Trans}$, $\mathbf{SPF}$, $\mathbf{EPF}$ and $\mathbf{KPF}$, for simple, exponent and Kripke polynomial functors:

\begin{enumerate}
  \item $\mathbf{SPF}$ is the least class of functors $\mathbf{Trans}\to\mathbf{Trans}$ satisfying:
  \begin{enumerate}
    \item The identity functor $\mathbf{Trans}\to\mathbf{Trans}$ is in $\mathbf{SPF}$.
    \item For each lmset $\langle A,\le^e_A,\ell_{A}\rangle$, the constant functor $A:\mathbf{Trans}\to\mathbf{Trans}$ is in $\mathbf{SPF}$, which maps every lmset $\langle X,\le^e_X,\ell_{X}\rangle$ to $\langle A,\le^e_A,\ell_{A}\rangle$ and every function $f$ to the identity $\mathrm{id}_{\langle A,\le^e_A,\ell_{A}\rangle}$ on $\langle A,\le^e_A,\ell_{A}\rangle$.
    \item If functors $F,G\in\mathbf{SPF}$, then $F\times G\in\mathbf{SPF}$, as $\langle X,\le^e_X,\ell_{X}\rangle\mapsto F(\langle X,\le^e_X,\ell_{X}\rangle)\times F(\langle X,\le^e_X,\ell_{X}\rangle)$ and for functions $f\mapsto F(f)\times G(f)$.
    \item For non-empty index set $I$, an $I$-indexed collection of functors $F_{i}\in\mathbf{SPF}$ for $i\in I$, then $\coprod_{i\in I}F_{i}\in\mathbf{SPF}$, as $\langle X,\le^e_X,\ell_{X}\rangle\mapsto\coprod_{i\in I}F_{i}(\langle X,\le^e_X,\ell_{X}\rangle)$ with $\coprod_{i\in I}F_{i}(\langle X,\le^e_X,\ell_{X}\rangle)=\{(i,u)|i\in I\wedge u\in F_{i}(\langle X,\le^e_X,\ell_{X}\rangle)\}$ and for a function $f:\langle X,\le^e_X,\ell_{X}\rangle\to\langle Y,\le^e_Y,\ell_{Y}\rangle$ to the mapping $(i,u)\mapsto(i,F_{i}(f)(u))$.
  \end{enumerate}
  \item $\mathbf{EPF}\supset\mathbf{SPF}$, and additionally, for each lmset $\langle A,\le^e_A,\ell_{A}\rangle$, if $F\in\mathbf{EPF}$, then so is the constant exponent $F^{\langle A,\le^e_A,\ell_{A}\rangle}$ as $\langle X,\le^e_X,\ell_{X}\rangle\mapsto F(\langle X,\le^e_X,\ell_{X}\rangle)^{\langle A,\le^e_A,\ell_{A}\rangle}$. It sends a function $f:\langle X,\le^e_X,\ell_{X}\rangle\to\langle Y,\le^e_Y,\ell_{Y}\rangle$ to the function $F(f)^{\langle A,\le^e_A,\ell_{A}\rangle}=F(f)^{\mathrm{id}_{\langle A,\le^e_A,\ell_{A}\rangle}}$, which maps $h:\langle A,\le^e_A,\ell_{A}\rangle\to F(\langle X,\le^e_X,\ell_{X}\rangle)$ to $F(f)\circ h:\langle A,\le^e_A,\ell_{A}\rangle\to F(\langle Y,\le^e_Y,\ell_{Y}\rangle)$.
  \item $\mathbf{KPF}\supset\mathbf{EPF}$, and additionally, 
  \begin{enumerate}
    \item If $F\in\mathbf{KPF}$, then so is the powerset $\mathcal{P}(F)\in\mathbf{KPF}$, as $\langle X,\le^e_X,\ell_{X}\rangle\mapsto\mathcal{P}(F(\langle X,\le^e_X,\ell_{X}\rangle))$ and functions $f\mapsto\mathcal{P}(F(f))$. 
    \item for each lmset $\langle A,\le^e_A,\ell_{A}\rangle$, if $F\in\mathbf{KPF}$, then so is the powerset exponent $F^{\mathcal{P}(\langle A,\le^e_A,\ell_{A}\rangle)}$ as $\langle X,\le^e_X,\ell_{X}\rangle\mapsto F(\langle X,\le^e_X,\ell_{X}\rangle)^{\mathcal{P}(\langle A,\le^e_A,\ell_{A}\rangle})$. It sends a function $f:\langle X,\le^e_X,\ell_{X}\rangle\to\langle Y,\le^e_Y,\ell_{Y}\rangle$ to the function $F(f)^{\mathcal{P}(\langle A,\le^e_A,\ell_{A}\rangle)}=F(f)^{\mathcal{P}(\mathrm{id}_{\langle A,\le^e_A,\ell_{A}\rangle})}$, which maps $h:\mathcal{P}(\langle A,\le^e_A,\ell_{A}\rangle)\to F(\langle X,\le^e_X,\ell_{X}\rangle)$ to $F(f)\circ h:\mathcal{P}(\langle A,\le^e_A,\ell_{A}\rangle)\to F(\langle Y,\le^e_Y,\ell_{Y}\rangle)$.
  \end{enumerate}
   If the powersets $\mathcal{P}$ in $F$ are $\mathcal{P}_{\mathrm{fin}}$, then $F$ is called a finite $\mathbf{KPF}$.
\end{enumerate}
\end{definition}

The list functor $F^{\star}$ is given as infinite coproduct of products:

\[F^{\star}=\coprod_{n\in\mathbb{N}}F^{n}\mbox{ where }F^{n}=F\times F^{n-1}\]

If $F$ is the identity functor, then $F^{\star}$ maps an lmset $\langle X,\le^e_X,\ell_{X}\rangle$ to the set of lists:

\[\langle X,\le^e_X,\ell_{X}\rangle^{\star}=1+\langle X,\le^e_X,\ell_{X}\rangle+(\langle X,\le^e_X,\ell_{X}\rangle\times\langle X,\le^e_X,\ell_{X}\rangle)+(\langle X,\le^e_X,\ell_{X}\rangle\times\langle X,\le^e_X,\ell_{X}\rangle\times\langle X,\le^e_X,\ell_{X}\rangle)+\cdots\]

\begin{definition}[Final coalgebra]
Let $\mathbb{C}$ be an arbitrary category with an endofunctor $F:\mathbb{C}\to\mathbb{C}$. A final $F$-coalgebra is simply a final object in the associated category $\mathbf{CoAlg}(F)$ of $F$-coalgebras. Thus, it is a coalgebra $\zeta:Z\to F(Z)$ such that for any coalgebra $c:X\to F(X)$, there is a unique homomorphism $beh_{c}:X \to Z$ of coalgebras, making the following diagram commute:

\begin{center}
\begin{tikzcd}[row sep=huge, column sep=huge]
F(X) \arrow[dashed,r,"\ensuremath{F(beh_{c})}"] & F(Z)  \\
X \arrow[dashed,r,"\ensuremath{beh_{c}}"'] \arrow[u,"c"] & Z \arrow[u,"\ensuremath{\zeta}"']
\end{tikzcd}
\end{center}
\end{definition}

\begin{lemma}
A final coalgebra, if it exists, is determined up to isomorphism.
\end{lemma}

\begin{lemma}
A final coalgeba $\zeta:Z\to F(Z)$ is necessarily an isomorphism $\zeta:Z\xrightarrow{\cong}F(Z)$.
\end{lemma}

\begin{theorem}
Each finite Kripke polynomail functor $\mathbf{Trans}\to\mathbf{Trans}$ has a final coalgebra.
\end{theorem}

\begin{definition}[Initial algebra]
Let $\mathbb{C}$ be an arbitrary category with an endofunctor $F:\mathbb{C}\to\mathbb{C}$. An initial $F$-algebra is simply an initial object in the associated category $\mathbf{Alg}(F)$ of $F$-algebras. Thus, it is an algebra $\alpha:F(A)\to A$ such that for any algebra $b:F(X)\to X$, there is a unique homomorphism $int_{b}:\alpha\to b$ of algebras, making the following diagram commute:

\begin{center}
\begin{tikzcd}[row sep=huge, column sep=huge]
F(A) \arrow[dashed,r,"\ensuremath{F(int_{b})}"] \arrow[d,"\ensuremath{\alpha}"'] & F(X) \arrow[d,"\ensuremath{b}"] \\
A \arrow[dashed,r,"\ensuremath{int_{b}}"']  & X 
\end{tikzcd}
\end{center}
\end{definition}

\begin{lemma}
An initial algebra $F(A)\to A$ of a functor $F:\mathbb{C}\to\mathbb{C}$ is an isomorphism $F(A)\xrightarrow{\cong}A$ in $\mathbb{C}$.
\end{lemma}

\begin{proposition}
An arbitrary initial algebra $\alpha:F(\langle A,\le^e_A,\ell_{A}\rangle)\xrightarrow{\cong}\langle A,\le^e_A,\ell_{A}\rangle$ satisfying the following recursion property: for each map $h:F(\langle X,\le^e_X,\ell_{X}\rangle\times\langle A,\le^e_A,\ell_{A}\rangle)\to \langle X,\le^e_X,\ell_{X}\rangle$, there is a unique map $f:\langle A,\le^e_A,\ell_{A}\rangle\to \langle X,\le^e_X,\ell_{X}\rangle$ making the following diagram commute:

\begin{center}
\begin{tikzcd}[row sep=huge, column sep=huge]
F(\langle A,\le^e_A,\ell_{A}\rangle) \arrow[dashed,r,"\ensuremath{F(\langle f,\mathrm{id}\rangle)}"] \arrow[d,"\ensuremath{\alpha}"',"\cong"] & F(\langle X,\le^e_X,\ell_{X}\rangle\times\langle A,\le^e_A,\ell_{A}\rangle) \arrow[d,"\ensuremath{h}"] \\
\langle A,\le^e_A,\ell_{A}\rangle \arrow[dashed,r,"\ensuremath{f}"']  & \langle X,\le^e_X,\ell_{X}\rangle 
\end{tikzcd}
\end{center}
\end{proposition}

\subsection{Algebra of Transformers $AT$}

\subsubsection{Step Bisimulation}

\begin{definition}[Relation lifting]
Let $F:\mathbf{Trans}\to\mathbf{Trans}$ be a polynomial functor, and let $\langle X,\le^e_X,\ell_{X}\rangle$, $\langle Y,\le^e_Y,\ell_{Y}\rangle$ be arbitrary lmsets. The mapping $\mathrm{Rel}(F)$ sends a relation $R\subseteq\langle X,\le^e_X,\ell_{X}\rangle\times\langle Y,\le^e_Y,\ell_{Y}\rangle$ to a lifted relation $\mathrm{Rel}(F)(R)\subseteq F(\langle X,\le^e_X,\ell_{X}\rangle)\times F(\langle Y,\le^e_Y,\ell_{Y}\rangle)$, which is defined by induction on the structure of the functor $F$ as follows:

\begin{enumerate}
  \item If $F$ is the identity functor, then 
  \[\mathrm{Rel}(F)(R)=R\]
  \item If $F$ is a constant functor $\langle X,\le^e_X,\ell_{X}\rangle\mapsto\mathcal{P}(\langle A,\le^e_A,\ell_{A}\rangle)$, then
  \[\mathrm{Rel}(F)(R)=\{(U,U)|U\in\mathcal{P}(\langle A,\le^e_A,\ell_{A}\rangle)\}\]
  \item If $F$ is a product $F_{1}\times F_{2}$, then
  \[\mathrm{Rel}(F)(R)=\{((u_{1},u_{2}),(v_{1},v_{2}))|\mathrm{Rel}(F_{1})(R)(u_{1},v_{1})\wedge\mathrm{Rel}(F_{2})(R)(u_{2},v_{2})\}\]
  \item If $F$ is a set-indexed coproduct $\coprod_{i\in I}F_{i}$, then
  \[\mathrm{Rel}(F)(R)=\bigcup_{i\in I}\{(\kappa_{i}(u),\kappa_{i}(v))|\mathrm{Rel}(F_{i})(R)(u,v)\}\]
  \item If $F$ is an exponent $G^{\mathcal{P}(\langle A,\le^e_A,\ell_{A}\rangle)}$, then
  \[\mathrm{Rel}(F)(R)=\{(f,g)|\forall U\in\mathcal{P}(\langle A,\le^e_A,\ell_{A}\rangle).\mathrm{Rel}(G)(R)(f(U),g(U))\}\]
  \item If $F$ is a powerset $\mathcal{P}(G)$, then
  \[\mathrm{Rel}(F)(R)=\{(U,V)|\forall u\in U.\exists v\in V.\mathrm{Rel}(G)(R)(u,v)\wedge \forall v\in V.\exists u\in U.\mathrm{Rel}(G)(R)(u,v)\}\]
\end{enumerate}
\end{definition}

\begin{definition}[Step bisimulation]
Let $F:\mathbf{Trans}\to\mathbf{Trans}$ be a polynomial functor. A step bisimulation for coalgebras $c:\langle X,\le^e_X,\ell_{X}\rangle\to F(\langle X,\le^e_X,\ell_{X}\rangle)$ and $d:\langle Y,\le^e_Y,\ell_{Y}\rangle\to F(\langle Y,\le^e_Y,\ell_{Y}\rangle)$ is a relation $R\subseteq\langle X,\le^e_X,\ell_{X}\rangle\times\langle Y,\le^e_Y,\ell_{Y}\rangle$ satisfying:

\[(x,y)\in R\Rightarrow(c(x),d(y))\in\mathrm{Rel}(F)(R)\]

A step bisimulation equivalence is a step bisimulation on a single coalgebra which is an equivalence relation.
\end{definition}

\begin{definition}[Step bisimilarity]
Let $c:\langle X,\le^e_X,\ell_{X}\rangle\to F(\langle X,\le^e_X,\ell_{X}\rangle)$ and $d:\langle Y,\le^e_Y,\ell_{Y}\rangle\to F(\langle Y,\le^e_Y,\ell_{Y}\rangle)$ be two coalgebras of a polynomial functor $F$. The step bisimilarity relation $\underline{\leftrightarrow}$ is the union of all step bisimulations:

\[x\underline{\leftrightarrow}y\Leftrightarrow\exists R\subseteq \langle X,\le^e_X,\ell_{X}\rangle\times\langle Y,\le^e_Y,\ell_{Y}\rangle.R\mbox{ is a step bisimulation for }c\mbox{ and }d,\mbox{ and }R(x,y)\] 
\end{definition}

\begin{theorem}\label{tbisimulation1}
Let $F:\mathbf{Trans}\to\mathbf{Trans}$ be a finite polynomial functor which has a final coalgebra $\zeta:\langle Z,\le^e_Z,\ell_{Z}\rangle\xrightarrow{\cong}F(\langle Z,\le^e_Z,\ell_{Z}\rangle)$. Let $c:\langle X,\le^e_X,\ell_{X}\rangle\to F(\langle X,\le^e_X,\ell_{X}\rangle)$ and $d:\langle Y,\le^e_Y,\ell_{Y}\rangle\to F(\langle Y,\le^e_Y,\ell_{Y}\rangle)$ be arbitrary coalgebras, with associated behaviour homomorphism $\mathrm{beh}_{c}:\langle X,\le^e_X,\ell_{X}\rangle\to\langle Z,\le^e_Z,\ell_{Z}\rangle$ and $\mathrm{beh}_{d}:\langle Y,\le^e_Y,\ell_{Y}\rangle\to\langle Z,\le^e_Z,\ell_{Z}\rangle$. Two states $x\in \langle X,\le^e_X,\ell_{X}\rangle$ and $y\in\langle Y,\le^e_Y,\ell_{Y}\rangle$ are then step bisimilar if and only if they have the same truly concurrent behaviour:

\[x\prescript{}{c}{\underline{\leftrightarrow}}_{d}y\Leftrightarrow\mathrm{beh}_{c}(x)=\mathrm{beh}_{d}(y)\]
\end{theorem}

\begin{definition}[Congruence]
Let $F:\mathbf{Trans}\to\mathbf{Trans}$ be a polynomial functor. A congruence for algebras $a:F(\langle X,\le^e_X,\ell_{X}\rangle)\to\langle X,\le^e_X,\ell_{X}\rangle$ and $b:F(\langle Y,\le^e_Y,\ell_{Y}\rangle)\to\langle Y,\le^e_Y,\ell_{Y}\rangle$ is a relation $R\subseteq\langle X,\le^e_X,\ell_{X}\rangle\times\langle Y,\le^e_Y,\ell_{Y}\rangle$ satisfying:

\[(u,v)\in\mathrm{Rel}(F)(R)\Rightarrow(a(u),b(v))\in R\]

A congruence equivalence is a congruence which is an equivalence relation.
\end{definition}

\begin{theorem}\label{tcongruence1}
Assume two algebras $a:F(\langle X,\le^e_X,\ell_{X}\rangle)\to\langle X,\le^e_X,\ell_{X}\rangle$ and $b:F(\langle Y,\le^e_Y,\ell_{Y}\rangle)\to\langle Y,\le^e_Y,\ell_{Y}\rangle$ of a Kripke polynomial functor $F$. A relation $R\subseteq \langle X,\le^e_X,\ell_{X}\rangle\times\langle Y,\le^e_Y,\ell_{Y}\rangle$ is a congruence if and only if it carries a unique algebra $F(R)\to R$ itself, making $\langle r_1,r_2\rangle:R\rightarrowtail \langle X,\le^e_X,\ell_{X}\rangle\times\langle Y,\le^e_Y,\ell_{Y}\rangle$ of the inclusion homomorphisms of algebras as:

\[\begin{tikzcd}[row sep=huge, column sep=huge]
F(\langle X,\le^e_X,\ell_{X}\rangle)  \arrow[d,"a"']& F(R) \arrow[l, "\ensuremath{F(r_1)}"']\arrow[r, "\ensuremath{F(r_2)}"] \arrow[d,"\ensuremath{\langle r_1,r_2\rangle}"']& F(\langle Y,\le^e_Y,\ell_{Y}\rangle)\arrow[d,"b"]\\
\langle X,\le^e_X,\ell_{X}\rangle & R \arrow[l, "\ensuremath{r_1}"]\arrow[r, "\ensuremath{r_2}"'] &\langle Y,\le^e_Y,\ell_{Y}\rangle
\end{tikzcd}\]
\end{theorem} 

\subsubsection{Algebra of Transformers $AT$}

Since the functor $\langle X,\le^e_X,\ell_{X}\rangle\mapsto\mathcal{P}_{\mathrm{fin}}(\langle X,\le^e_X,\ell_{X}\rangle)^{\mathcal{P}_{\mathrm{fin}}(\langle A,\le^e_A,\ell_{A}\rangle)}$ is a finite Kripke polynomial functor which has a final coalgebra:

\[Z_{\mathcal{P}_{\mathrm{fin}}(\langle A,\le^e_A,\ell_{A}\rangle)}\xlongrightarrow[\cong]{\zeta_{\mathcal{P}_{\mathrm{fin}}(\langle A,\le^e_A,\ell_{A}\rangle)}}\mathcal{P}_{\mathrm{fin}}(Z_{\mathcal{P}_{\mathrm{fin}}(\langle A,\le^e_A,\ell_{A}\rangle)})^{\mathcal{P}_{\mathrm{fin}}(\langle A,\le^e_A,\ell_{A}\rangle)}\]

The elements of $Z_{\mathcal{P}_{\mathrm{fin}}(\langle A,\le^e_A,\ell_{A}\rangle)}$ of the final coalgebra is called truly concurrent processes.

The null process is a trivial process which does nothing:

\[\mathbf{0}\triangleq\zeta_{\mathcal{P}_{\mathrm{fin}}(\langle A,\le^e_A,\ell_{A}\rangle)}^{-1}(\lambda U\in\mathcal{P}_{\mathrm{fin}}(\langle A,\le^e_A,\ell_{A}\rangle).\emptyset)\]

Given two processes $z_1,z_2\in Z_{\mathcal{P}_{\mathrm{fin}}(\langle A,\le^e_A,\ell_{A}\rangle)}$, the sum process $z_1+z_2\in Z_{\mathcal{P}_{\mathrm{fin}}(\langle A,\le^e_A,\ell_{A}\rangle)}$ is defined as follows:

\[z_1+z_2\triangleq\zeta_{\mathcal{P}_{\mathrm{fin}}(\langle A,\le^e_A,\ell_{A}\rangle)}^{-1}(\lambda U\in \mathcal{P}_{\mathrm{fin}}(\langle A,\le^e_A,\ell_{A}\rangle).\zeta_{\mathcal{P}_{\mathrm{fin}}(\langle A,\le^e_A,\ell_{A}\rangle)}(z_1)(U)\cup\zeta_{\mathcal{P}_{\mathrm{fin}}(\langle A,\le^e_A,\ell_{A}\rangle)}(z_2)(U))\]

Then, we have:

\[\begin{aligned}
(z_1+z_2)\xrightarrow{U}w &\Leftrightarrow w\in\zeta_{\mathcal{P}_{\mathrm{fin}}(\langle A,\le^e_A,\ell_{A}\rangle)}(z_1+z_2)(U)\\
&\Leftrightarrow w\in\zeta_{\mathcal{P}_{\mathrm{fin}}(\langle A,\le^e_A,\ell_{A}\rangle)}(z_1)(U)\cup\zeta_{\mathcal{P}_{\mathrm{fin}}(\langle A,\le^e_A,\ell_{A}\rangle)}(z_2)(U)\\
&\Leftrightarrow w\in\zeta_{\mathcal{P}_{\mathrm{fin}}(\langle A,\le^e_A,\ell_{A}\rangle)}(z_1)(U)\mbox{ or }w\in\zeta_{\mathcal{P}_{\mathrm{fin}}(\langle A,\le^e_A,\ell_{A}\rangle)}(z_2)(U)\\
&\Leftrightarrow z_1\xrightarrow{U}w\mbox{ or }z_2\xrightarrow{U}w
\end{aligned}\]

which can be described by the following two rules:

\[\frac{z_1\xrightarrow{U}w}{z_1+z_2\xrightarrow{U}w}\qquad\frac{z_2\xrightarrow{U}w}{z_1+z_2\xrightarrow{U}w}\]

Given a process $z\in Z_{\mathcal{P}_{\mathrm{fin}}(\langle A,\le^e_A,\ell_{A}\rangle)}$ and a step $V\in\mathcal{P}_{\mathrm{fin}}(\langle A,\le^e_A,\ell_{A}\rangle)$, the prefixing process $V.z$ is defined as follows:

\[V.z\triangleq\zeta_{\mathcal{P}_{\mathrm{fin}}(\langle A,\le^e_A,\ell_{A}\rangle)}^{-1}(\lambda U\in \mathcal{P}_{\mathrm{fin}}(\langle A,\le^e_A,\ell_{A}\rangle)).\{z|U\cong V\}\]

Then, we have:

\[\begin{aligned}
(V.z)\xrightarrow{U}w &\Leftrightarrow w\in\zeta_{\mathcal{P}_{\mathrm{fin}}(\langle A,\le^e_A,\ell_{A}\rangle)}(V.z)(U)\\
&\Leftrightarrow w\in\{z|U\cong V\}\\
&\Leftrightarrow U\cong V\wedge w=z
\end{aligned}\]

which can be described by the following rule:

\[\frac{}{V.z\xrightarrow{V}z}\]

Given two processes $z_1,z_2\in Z_{\mathcal{P}_{\mathrm{fin}}(\langle A,\le^e_A,\ell_{A}\rangle)}$, the parallel process $z_1\oplus z_2\in Z_{\mathcal{P}_{\mathrm{fin}}(\langle A,\le^e_A,\ell_{A}\rangle)}$ is defined as follows:

\[z_1\oplus z_2\triangleq\zeta_{\mathcal{P}_{\mathrm{fin}}(\langle A,\le^e_A,\ell_{A}\rangle)}^{-1}(\lambda U,V\in \mathcal{P}_{\mathrm{fin}}(\langle A,\le^e_A,\ell_{A}\rangle).\zeta_{\mathcal{P}_{\mathrm{fin}}(\langle A,\le^e_A,\ell_{A}\rangle)}(z_1)(U)\cup\zeta_{\mathcal{P}_{\mathrm{fin}}(\langle A,\le^e_A,\ell_{A}\rangle)}(z_2)(V))\]

Then, we have:

\[\begin{aligned}
(z_1\oplus z_2)\xrightarrow{U\parallel V}z_1'\oplus z_2' &\Leftrightarrow z_1'\oplus z_2'\in\zeta_{\mathcal{P}_{\mathrm{fin}}(\langle A,\le^e_A,\ell_{A}\rangle)}(z_1\oplus z_2)(U\parallel V)\\
&\Leftrightarrow z_1'\oplus z_2'\in\zeta_{\mathcal{P}_{\mathrm{fin}}(\langle A,\le^e_A,\ell_{A}\rangle)}(z_1)(U)\cup\zeta_{\mathcal{P}_{\mathrm{fin}}(\langle A,\le^e_A,\ell_{A}\rangle)}(z_2)(V)\\
&\Leftrightarrow z_1'\in\zeta_{\mathcal{P}_{\mathrm{fin}}(\langle A,\le^e_A,\ell_{A}\rangle)}(z_1)(U)\mbox{ and }z_2'\in\zeta_{\mathcal{P}_{\mathrm{fin}}(\langle A,\le^e_A,\ell_{A}\rangle)}(z_2)(V)\\
&\Leftrightarrow z_1\xrightarrow{U}z_1'\mbox{ and }z_2\xrightarrow{V}z_2'
\end{aligned}\]

which can be described by the following two rules:

\[\frac{z_1\xrightarrow{U}z_1'\quad z_2\xrightarrow{V}z_2'}{z_1\oplus z_2\xrightarrow{U\parallel V}z_1'\oplus z_2'}\]

For an arbitrary powerset $\mathcal{P}_{\mathrm{fin}}(\langle A,\le^e_A,\ell_{A}\rangle)$ of steps, let polynomial functor $\Sigma_{\mathcal{P}_{\mathrm{fin}}(\langle A,\le^e_A,\ell_{A}\rangle)}:\mathbf{Trans}\to\mathbf{Trans}$ given by

$\Sigma_{\mathcal{P}_{\mathrm{fin}}(\langle A,\le^e_A,\ell_{A}\rangle)}(\langle X,\le^e_X,\ell_{X}\rangle)
=1+(\mathcal{P}_{\mathrm{fin}}(\langle A,\le^e_A,\ell_{A}\rangle)\times\langle X,\le^e_X,\ell_{X}\rangle)+(\langle X,\le^e_X,\ell_{X}\rangle\times\langle X,\le^e_X,\ell_{X}\rangle)+(\langle X,\le^e_X,\ell_{X}\rangle\times\langle X,\le^e_X,\ell_{X}\rangle)$

An algebra $\Sigma_{\mathcal{P}_{\mathrm{fin}}(\langle A,\le^e_A,\ell_{A}\rangle)}(\langle X,\le^e_X,\ell_{X}\rangle)\to\langle X,\le^e_X,\ell_{X}\rangle$ for this functor consists of four operations: $\mathbf{0}:1\to\langle X,\le^e_X,\ell_{X}\rangle$ for null process, $.:\mathcal{P}_{\mathrm{fin}}(\langle A,\le^e_A,\ell_{A}\rangle)\times \langle X,\le^e_X,\ell_{X}\rangle\to\langle X,\le^e_X,\ell_{X}\rangle$ for prefixing process, $+:\langle X,\le^e_X,\ell_{X}\rangle\times\langle X,\le^e_X,\ell_{X}\rangle\to\langle X,\le^e_X,\ell_{X}\rangle$, $\oplus:\langle X,\le^e_X,\ell_{X}\rangle\times\langle X,\le^e_X,\ell_{X}\rangle\to\langle X,\le^e_X,\ell_{X}\rangle$ for parallel process. The final coalgebra $Z_{\mathcal{P}_{\mathrm{fin}}(\langle A,\le^e_A,\ell_{A}\rangle)}\xrightarrow{\cong}\mathcal{P}_{\mathrm{fin}}(Z_{\mathcal{P}_{\mathrm{fin}}(\langle A,\le^e_A,\ell_{A}\rangle)})^{\mathcal{P}_{\mathrm{fin}}(\langle A,\le^e_A,\ell_{A}\rangle)}$ carries a $\Sigma_{\mathcal{P}_{\mathrm{fin}}(\langle A,\le^e_A,\ell_{A}\rangle)}$ algebra written as $\xi_{\mathcal{P}_{\mathrm{fin}}(\langle A,\le^e_A,\ell_{A}\rangle)}:\Sigma_{\mathcal{P}_{\mathrm{fin}}(\langle A,\le^e_A,\ell_{A}\rangle)}(Z_{\mathcal{P}_{\mathrm{fin}}(\langle A,\le^e_A,\ell_{A}\rangle)})\to Z_{\mathcal{P}_{\mathrm{fin}}(\langle A,\le^e_A,\ell_{A}\rangle)}$. Thus, a bialgebra is obtained:

\[\begin{tikzcd}[row sep=huge, column sep=huge]
\Sigma_{\mathcal{P}_{\mathrm{fin}}(\langle A,\le^e_A,\ell_{A}\rangle)}(Z_{\mathcal{P}_{\mathrm{fin}}(\langle A,\le^e_A,\ell_{A}\rangle)}) \arrow[r, "\ensuremath{\xi_{\mathcal{P}_{\mathrm{fin}}(\langle A,\le^e_A,\ell_{A}\rangle)}}"] & Z_{\mathcal{P}_{\mathrm{fin}}(\langle A,\le^e_A,\ell_{A}\rangle)}\arrow[r,"\ensuremath{\zeta_{\mathcal{P}_{\mathrm{fin}}(\langle A,\le^e_A,\ell_{A}\rangle)}}","\ensuremath{\cong}"'] & \mathcal{P}_{\mathrm{fin}}(Z_{\mathcal{P}_{\mathrm{fin}}(\langle A,\le^e_A,\ell_{A}\rangle)})^{\mathcal{P}_{\mathrm{fin}}(\langle A,\le^e_A,\ell_{A}\rangle)} 
\end{tikzcd}\] 

The initial $\Sigma_{\mathcal{P}_{\mathrm{fin}}(\langle A,\le^e_A,\ell_{A}\rangle)}$-algebra denoted $P_{\mathcal{P}_{\mathrm{fin}}(\langle A,\le^e_A,\ell_{A}\rangle)}$ as $\alpha:\Sigma_{\mathcal{P}_{\mathrm{fin}}(\langle A,\le^e_A,\ell_{A}\rangle)}(P_{\mathcal{P}_{\mathrm{fin}}(\langle A,\le^e_A,\ell_{A}\rangle)})\xrightarrow{\cong}P_{\mathcal{P}_{\mathrm{fin}}(\langle A,\le^e_A,\ell_{A}\rangle)}$, contains the closed process terms, which are built up from $\mathbf{0}$ and $U\in \mathcal{P}_{\mathrm{fin}}(\langle A,\le^e_A,\ell_{A}\rangle)$. $P_{\mathcal{P}_{\mathrm{fin}}(\langle A,\le^e_A,\ell_{A}\rangle)}$ also carries a bialgebra structure:

\[\begin{tikzcd}[row sep=huge, column sep=huge]
\Sigma_{\mathcal{P}_{\mathrm{fin}}(\langle A,\le^e_A,\ell_{A}\rangle)}(P_{\mathcal{P}_{\mathrm{fin}}(\langle A,\le^e_A,\ell_{A}\rangle)}) \arrow[r, "\ensuremath{\alpha}","\ensuremath{\cong}"'] & P_{\mathcal{P}_{\mathrm{fin}}(\langle A,\le^e_A,\ell_{A}\rangle)}\arrow[r,"\ensuremath{\beta}"] & \mathcal{P}_{\mathrm{fin}}(P_{\mathcal{P}_{\mathrm{fin}}(\langle A,\le^e_A,\ell_{A}\rangle)})^{\mathcal{P}_{\mathrm{fin}}(\langle A,\le^e_A,\ell_{A}\rangle)} 
\end{tikzcd}\]

The coalgebra $\beta$ is defined by induction:

\[\beta(\mathbf{0})(U)=\emptyset\]
\[\beta(V.s)(U)=\{s|V\cong U\}\]
\[\beta(s+t)(U)=\beta(s)(U)\cup\beta(t)(U)\]
\[\beta(s\oplus t)(U\parallel V)=\beta(s)(U)\cup\beta(t)(V)\]

The denotational semantics given by initiality and operational semantics given by finality for process terms coincide.

\begin{proposition}
In the above situation, we obtain two maps $P_{\mathcal{P}_{\mathrm{fin}}(\langle A,\le^e_A,\ell_{A}\rangle)}\to Z_{\mathcal{P}_{\mathrm{fin}}(\langle A,\le^e_A,\ell_{A}\rangle)}$, one by initiality making the following diagram commute:

\begin{center}
\begin{tikzcd}[row sep=huge, column sep=huge]
\Sigma_{\mathcal{P}_{\mathrm{fin}}(\langle A,\le^e_A,\ell_{A}\rangle)}(P_{\mathcal{P}_{\mathrm{fin}}(\langle A,\le^e_A,\ell_{A}\rangle)}) \arrow[dashed,r,"\ensuremath{\Sigma_{\mathcal{P}_{\mathrm{fin}}(\langle A,\le^e_A,\ell_{A}\rangle)}(\mathrm{int}_{\alpha})}"] \arrow[d,"\ensuremath{\alpha}"',"\cong"] & \Sigma_{\mathcal{P}_{\mathrm{fin}}(\langle A,\le^e_A,\ell_{A}\rangle)}(Z_{\mathcal{P}_{\mathrm{fin}}(\langle A,\le^e_A,\ell_{A}\rangle)})  \arrow[d,"\ensuremath{\xi}"]\\
P_{\mathcal{P}_{\mathrm{fin}}(\langle A,\le^e_A,\ell_{A}\rangle)} \arrow[dashed,r,"\ensuremath{\mathrm{int}_{\alpha}}"'] & Z_{\mathcal{P}_{\mathrm{fin}}(\langle A,\le^e_A,\ell_{A}\rangle)}
\end{tikzcd}
\end{center}

the other by finality making the following diagram commute:

\begin{center}
\begin{tikzcd}[row sep=huge, column sep=huge]
\mathcal{P}_{\mathrm{fin}}(P_{\mathcal{P}_{\mathrm{fin}}(\langle A,\le^e_A,\ell_{A}\rangle)})^{\mathcal{P}_{\mathrm{fin}}(\langle A,\le^e_A,\ell_{A}\rangle)} \arrow[dashed,r,"\ensuremath{\mathcal{P}_{\mathrm{fin}}(\mathrm{beh}_{\beta})^{\mathcal{P}_{\mathrm{fin}}(\langle A,\le^e_A,\ell_{A}\rangle)}}"] & \mathcal{P}_{\mathrm{fin}}(Z_{\mathcal{P}_{\mathrm{fin}}(\langle A,\le^e_A,\ell_{A}\rangle)})^{\mathcal{P}_{\mathrm{fin}}(\langle A,\le^e_A,\ell_{A}\rangle)}\\
P_{\mathcal{P}_{\mathrm{fin}}(\langle A,\le^e_A,\ell_{A}\rangle)} \arrow[dashed,r,"\ensuremath{\mathrm{beh}_{\beta}}"'] \arrow[u,"\ensuremath{\beta}"]& Z_{\mathcal{P}_{\mathrm{fin}}(\langle A,\le^e_A,\ell_{A}\rangle)}\arrow[u,"\ensuremath{\cong}","\ensuremath{\zeta}"']
\end{tikzcd}
\end{center}
\end{proposition}

\begin{proof}
By induction on the structure of a term $s\in P_{\mathcal{P}_{\mathrm{fin}}(\langle A,\le^e_A,\ell_{A}\rangle)}$, we prove that $\mathrm{beh}_{\beta}(s)=\mathrm{int}_{\alpha}(s)$, or equivalently, $\zeta(\mathrm{beh}_{\beta}(s))(U)=\zeta(\mathrm{int}_{\alpha}(s))(U)$:

\[\begin{aligned}
\zeta(\mathrm{beh}_{\beta}(\mathbf{0}))(U)&=\mathcal{P}_{\mathrm{fin}}(\mathrm{beh}_{\beta})^{\mathcal{P}_{\mathrm{fin}}(\langle A,\le^e_A,\ell_{A}\rangle)}(\beta(\mathbf{0}))(U)\\
&=\mathcal{P}_{\mathrm{fin}}(\mathrm{beh}_{\beta})(\beta(\mathbf{0})(U))\\
&=\mathcal{P}_{\mathrm{fin}}(\mathrm{beh}_{\beta}(\emptyset))\\
&=\emptyset\\
&=\zeta(\mathbf{0})(U)\\
&=\zeta(\mathrm{int}_{\alpha}(\mathbf{0}))(U)
\end{aligned}\]

\[\begin{aligned}
\zeta(\mathrm{beh}_{\beta}(V.s))(U)&=\mathcal{P}_{\mathrm{fin}}(\mathrm{beh}_{\beta})(\beta(V.s))(U)\\
&=\mathcal{P}_{\mathrm{fin}}(\mathrm{beh}_{\beta})(\{s|V\cong U\})\\
&=\{\mathrm{beh}_{\beta}(s)|V\cong U\}\\
&=\{\mathrm{int}_{\alpha}(s)|V\cong U\}\\
&=\zeta(V.\mathrm{int}_{\alpha}(s))(U)\\
&=\zeta(\mathrm{int}_{\alpha}(V.s))(U)
\end{aligned}\]

\[\begin{aligned}
\zeta(\mathrm{beh}_{\beta}(s+t))(U)&=\mathcal{P}_{\mathrm{fin}}(\mathrm{beh}_{\beta})(\beta(s+t))(U)\\
&=\mathcal{P}_{\mathrm{fin}}(\mathrm{beh}_{\beta})(\beta(s)(U)\cup\beta(t)(U))\\
&=\mathcal{P}_{\mathrm{fin}}(\mathrm{beh}_{\beta})(\beta(s)(U))\cup \mathcal{P}_{\mathrm{fin}}(\mathrm{beh}_{\beta})(\beta(t)(U))\\
&=\zeta(\mathrm{beh}_{\beta}(s))(U)\cup \zeta(\mathrm{beh}_{\beta}(t))(U)\\
&=\zeta(\mathrm{int}_{\alpha}(s))(U)\cup \zeta(\mathrm{int}_{\alpha}(t))(U)\\
&=\zeta(\mathrm{int}_{\alpha}(s)+ \mathrm{int}_{\alpha}(t))(U)\\
&=\zeta(\mathrm{int}_{\alpha}(s+t))(U)
\end{aligned}\]

\[\begin{aligned}
\zeta(\mathrm{beh}_{\beta}(s\oplus t))(U\parallel V)&=\mathcal{P}_{\mathrm{fin}}(\mathrm{beh}_{\beta})(\beta(s\oplus t))(U\parallel V)\\
&=\mathcal{P}_{\mathrm{fin}}(\mathrm{beh}_{\beta})(\beta(s)(U)\cup\beta(t)(V))\\
&=\mathcal{P}_{\mathrm{fin}}(\mathrm{beh}_{\beta})(\beta(s)(U))\cup \mathcal{P}_{\mathrm{fin}}(\mathrm{beh}_{\beta})(\beta(t)(V))\\
&=\zeta(\mathrm{beh}_{\beta}(s))(U)\cup \zeta(\mathrm{beh}_{\beta}(t))(V)\\
&=\zeta(\mathrm{int}_{\alpha}(s))(U)\cup \zeta(\mathrm{int}_{\alpha}(t))(V)\\
&=\zeta(\mathrm{int}_{\alpha}(s)\oplus \mathrm{int}_{\alpha}(t))(U\parallel V)\\
&=\zeta(\mathrm{int}_{\alpha}(s\oplus t))(U\parallel V)
\end{aligned}\]
\end{proof}

\begin{proposition}
Still in the above situation, the step bisimilarity relation $\underline{\leftrightarrow}$ on the set $P_{\mathcal{P}_{\mathrm{fin}}(\langle A,\le^e_A,\ell_{A}\rangle)}$ of process terms is a congruence.
\end{proposition}

\begin{proof}
In the following diagram, $f=\mathrm{beh}_{\beta}=\mathrm{int}_{\alpha}$.

\begin{center}
\adjustbox{rotate=90}{
\begin{tikzcd}[row sep=huge, column sep=huge]
\Sigma_{\mathcal{P}_{\mathrm{fin}}(\langle A,\le^e_A,\ell_{A}\rangle)}(\underline{\leftrightarrow})\arrow[r,"d"] \arrow[dashed,d]&\Sigma_{\mathcal{P}_{\mathrm{fin}}(\langle A,\le^e_A,\ell_{A}\rangle)}(P_{\mathcal{P}_{\mathrm{fin}}(\langle A,\le^e_A,\ell_{A}\rangle)})\times\Sigma_{\mathcal{P}_{\mathrm{fin}}(\langle A,\le^e_A,\ell_{A}\rangle)}(P_{\mathcal{P}_{\mathrm{fin}}(\langle A,\le^e_A,\ell_{A}\rangle)}) \arrow[r,shift left,"\ensuremath{\Sigma_{\mathcal{P}_{\mathrm{fin}}(\langle A,\le^e_A,\ell_{A}\rangle)}(f\circ\pi_1)}"]\arrow[r,shift right,"\ensuremath{\Sigma_{\mathcal{P}_{\mathrm{fin}}(\langle A,\le^e_A,\ell_{A}\rangle)}(f\circ\pi_2)}"'] \arrow[d,"\ensuremath{\alpha\times\alpha}"] & \Sigma_{\mathcal{P}_{\mathrm{fin}}(\langle A,\le^e_A,\ell_{A}\rangle)}(Z_{\mathcal{P}_{\mathrm{fin}}(\langle A,\le^e_A,\ell_{A}\rangle)})  \arrow[d,"\ensuremath{\xi_{\mathcal{P}_{\mathrm{fin}}(\langle A,\le^e_A,\ell_{A}\rangle)}}"]\\
\underline{\leftrightarrow} \arrow[r,tail,"e"'] &P_{\mathcal{P}_{\mathrm{fin}}(\langle A,\le^e_A,\ell_{A}\rangle)}\times P_{\mathcal{P}_{\mathrm{fin}}(\langle A,\le^e_A,\ell_{A}\rangle)} \arrow[r,shift left,"\ensuremath{f\circ \pi_1}"]\arrow[r,shift right,"\ensuremath{f\circ \pi_2}"'] & Z_{\mathcal{P}_{\mathrm{fin}}(\langle A,\le^e_A,\ell_{A}\rangle)}
\end{tikzcd}}
\end{center}

The map $e$ is the equalizer of $f\circ\pi_1$ and $f\circ \pi_2$ by \cref{tbisimulation1}, and $d$ is the pair $\langle\Sigma_{\mathcal{P}_{\mathrm{fin}}(\langle A,\le^e_A,\ell_{A}\rangle)}(\pi_1\circ e),\Sigma_{\mathcal{P}_{\mathrm{fin}}(\langle A,\le^e_A,\ell_{A}\rangle)}(\pi_2\circ e)\rangle$. We show that $(\alpha\times\alpha)\circ d$ equalizes $f\circ\pi_1$ and $f\circ\pi_2$. The dashed arrow then exists by the universal property of the equalizer $e$, make $\underline{\leftrightarrow}$ a congruence by \cref{tcongruence1}.

\[\begin{aligned}
f\circ\pi_1\circ(\alpha\times\alpha)\circ d &=\mathrm{int}_{\alpha}\circ\alpha\circ\pi_1\circ d\\
&=\xi\circ\Sigma_{\mathcal{P}_{\mathrm{fin}}(\langle A,\le^e_A,\ell_{A}\rangle)}(\mathrm{int}_{\alpha})\circ\Sigma_{\mathcal{P}_{\mathrm{fin}}(\langle A,\le^e_A,\ell_{A}\rangle)}(\pi_1\circ e)\\
&=\xi\circ\Sigma_{\mathcal{P}_{\mathrm{fin}}(\langle A,\le^e_A,\ell_{A}\rangle)}(\mathrm{int}_{\alpha})\circ\Sigma_{\mathcal{P}_{\mathrm{fin}}(\langle A,\le^e_A,\ell_{A}\rangle)}(\pi_2\circ e)\\
&=\mathrm{int}_{\alpha}\circ\alpha\circ\pi_2\circ d\\
&=f\circ\pi_2\circ(\alpha\times\alpha)\circ d
\end{aligned}\]
\end{proof}

\begin{proposition}
The so-called exchange laws hold:

\[(U.s)\oplus(V.t)\underline{\leftrightarrow}(U\parallel V).(s\oplus t)\]
\end{proposition} 

\newpage\section{Execution Phase}\label{ep}

In this section, we introduce the execution phase. During the execution phase, the AI agent workflow is executed. The category of agents $\mathbf{Agnts}$ can still be deemed as $\mathbf{LMSets}$ equipped with an execution partial order, similarly to $\mathbf{Trans}$, and we only give the definition. The category of AI agent orchestration and choreography $\mathbf{OCs}$ is treated as $\mathbf{LMSets}$ equipped with an execution partial order and a communication partial order. We then introduce the coalgebra of $\mathbf{OCs}$, and also the algebra of orchestration and choreography which is introduced in \cite{APTC} \cite{APTC2} firstly. 

\subsection{Category of Agents $\mathbf{Agnts}$}

\begin{definition}[Category of Agents $\mathbf{Agnts}$]
An object in $\mathbf{Obj}(\mathbf{Agnts})$ is a triple $\langle X,\le^e_X,\ell_X:X\to \Sigma_X\rangle$ where:

\begin{itemize}
  \item $X$ is a set (the set of occurrences / instances of the multiset).
  \item $\le^e_X$ is an execution partial order on $X$ (reflexive, antisymmetric, transitive), and $\langle X,\le^e_X\rangle$ is an N-free poset.
  \item $\Sigma_X$ is a set (label alphabet of $X$).
  \item $\ell_X: X\to \Sigma_X$ is a function (labelling map).
\end{itemize}

Given $\langle X,\le^e_X,\ell_X:X\to\Sigma_X\rangle$, $\langle Y,\le^e_Y,\ell_Y:Y\to\Sigma_Y\rangle$, a morphism $\langle f,\sigma\rangle : \langle X,\le^e_X,\ell_X:X\to\Sigma_X\rangle\to \langle Y,\le^e_Y,\ell_Y:Y\to\Sigma_Y\rangle$ is a pair $(f,\sigma)$ with $f: X\to Y,\sigma: \Sigma_X\to \Sigma_Y$ satisfying:

\begin{enumerate}
  \item Commutativity: $\ell_Y\circ f = \sigma \circ \ell_X$.
  \item Monotonicity: $\forall x_1,x_2\in X.x_1\le^e_X x_2 \Leftrightarrow f(x_1)\le^e_Y f(x_2)$.
\end{enumerate}

If $\langle f,\sigma\rangle: \langle X,\le^e_X,\ell_X:X\to\Sigma_X\rangle\to \langle Y,\le^e_Y,\ell_Y:Y\to\Sigma_Y\rangle$, $\langle g,\xi\rangle: \langle X,\le^e_Y,\ell_Y:Y\to\Sigma_Y\rangle\to \langle Z,\le^e_Z,\ell_Z:Z\to\Sigma_Z\rangle$, define $\langle g,\xi\rangle \circ \langle f,\sigma\rangle \triangleq \langle g\circ f, \xi\circ \sigma\rangle$.

For $A=\langle X,\le^e_X,\ell_X:X\to\Sigma_X\rangle$, $\mathrm{id}_A \triangleq \langle \mathrm{id}_X,\mathrm{id}_{\Sigma_X}\rangle$.

$\mathbf{Agnts}$ forms a category.
\end{definition}

\subsection{Category of Orchestrations and Choreographies $\mathbf{OCs}$}

\begin{definition}[Category of orchestrations and choreographies $\mathbf{OCs}$]
An object in $\mathbf{Obj}(\mathbf{OCs})$ is a quadruple $\langle X,\le^e_X,\le^c_X,\ell_X:X\to \Sigma_X\rangle$ where:

\begin{itemize}
  \item $X$ is a set (the set of occurrences / instances of the multiset).
  \item $\le^e_X$ is an execution partial order on $X$ (reflexive, antisymmetric, transitive), and $\langle X,\le^e_X\rangle$ is an N-free poset.
  \item $\le^c_X$ is a communication partial order on $X$ (reflexive, antisymmetric, transitive), and $\langle X, \le^e_X,\le^c_X\rangle$ may contain N-shapes in which partial orders between different parallel branches are all the forms of $\le^c_X$.
  \item $\Sigma_X$ is a set (label alphabet of $X$).
  \item $\ell_X: X\to \Sigma_X$ is a function (labelling map).
\end{itemize}

Given $\langle X,\le^e_X,\le^c_X,\ell_X:X\to\Sigma_X\rangle$, $\langle Y,\le^e_Y,\le^c_Y,\ell_Y:Y\to\Sigma_Y\rangle$, a morphism $\langle f,\sigma\rangle : \langle X,\le^e_X,\le^c_X,\ell_X:X\to\Sigma_X\rangle\to \langle Y,\le^e_Y,\le^c_Y,\ell_Y:Y\to\Sigma_Y\rangle$ is a pair $(f,\sigma)$ with $f: X\to Y,\sigma: \Sigma_X\to \Sigma_Y$ satisfying:

\begin{enumerate}
  \item Commutativity: $\ell_Y\circ f = \sigma \circ \ell_X$.
  \item Monotonicity: $\forall x_1,x_2\in X.x_1\le^e_X x_2 \Leftrightarrow f(x_1)\le^e_Y f(x_2)$ and $\forall x_1,x_2\in X.x_1\le^c_X x_2 \Leftrightarrow f(x_1)\le^c_Y f(x_2)$.
\end{enumerate}

If $\langle f,\sigma\rangle: \langle X,\le^e_X,\le^c_X,\ell_X:X\to\Sigma_X\rangle\to \langle Y,\le^e_Y,\le^c_Y,\ell_Y:Y\to\Sigma_Y\rangle$, $\langle g,\xi\rangle: \langle X,\le^e_Y,\ell_Y:Y\to\Sigma_Y\rangle\to \langle Z,\le^e_Z,\le^c_Z,\ell_Z:Z\to\Sigma_Z\rangle$, define $\langle g,\xi\rangle \circ \langle f,\sigma\rangle \triangleq \langle g\circ f, \xi\circ \sigma\rangle$.

For $A=\langle X,\le^e_X,\le^c_X,\ell_X:X\to\Sigma_X\rangle$, $\mathrm{id}_A \triangleq \langle \mathrm{id}_X,\mathrm{id}_{\Sigma_X}\rangle$.

$\mathbf{OCs}$ forms a category.
\end{definition}

\begin{proof}
We assume $\langle f,\sigma\rangle: \langle X,\le^e_X,\le^c_X,\ell_X:X\to\Sigma_X\rangle\to \langle Y,\le^e_Y,\le^c_Y,\ell_Y:Y\to\Sigma_Y\rangle, \langle g,\xi\rangle: \langle Y,\le^e_Y,\le^c_Y,\ell_Y:Y\to\Sigma_Y\rangle\to \langle Z,\le^e_Z,\le^c_Z,\ell_Z:Z\to\Sigma_Z\rangle$ are morphisms, and prove $\langle g\circ f,\xi\circ\sigma\rangle$ satisfies the two morphism conditions.

\begin{enumerate}
  \item Commutativity:
\[\begin{aligned}
\ell_C \circ (g\circ f)
&= (\ell_C \circ g)\circ f \\
&= (\xi \circ \ell_Y)\circ f \\
&= \xi \circ (\ell_Y \circ f) \\
&= \xi \circ (\sigma \circ \ell_X) \\
&= (\xi\circ \sigma)\circ \ell_X.
\end{aligned}\]

So $\ell_C \circ (g\circ f) = (\xi\circ\sigma)\circ \ell_X$.
  \item Monotonicity: take $x_1,x_2\in X$ with $x_1\le^e_X x_2$ and/or $x_1\le^c_X x_2$. Since $\langle f,\sigma\rangle$ is monotone, $f(x_1)\le^e_Y f(x_2)$ and/or $f(x_1)\le^c_Y f(x_2)$. Since $\langle g,\xi\rangle$ is monotone, $g(f(x_1)) \le_Z^e g(f(x_2))$ and/or $g(f(x_1)) \le_Z^c g(f(x_2))$. Thus $g\circ f$ is monotone.
\end{enumerate}

Therefore $\langle g\circ f,\xi\circ\sigma\rangle$ is indeed a morphism $\langle X,\le^e_X,\le^c_X,\ell_X:X\to\Sigma_X\rangle\to \langle Z,\le^e_Z,\le^c_Z,\ell_Z:Z\to\Sigma_Z\rangle$.

For $A=\langle X,\le^e_X,\le^c_X,\ell_X:X\to\Sigma_X\rangle$, $\mathrm{id}_A = \langle \mathrm{id}_X,\mathrm{id}_{\Sigma_X}\rangle$. Check the two conditions:

\begin{enumerate}
  \item Commutativity: $\ell_X \circ \mathrm{id}_X= \ell_X= \mathrm{id}_{\Sigma_X}\circ \ell_X$.
  \item Monotonicity: if $x_1\le^e_X x_2$, then $\mathrm{id}_X(x_1)=x_1\le^e_X x_2=\mathrm{id}_X(x_2)$; if $x_1\le^c_X x_2$, then $\mathrm{id}_X(x_1)=x_1\le^c_X x_2=\mathrm{id}_X(x_2)$.
\end{enumerate}

Hence $\mathrm{id}_A : A\to A$ is a morphism.

Then, we prove the associativity of composition.

Take three composable morphisms: $\langle f,\sigma\rangle: \langle X,\le^e_X,\le^c_X,\ell_X:X\to\Sigma_X\rangle\to \langle Y,\le^e_Y,\le^c_Y,\ell_Y:Y\to\Sigma_Y\rangle$,$\langle g,\xi\rangle: \langle Y,\le^e_Y,\le^c_Y,\ell_Y:Y\to\Sigma_Y\rangle\to \langle Z,\le^e_Z,\le^c_Z,\ell_Z:Z\to\Sigma_Z\rangle$,$\langle h,\zeta\rangle: \langle Z,\le^e_Z,\le^c_Z,\ell_Z:Z\to\Sigma_Z\rangle\to \langle W,\le^e_W,\le^c_W,\ell_W:W\to\Sigma_W\rangle$. By definition:

\[\begin{aligned}
(\langle h,\zeta\rangle \circ \langle g,\xi\rangle)\circ \langle f,\sigma\rangle&=\langle h\circ g,\zeta\circ\xi\rangle\circ \langle f,\sigma\rangle \\
&=\langle (h\circ g)\circ f, (\zeta\circ\xi)\circ \sigma\rangle
\end{aligned}\]

and

\[\begin{aligned}
\langle h,\zeta\rangle\circ(\langle g,\xi\rangle \circ \langle f,\sigma\rangle)&=\langle h,\zeta\rangle\circ\langle g\circ f,\xi\circ\sigma\rangle \\
&=\langle h\circ(g\circ f), \zeta\circ(\xi\circ\sigma)\rangle
\end{aligned}\]

Function composition on sets is associative: $(h\circ g)\circ f = h\circ(g\circ f),(\zeta\circ\xi)\circ\sigma = \zeta\circ(\xi\circ\sigma)$. Therefore the two composite morphisms are equal: $(\langle h,\zeta\rangle \circ \langle g,\xi\rangle)\circ \langle f,\sigma\rangle=\langle h,\zeta\rangle\circ(\langle g,\xi\rangle \circ \langle f,\sigma\rangle)$.

Lastly, we prove the identity laws.

Let $\langle f,\sigma\rangle: A=\langle X,\le^e_X,\le^c_X,\ell_X:X\to\Sigma_X\rangle\to B=\langle Y,\le^e_Y,\ell_X:Y\to\Sigma_Y\rangle$.

\[\begin{aligned}
\langle f,\sigma\rangle \circ \mathrm{id}_A&=\langle f,\sigma\rangle\circ\langle \mathrm{id}_X,\mathrm{id}_{\Sigma_X}\rangle \\
&=\langle f\circ \mathrm{id}_X, \sigma\circ \mathrm{id}_{\Sigma_X}\rangle
&=\langle f,\sigma\rangle
\end{aligned}\]

\[\begin{aligned}
\mathrm{id}_B \circ \langle f,\sigma\rangle&=\langle \mathrm{id}_Y,\mathrm{id}_{\Sigma_Y}\rangle\circ\langle f,\sigma\rangle \\
&=\langle \mathrm{id}_Y \circ f,\; \mathrm{id}_{\Sigma_Y}\circ \sigma\rangle
&=\langle f,\sigma\rangle
\end{aligned}\]

Therefore, $\mathbf{OCs}$ is a category.
\end{proof}

\subsubsection{Polynomial Functors of $\mathbf{OCs}$}

\runhead{Products}

For lmsets $\langle X,\le^e_X,\le^c_X,\ell_{X}\rangle$ and $\langle Y,\le^e_Y,\le^c_Y,\ell_{Y}\rangle$, their products written as $\langle X,\le^e_X,\le^c_X,\ell_{X}\rangle\times \langle Y,\le^e_Y,\le^c_Y,\ell_{Y}\rangle$, is defined as:

\[\langle X,\le^e_X,\le^c_X,\ell_{X}\rangle\times \langle Y,\le^e_Y,\le^c_Y,\ell_{Y}\rangle=\{(x,y)|x\in \langle X,\le^e_X,\le^c_X,\ell_{X}\rangle\wedge y\in \langle Y,\le^e_Y,\le^c_Y,\ell_{Y}\rangle\}\]

The product operation $\times$ is a functor $\mathbf{OCs}\times\mathbf{OCs}\to\mathbf{OCs}$.

\begin{definition}[Product]\label{definitionofproduct}
Let us take two objects $\langle X,\le^e_X,\le^c_X,\ell_X\rangle$ and $\langle Y,\le^e_Y,\le^c_Y,\ell_Y\rangle$ in $\mathbf{OCs}$. The product of these two objects is a new object $\langle X,\le^e_X,\le^c_X,\ell_{X}\rangle\times\langle Y,\le^e_Y,\le^c_Y,\ell_{Y}\rangle$ with two projection morphisms:

\[\begin{tikzcd}[row sep=huge, column sep=huge]
\langle X,\le^e_X,\le^c_X,\ell_{X}\rangle & \langle X,\le^e_X,\le^c_X,\ell_{X}\rangle\times\langle Y,\le^e_Y,\le^c_Y,\ell_{Y}\rangle \arrow[l, "\pi_1"'] \arrow[r, "\pi_2"] & \langle Y,\le^e_Y,\le^c_Y,\ell_{Y}\rangle 
\end{tikzcd}\]

For functions $f:\langle Z,\le^e_Z,\le^c_Z,\ell_{Z}\rangle\to\langle X,\le^e_X,\le^c_X,\ell_{X}\rangle$ and $g:\langle Z,\le^e_Z,\le^c_Z,\ell_{Z}\rangle\to\langle Y,\le^e_Y,\le^c_Y,\ell_{Y}\rangle$, (where $f$ contains a pair of functions $f:\langle f_{f},f_{\sigma}\rangle : \langle Z,\le^e_Z,\le^c_Z,\ell_{Z}:Z\to\Sigma_{Z}\rangle \to \langle X,\le^e_X,\le^c_X,\ell_{X}:X\to\Sigma_{X}\rangle$ and where $g$ also contains a pair of functions $g:\langle g_{g},g_{\xi}\rangle : \langle Z,\le^e_Z,\le^c_Z,\ell_{Z}:Z\to\Sigma_{Z}\rangle \to \langle Y,\le^e_Y,\le^c_Y,\ell_{Y}:Y\to\Sigma_{Y}\rangle$, in the following chapters, if no needs, we do not expand them and use their collapsed forms), there is a unique pairing function $\langle f,g\rangle:\langle Z,\le^e_Z,\le^c_Z,\ell_{Z}\rangle\to\langle X,\le^e_X,\le^c_X,\ell_{X}\rangle\times\langle Y,\le^e_Y,\le^c_Y,\ell_{Y}\rangle$ with $\langle f,g\rangle(z)=(f(z),g(z))\in \langle X,\le^e_X,\le^c_X,\ell_{X}\rangle\times\langle Y,\le^e_Y,\le^c_Y,\ell_{Y}\rangle$ for $z\in\langle Z,\le^e_Z,\le^c_Z,\ell_{Z}\rangle$, such that the following equations hold:

\[\pi_1\circ\langle f,g\rangle=f\qquad \langle \pi_1,\pi_2\rangle=\mathrm{id}_{\langle X,\le^e_X,\le^c_X,\ell_{X}\rangle\times\langle Y,\le^e_Y,\le^c_Y,\ell_{Y}\rangle}\]
\[\pi_2\circ\langle f,g\rangle=g\qquad \langle f,g\rangle\circ h=\langle f\circ h,g\circ h\rangle\]

where $h:\langle W,\le^e_W,\le^c_W,\ell_{W}\rangle\to\langle Z,\le^e_Z,\le^c_Z,\ell_{Z}\rangle$.

For the above morphisms $f$ and $g$, there is a unique pairing morphism $\langle f,g\rangle$ defined as above, making the following diagram commute:

\[\begin{tikzcd}[row sep=huge, column sep=huge]
\langle X,\le^e_X,\le^c_X,\ell_{X}\rangle & \langle X,\le^e_X,\le^c_X,\ell_{X}\rangle\times\langle Y,\le^e_Y,\le^c_Y,\ell_{Y}\rangle \arrow[l, "\pi_1"'] \arrow[r, "\pi_2"] & \langle Y,\le^e_Y,\le^c_Y,\ell_{Y}\rangle \\
&\langle Z,\le^e_Z,\le^c_Z,\ell_{Z}\rangle \arrow[lu,"f"] \arrow[dashed,u,"\ensuremath{\langle f,g\rangle}"description] \arrow[ru,"g"']&
\end{tikzcd}\]
\end{definition}

For functions $f:\langle X,\le^e_X,\le^c_X,\ell_{X}\rangle\to\langle X',\le^e_{X'},\le^c_{X'},\ell_{X'}\rangle$ and $g:\langle Y,\le^e_Y,\le^c_Y,\ell_{Y}\rangle\to\langle Y',\le^e_{Y'},\le^c_{Y'},\ell_{Y'}\rangle$, the product of functions $f\times g$ is defined as:

\[\begin{tikzcd}[row sep=huge, column sep=huge]
\langle X,\le^e_X,\le^c_X,\ell_{X}\rangle\times\langle Y,\le^e_Y,\le^c_Y,\ell_{Y}\rangle \arrow[r, "f\times g"] & \langle X',\le^e_{X'},\le^c_{X'},\ell_{X'}\rangle\times\langle Y',\le^e_{Y'},\le^c_{Y'},\ell_{Y'}\rangle 
\end{tikzcd}\] 

which is given by $(x,y)\mapsto(f(x),g(y))$, where $x\in\langle X,\le^e_X,\le^c_X,\ell_{X}\rangle$ and $y\in\langle Y,\le^e_Y,\le^c_Y,\ell_{Y}\rangle$, such that the following equations hold:

\[f\times g=\langle f\circ\pi_1,g\circ\pi_2\rangle\qquad \mathrm{id}_{\langle X,\le^e_X,\le^c_X,\ell_{X}\rangle}\times\mathrm{id}_{\langle Y,\le^e_Y,\le^c_Y,\ell_{Y}\rangle}=\mathrm{id}_{\langle X,\le^e_X,\le^c_X,\ell_{X}\rangle\times\langle Y,\le^e_Y,\le^c_Y,\ell_{Y}\rangle}\]
\[(f\circ h)\times(g\circ k)=(f\times g)\circ(h\times k)\]

Two functions $\langle Z,\le^e_Z,\le^c_Z,\ell_{Z}\rangle\to\langle X,\le^e_X,\le^c_X,\ell_{X}\rangle$ and $\langle Z,\le^e_Z,\le^c_Z,\ell_{Z}\rangle\to\langle Y,\le^e_Y,\le^c_Y,\ell_{Y}\rangle$ can be composed into $\langle Z,\le^e_Z,\le^c_Z,\ell_{Z}\rangle\to\langle X,\le^e_X,\le^c_X,\ell_{X}\rangle\times\langle Y,\le^e_Y,\le^c_Y,\ell_{Y}\rangle$ by \cref{definitionofproduct} and $\langle Z,\le^e_Z,\le^c_Z,\ell_{Z}\rangle\to\langle X,\le^e_X,\le^c_X,\ell_{X}\rangle\times\langle Y,\le^e_Y,\le^c_Y,\ell_{Y}\rangle$ can be decomposed into $\langle Z,\le^e_Z,\le^c_Z,\ell_{Z}\rangle\to\langle X,\le^e_X,\le^c_X,\ell_{X}\rangle$ and $\langle Z,\le^e_Z,\le^c_Z,\ell_{Z}\rangle\to\langle Y,\le^e_Y,\le^c_Y,\ell_{Y}\rangle$ by use of projections $\pi_1$ and $\pi_2$, which can be expressed by the following doubled line rule working in two directions:

\[\doublefrac{\langle Z,\le^e_Z,\le^c_Z,\ell_{Z}\rangle\to\langle X,\le^e_X,\le^c_X,\ell_{X}\rangle\quad \langle Z,\le^e_Z,\le^c_Z,\ell_{Z}\rangle\to\langle Y,\le^e_Y,\le^c_Y,\ell_{Y}\rangle}{\langle Z,\le^e_Z,\le^c_Z,\ell_{Z}\rangle\to\langle X,\le^e_X,\le^c_X,\ell_{X}\rangle\times\langle Y,\le^e_Y,\le^c_Y,\ell_{Y}\rangle}\] 

The $n$-fold product called power is defined as $\langle X,\le^e_X,\le^c_X,\ell_{X}\rangle^{n}=\langle X,\le^e_X,\le^c_X,\ell_{X}\rangle\times\cdots\times\langle X,\le^e_X,\le^c_X,\ell_{X}\rangle$ for a given $\langle X,\le^e_X,\le^c_X,\ell_{X}\rangle$ and $n\in\mathbb{N}$. And the empty product $\langle X,\le^e_X,\le^c_X,\ell_{X}\rangle^{0}$ is called a final or terminal object.

\begin{definition}[Final object]
A final object in a category $\mathbb{C}$ is an object $1\in\mathbb{C}$, such that for each object $X\in\mathbb{C}$ there is a unique morphism $!_{X}:X\to 1$ in $\mathbb{C}$.
\end{definition} 

\begin{proposition}[Final object of $\mathbf{OCs}$]
$\mathbf{OCs}$ has a final object.
\end{proposition}

\begin{proof}
Any singleton labelled occurrence multiset is final.
\end{proof}

$\mathbf{OCs}$ has finite products because it has binary products and a final object $1$. Finite products can be generalized to set-indexed products: given an index set $I$, a collection $(\langle X,\le^e_X,\le^c_X,\ell_{X}\rangle)_{i\in I}$ of $I$-indexed objects has $I$-indexed products $\langle X,\le^e_X,\le^c_X,\ell_{X}\rangle=\prod_{i\in I}\langle X_{i},\le^e_{X_{i}},\le^c_{X_{i}},\ell_{X_{i}}\rangle$ with projections $\pi_{i}:\langle X,\le^e_X,\le^c_X,\ell_{X}\rangle\to\langle X_{i},\le^e_{X_{i}},\le^c_{X_{i}},\ell_{X_{i}}\rangle$ for $i\in I$, which are universal as in \cref{definitionofproduct}: for an arbitrary object $\langle Y,\le^e_Y,\le^c_Y,\ell_{Y}\rangle$ and an $I$-indexed collection $f_{i}:\langle Y,\le^e_Y,\le^c_Y,\ell_{Y}\rangle\to\langle X_{i},\le^e_{X_{i}},\le^c_{X_{i}},\ell_{X_{i}}\rangle$ of morphisms there is a unique morphism $f=\langle f_{i}\rangle:\langle Y,\le^e_Y,\le^c_Y,\ell_{Y}\rangle\to\langle X,\le^e_X,\le^c_X,\ell_{X}\rangle$ with $\pi_{i}\circ f=f_{i}$ for $i\in I$. In the category $\mathbf{OCs}$, such products exist and may be described as:

\[\prod_{i\in I}\langle X_{i},\le^e_{X_{i}},\le^c_{X_{i}},\ell_{X_{i}}\rangle=\{t:I\to\bigcup_{i\in I}\langle X_{i},\le^e_{X_{i}},\le^c_{X_{i}},\ell_{X_{i}}\rangle|\forall i\in I.t(i)\in\langle X_{i},\le^e_{X_{i}},\le^c_{X_{i}},\ell_{X_{i}}\rangle\}\] 

\runhead{Coproducts}

For lmsets $\langle X,\le^e_X,\le^c_X,\ell_{X}\rangle$ and $\langle Y,\le^e_Y,\le^c_Y,\ell_{Y}\rangle$, their coproducts written as $\langle X,\le^e_X,\le^c_X,\ell_{X}\rangle+\langle Y,\le^e_Y,\le^c_Y,\ell_{Y}\rangle$, is defined as:

\[\langle X,\le^e_X,\le^c_X,\ell_{X}\rangle+\langle Y,\le^e_Y,\le^c_Y,\ell_{Y}\rangle=\{(x,1)|x\in \langle X,\le^e_X,\le^c_X,\ell_{X}\rangle\}\cup\{(y,2)|y\in \langle Y,\le^e_Y,\le^c_Y,\ell_{Y}\rangle\}\]

where $1$ and $2$ force the above union to be disjoint.

The coproduct operation $+$ is also a functor $\mathbf{OCs}\times\mathbf{OCs}\to\mathbf{OCs}$, actually, it is the dual of product $\times$.

\begin{definition}[Coproduct]\label{definitionofcoproduct}
Let us take two objects $\langle X,\le^e_X,\le^c_X,\ell_X\rangle $ and $\langle Y,\le^e_Y,\le^c_Y,\ell_Y\rangle $ in $\mathbf{OCs}$. The coproduct of these two objects is a new object $\langle X,\le^e_X,\le^c_X,\ell_{X}\rangle+\langle Y,\le^e_Y,\le^c_Y,\ell_{Y}\rangle$ with two coprojection morphisms:

\[\begin{tikzcd}[row sep=huge, column sep=huge]
\langle X,\le^e_X,\le^c_X,\ell_{X}\rangle \arrow[r, "\kappa_1"] & \langle X,\le^e_X,\le^c_X,\ell_{X}\rangle+\langle Y,\le^e_Y,\le^c_Y,\ell_{Y}\rangle & \langle Y,\le^e_Y,\le^c_Y,\ell_{Y}\rangle \arrow[l, "\kappa_2"']
\end{tikzcd}\]

For functions $f:\langle X,\le^e_X,\le^c_X,\ell_{X}\rangle\to\langle Z,\le^e_Z,\le^c_Z,\ell_{Z}\rangle$ and $g:\langle Y,\le^e_Y,\le^c_Y,\ell_{Y}\rangle\to\langle Z,\le^e_Z,\le^c_Z,\ell_{Z}\rangle$, there is a unique copairing function $[f,g]:\langle X,\le^e_X,\le^c_X,\ell_{X}\rangle+\langle Y,\le^e_Y,\le^c_Y,\ell_{Y}\rangle\to\langle Z,\le^e_Z,\le^c_Z,\ell_{Z}\rangle$ with:

\[[f,g](w)=\begin{cases}
             f(x), & \mbox{if } w=(x,1); \\
             g(y), & \mbox{if } w=(y,2).
           \end{cases}\]
           
where $x\in\langle X,\le^e_X,\le^c_X,\ell_{X}\rangle$ and $y\in\langle Y,\le^e_Y,\le^c_Y,\ell_{Y}\rangle$.

And the following equations hold:

\[[f,g]\circ\kappa_1=f\qquad [\kappa_1,\kappa_2]=\mathrm{id}_{\langle X,\le^e_X,\le^c_X,\ell_{X}\rangle+\langle Y,\le^e_Y,\le^c_Y,\ell_{Y}\rangle}\]
\[[f,g]\circ\kappa_2=g\qquad h\circ[f,g]=[h\circ f,h\circ g]\]

where $h:\langle Z,\le^e_Z,\le^c_Z,\ell_{Z}\rangle\to\langle W,\le^e_W,\le^c_W,\ell_{W}\rangle$.

For the above morphisms $f$ and $g$, there is a unique pairing morphism $[f,g]$ defined as above, making the following diagram commute:

\[\begin{tikzcd}[row sep=huge, column sep=huge]
\langle X,\le^e_X,\le^c_X,\ell_{X}\rangle \arrow[r, "\kappa_1"]\arrow[rd,"f"'] & \langle X,\le^e_X,\le^c_X,\ell_{X}\rangle+\langle Y,\le^e_Y,\le^c_Y,\ell_{Y}\rangle \arrow[dashed,d,"\ensuremath{[f,g]}"description] & \langle Y,\le^e_Y,\le^c_Y,\ell_{Y}\rangle \arrow[l, "\kappa_2"']\arrow[ld,"g"]\\
&\langle Z,\le^e_Z,\le^c_Z,\ell_{Z}\rangle&
\end{tikzcd}\]
\end{definition}

For functions $f:\langle X,\le^e_X,\le^c_X,\ell_{X}\rangle\to\langle X',\le^e_{X'},\le^c_{X'},\ell_{X'}\rangle$ and $g:\langle Y,\le^e_Y,\le^c_Y,\ell_{Y}\rangle\to\langle Y',\le^e_{Y'},\le^c_{Y'},\ell_{Y'}\rangle$, the coproduct of functions $f+ g$ is defined as:

\[(f+g)(w)=\begin{cases}
             (f(x),1), & \mbox{if } w=(x,1); \\
             (g(y),2), & \mbox{if } w=(y,2).
           \end{cases}\]
           
where $x\in\langle X,\le^e_X,\le^c_X,\ell_{X}\rangle$ and $y\in\langle Y,\le^e_Y,\le^c_Y,\ell_{Y}\rangle$, such that the following equations hold:

\[f+ g=\langle \kappa_1\circ f,\kappa_2\circ g\rangle\qquad \mathrm{id}_{\langle X,\le^e_X,\le^c_X,\ell_{X}\rangle}+\mathrm{id}_{\langle Y,\le^e_Y,\le^c_Y,\ell_{Y}\rangle}=\mathrm{id}_{\langle X,\le^e_X,\le^c_X,\ell_{X}\rangle+\langle Y,\le^e_Y,\le^c_Y,\ell_{Y}\rangle}\]
\[(f\circ h)+(g\circ k)=(f+ g)\circ(h+ k)\]

Two functions $\langle X,\le^e_X,\le^c_X,\ell_{X}\rangle\to\langle Z,\le^e_Z,\le^c_Z,\ell_{Z}\rangle$ and $\langle Y,\le^e_Y,\le^c_Y,\ell_{Y}\rangle\to\langle Z,\le^e_Z,\le^c_Z,\ell_{Z}\rangle$ can be composed into $\langle X,\le^e_X,\le^c_X,\ell_{X}\rangle+\langle Y,\le^e_Y,\le^c_Y,\ell_{Y}\rangle\to\langle Z,\le^e_Z,\le^c_Z,\ell_{Z}\rangle$ by \cref{definitionofcoproduct} and $\langle X,\le^e_X,\le^c_X,\ell_{X}\rangle+\langle Y,\le^e_Y,\le^c_Y,\ell_{Y}\rangle\to\langle Z,\le^e_Z,\le^c_Z,\ell_{Z}\rangle$ can be decomposed into $\langle X,\le^e_X,\le^c_X,\ell_{X}\rangle\to\langle Z,\le^e_Z,\le^c_Z,\ell_{Z}\rangle$ and $\langle Y,\le^e_Y,\le^c_Y,\ell_{Y}\rangle\to\langle Z,\le^e_Z,\le^c_Z,\ell_{Z}\rangle$ by use of coprojections $\kappa_1$ and $\kappa_2$, which can be expressed by the following doubled line rule working in two directions:

\[\doublefrac{\langle X,\le^e_X,\le^c_X,\ell_{X}\rangle\to\langle Z,\le^e_Z,\le^c_Z,\ell_{Z}\rangle\quad \langle Y,\le^e_Y,\le^c_Y,\ell_{Y}\rangle\to\langle Z,\le^e_Z,\le^c_Z,\ell_{Z}\rangle}{\langle X,\le^e_X,\le^c_X,\ell_{X}\rangle+\langle Y,\le^e_Y,\le^c_Y,\ell_{Y}\rangle\to\langle Z,\le^e_Z,\le^c_Z,\ell_{Z}\rangle}\] 

The $n$-fold coproduct is defined as $n\cdot\langle X,\le^e_X,\le^c_X,\ell_{X}\rangle=\langle X,\le^e_X,\le^c_X,\ell_{X}\rangle+\cdots+\langle X,\le^e_X,\le^c_X,\ell_{X}\rangle$ for a given $\langle X,\le^e_X,\le^c_X,\ell_{X}\rangle$ and $n\in\mathbb{N}$. And the empty coproduct $0\cdot\langle X,\le^e_X,\le^c_X,\ell_{X}\rangle$ is called an initial object.

\begin{definition}[Initial object]
An initial object in a category $\mathbb{C}$ is an object $0\in\mathbb{C}$, such that for each object $X\in\mathbb{C}$ there is a unique morphism $!_{X}:0\to X$ in $\mathbb{C}$.
\end{definition} 

\begin{proposition}[Initial object of $\mathbf{OCs}$]
$\mathbf{OCs}$ has an initial object.
\end{proposition}

\begin{proof}
The empty labelled occurrence multiset $0$ is initial.
\end{proof}

$\mathbf{OCs}$ has finite coproducts because it has binary coproducts and an initial object $0$. Finite coproducts can be generalized to set-indexed coproducts: given an index set $I$, a collection $(\langle X,\le^e_X,\le^c_X,\ell_{X}\rangle)_{i\in I}$ of $I$-indexed objects has $I$-indexed coproducts $\langle X,\le^e_X,\le^c_X,\ell_{X}\rangle=\coprod_{i\in I}\langle X_{i},\le^e_{X_{i}},\le^c_{X_{i}},\ell_{X_{i}}\rangle$ with coprojections $\kappa_{i}:\langle X_{i},\le^e_{X_{i}},\le^c_{X_{i}},\ell_{X_{i}}\rangle\to\langle X,\le^e_X,\le^c_X,\ell_{X}\rangle$ for $i\in I$, which are universal as in \cref{definitionofcoproduct}: for an arbitrary object $\langle Y,\le^e_Y,\le^c_Y,\ell_{Y}\rangle$ and an $I$-indexed collection $f_{i}:\langle X_{i},\le^e_{X_{i}},\le^c_{X_{i}},\ell_{X_{i}}\rangle\to\langle Y,\le^e_Y,\le^c_Y,\ell_{Y}\rangle$ of morphisms there is a unique morphism $f=[f_{i}]:\langle X,\le^e_X,\le^c_X,\ell_{X}\rangle\to\langle Y,\le^e_Y,\le^c_Y,\ell_{Y}\rangle$ with $f\circ\kappa_{i}=f_{i}$ for $i\in I$. In the category $\mathbf{OCs}$, such coproducts are disjoint unions and may be described as:

\[\coprod_{i\in I}\langle X_{i},\le^e_{X_{i}},\le^c_{X_{i}},\ell_{X_{i}}\rangle=\{(i,x):i\in I \wedge x\in\langle X_{i},\le^e_{X_{i}},\le^c_{X_{i}},\ell_{X_{i}}\rangle\}\] 

\runhead{Exponents}

For lmsets $\langle X,\le^e_X,\le^c_X,\ell_{X}\rangle$ and $\langle Y,\le^e_Y,\le^c_Y,\ell_{Y}\rangle$, their exponents written $\langle Y,\le^e_Y,\le^c_Y,\ell_{Y}\rangle^{\langle X,\le^e_X,\le^c_X,\ell_{X}\rangle}$, is defined as:

\[\langle Y,\le^e_Y,\le^c_Y,\ell_{Y}\rangle^{\langle X,\le^e_X,\le^c_X,\ell_{X}\rangle}=\{f|f\mbox{ is a total function }\langle X,\le^e_X,\le^c_X,\ell_{X}\rangle\to \langle Y,\le^e_Y,\le^c_Y,\ell_{Y}\rangle\}\]

The exponent operation $\langle Y,\le^e_Y,\le^c_Y,\ell_{Y}\rangle^{\langle X,\le^e_X,\le^c_X,\ell_{X}\rangle}$ is a functor $\mathbf{OCs}^{\mathrm{op}}\times\mathbf{OCs}\to\mathbf{OCs}$, where $\mathbf{OCs}^{\mathrm{op}}$ is the opposite category of $\mathbf{OCs}$.

\begin{definition}[Exponent]\label{definitionofexponent}
Let us take two objects $\langle X,\le^e_X,\le^c_X,\ell_X\rangle $ and $\langle Y,\le^e_Y,\le^c_Y,\ell_Y\rangle $ in $\mathbf{OCs}$. The exponent of these two objects is a new object $\langle Y,\le^e_Y,\le^c_Y,\ell_{Y}\rangle^{\langle X,\le^e_X,\le^c_X,\ell_{X}\rangle}$ with an evaluation morphism:

\[\begin{tikzcd}[row sep=huge, column sep=huge]
\langle Y,\le^e_Y,\le^c_Y,\ell_{Y}\rangle^{\langle X,\le^e_X,\le^c_X,\ell_{X}\rangle}\times\langle X,\le^e_X,\le^c_X,\ell_{X}\rangle \arrow[r, "\ensuremath{\mathrm{ev}}"] & \langle Y,\le^e_Y,\le^c_Y,\ell_{Y}\rangle 
\end{tikzcd}\]

For each function $f:\langle Z,\le^e_Z,\le^c_Z,\ell_{Z}\rangle\times\langle X,\le^e_X,\le^c_X,\ell_{X}\rangle\to\langle Y,\le^e_Y,\le^c_Y,\ell_{Y}\rangle$, there is a unique abstraction morphism $\Lambda(f):\langle X,\ell_{Z}\rangle\to\langle Y,\le^e_Y,\le^c_Y,\ell_{Y}\rangle^{\langle X,\le^e_X,\le^c_X,\ell_{X}\rangle}$, such that the following equations hold:

\[\mathrm{ev}\circ(\Lambda(f)\times\mathrm{id}_{\langle X,\le^e_X,\le^c_X,\ell_{X}\rangle})=f\qquad \Lambda(\mathrm{ev})=\mathrm{id}_{\langle Y,\le^e_Y,\le^c_Y,\ell_{Y}\rangle^{\langle X,\le^e_X,\le^c_X,\ell_{X}\rangle}}\]
\[\Lambda(f)\circ h=\Lambda(f\circ(h\times\mathrm{id}_{\langle X,\le^e_X,\le^c_X,\ell_{X}\rangle}))\]

For the above morphism $f$, there is a unique abstraction morphism $\Lambda(f)$ defined as above, making the following diagram commute:

\[\begin{tikzcd}[row sep=huge, column sep=huge]
\langle Y,\le^e_Y,\le^c_Y,\ell_{Y}\rangle^{\langle X,\le^e_X,\le^c_X,\ell_{X}\rangle}\times\langle X,\le^e_X,\le^c_X,\ell_{X}\rangle \arrow[r, "\ensuremath{\mathrm{ev}}"]  & \langle Y,\le^e_Y,\le^c_Y,\ell_{Y}\rangle \\
\langle Z,\le^e_Z,\le^c_Z,\ell_{Z}\rangle\times\langle X,\le^e_X,\le^c_X,\ell_{X}\rangle \arrow[dashed,u,"\ensuremath{\Lambda(f)\times\mathrm{id}_{\langle X,\le^e_X,\le^c_X,\ell_{X}\rangle}}"description] \arrow[ru,"f"']&
\end{tikzcd}\]
\end{definition}

For two maps $k:\langle X,\le^e_X,\le^c_X,\ell_{X}\rangle\to\langle U,\le^e_U,\le^c_U,\ell_{U}\rangle$ in $\mathbf{OCs}^{\mathrm{op}}$ and $h:\langle Y,\le^e_Y,\le^c_Y,\ell_{Y}\rangle\to\langle V,\le^e_V,\le^c_V,\ell_{V}\rangle$ in $\mathbf{OCs}$, the exponent of functions $h^{k}$ is defined as:

\[h^{k}(f)=h\circ f\circ k\]

such that the following equations hold:

\[(\mathrm{id}^{\mathrm{id}}(f))=\mathrm{id}\circ f\circ\mathrm{id}=f\]
\[\begin{aligned}
(h_{2}^{k_{2}}\circ h_{1}^{k_{1}})(f)&=(h_{2}^{k_{2}}(h_{1}^{k_{1}}(f))\\
&=h_{2}^{k_{2}}(h_{1}\circ f\circ k_{1})\\
&=h_{2}\circ h_{1}\circ f\circ k_{1}\circ k_{2}\\
&=(h_{2}\circ h_{1})^{k_{1}\circ k_{2}}(f)\\
&=(h_{2}\circ_{\mathbb{C}} h_{1})^{k_{1}\circ_{\mathbb{C}^{\mathrm{op}}} k_{2}}(f)
\end{aligned}\]

The exponent, also called currying, can be expressed by the following doubled line rule:

\[\doublefrac{\langle Z,\le^e_Z,\le^c_Z,\ell_{Z}\rangle\times\langle X,\le^e_X,\le^c_X,\ell_{X}\rangle\to\langle Y,\le^e_Y,\le^c_Y,\ell_{Y}\rangle}{\langle Z,\le^e_Z,\le^c_Z,\ell_{Z}\rangle\to\langle Y,\le^e_Y,\le^c_Y,\ell_{Y}\rangle^{\langle X,\le^e_X,\le^c_X,\ell_{X}\rangle}}\] 

$\mathbf{OCs}$ is also a $\mathbf{BiCCC}$.

\runhead{Merges}

For lmsets $\langle X,\le^e_X,\le^c_X,\ell_{X}\rangle$ and $\langle Y,\le^e_Y,\le^c_Y,\ell_{Y}\rangle$, their merges written as $\langle X,\le^e_X,\le^c_X,\ell_{X}\rangle\mid \langle Y,\le^e_Y,\le^c_Y,\ell_{Y}\rangle$, is defined as:

\[\langle X,\le^e_X,\le^c_X,\ell_{X}\rangle\mid \langle Y,\le^e_Y,\le^c_Y,\ell_{Y}\rangle=\{\gamma(x,y)|x\in \langle X,\le^e_X,\le^c_X,\ell_{X}\rangle\wedge y\in \langle Y,\le^e_Y,\le^c_Y,\ell_{Y}\rangle\}\]

The merge operation $\mid$ is a functor $\mathbf{OCs}\times\mathbf{OCs}\to\mathbf{OCs}$.

\begin{definition}[Merge]\label{definitionofmerge}
Let us take two objects $\langle X,\le^e_X,\le^c_X,\ell_X\rangle$ and $\langle Y,\le^e_Y,\le^c_Y,\ell_Y\rangle$ in $\mathbf{OCs}$. The merge of these two objects is a new object $\langle X,\le^e_X,\le^c_X,\ell_{X}\rangle\mid\langle Y,\le^e_Y,\le^c_Y,\ell_{Y}\rangle$ with two projection morphisms:

\[\begin{tikzcd}[row sep=huge, column sep=huge]
\langle X,\le^e_X,\le^c_X,\ell_{X}\rangle & \langle X,\le^e_X,\le^c_X,\ell_{X}\rangle\mid\langle Y,\le^e_Y,\le^c_Y,\ell_{Y}\rangle \arrow[l, "\iota_1"'] \arrow[r, "\iota_2"] & \langle Y,\le^e_Y,\le^c_Y,\ell_{Y}\rangle 
\end{tikzcd}\]

For functions $f:\langle Z,\le^e_Z,\le^c_Z,\ell_{Z}\rangle\to\langle X,\le^e_X,\le^c_X,\ell_{X}\rangle$ and $g:\langle Z,\le^e_Z,\le^c_Z,\ell_{Z}\rangle\to\langle Y,\le^e_Y,\le^c_Y,\ell_{Y}\rangle$, there is a unique pairing function $\{f,g\}:\langle Z,\le^e_Z,\le^c_Z,\ell_{Z}\rangle\to\langle X,\le^e_X,\le^c_X,\ell_{X}\rangle\mid\langle Y,\le^e_Y,\le^c_Y,\ell_{Y}\rangle$ with $\{f,g\}(z)=(f(z),g(z))\in \langle X,\le^e_X,\le^c_X,\ell_{X}\rangle\mid\langle Y,\le^e_Y,\le^c_Y,\ell_{Y}\rangle$ for $z\in\langle Z,\le^e_Z,\le^c_Z,\ell_{Z}\rangle$, such that the following equations hold:

\[\iota_1\circ\{f,g\}=f\qquad \{\iota_1,\iota_2\}=\mathrm{id}_{\langle X,\le^e_X,\le^c_X,\ell_{X}\rangle\mid\langle Y,\le^e_Y,\le^c_Y,\ell_{Y}\rangle}\]
\[\iota_2\circ\{f,g\}=g\qquad \{f,g\}\circ h=\{f\circ h,g\circ h\}\]

where $h:\langle W,\le^e_W,\le^c_W,\ell_{W}\rangle\to\langle Z,\le^e_Z,\le^c_Z,\ell_{Z}\rangle$.

For the above morphisms $f$ and $g$, there is a unique pairing morphism $\{f,g\}$ defined as above, making the following diagram commute:

\[\begin{tikzcd}[row sep=huge, column sep=huge]
\langle X,\le^e_X,\le^c_X,\ell_{X}\rangle & \langle X,\le^e_X,\le^c_X,\ell_{X}\rangle\mid\langle Y,\le^e_Y,\le^c_Y,\ell_{Y}\rangle \arrow[l, "\iota_1"'] \arrow[r, "\iota_2"] & \langle Y,\le^e_Y,\le^c_Y,\ell_{Y}\rangle \\
&\langle Z,\le^e_Z,\le^c_Z,\ell_{Z}\rangle \arrow[lu,"f"] \arrow[dashed,u,"\ensuremath{\{f,g\}}"description] \arrow[ru,"g"']&
\end{tikzcd}\]
\end{definition}

For functions $f:\langle X,\le^e_X,\le^c_X,\ell_{X}\rangle\to\langle X',\le^e_{X'},\le^c_{X'},\ell_{X'}\rangle$ and $g:\langle Y,\le^e_Y,\le^c_Y,\ell_{Y}\rangle\to\langle Y',\le^e_{Y'},\le^c_{Y'},\ell_{Y'}\rangle$, the merge of functions $f\mid g$ is defined as:

\[\begin{tikzcd}[row sep=huge, column sep=huge]
\langle X,\le^e_X,\le^c_X,\ell_{X}\rangle\mid\langle Y,\le^e_Y,\le^c_Y,\ell_{Y}\rangle \arrow[r, "f\mid g"] & \langle X',\le^e_{X'},\le^c_{X'},\ell_{X'}\rangle\mid\langle Y',\le^e_{Y'},\le^c_{Y'},\ell_{Y'}\rangle 
\end{tikzcd}\] 

which is given by $\gamma(x,y)\mapsto(f(x),g(y))$, where $x\in\langle X,\le^e_X,\le^c_X,\ell_{X}\rangle$ and $y\in\langle Y,\le^e_Y,\le^c_Y,\ell_{Y}\rangle$, such that the following equations hold:

\[f\mid g=\{f\circ\iota_1,g\circ\iota_2\}\qquad \mathrm{id}_{\langle X,\le^e_X,\le^c_X,\ell_{X}\rangle}\mid\mathrm{id}_{\langle Y,\le^e_Y,\le^c_Y,\ell_{Y}\rangle}=\mathrm{id}_{\langle X,\le^e_X,\le^c_X,\ell_{X}\rangle\mid\langle Y,\le^e_Y,\le^c_Y,\ell_{Y}\rangle}\]
\[(f\circ h)\mid(g\circ k)=(f\mid g)\circ(h\mid k)\]

Two functions $\langle Z,\le^e_Z,\le^c_Z,\ell_{Z}\rangle\to\langle X,\le^e_X,\le^c_X,\ell_{X}\rangle$ and $\langle Z,\le^e_Z,\le^c_Z,\ell_{Z}\rangle\to\langle Y,\le^e_Y,\le^c_Y,\ell_{Y}\rangle$ can be composed into $\langle Z,\le^e_Z,\le^c_Z,\ell_{Z}\rangle\to\langle X,\le^e_X,\le^c_X,\ell_{X}\rangle\mid\langle Y,\le^e_Y,\le^c_Y,\ell_{Y}\rangle$ by \cref{definitionofmerge} and $\langle Z,\le^e_Z,\le^c_Z,\ell_{Z}\rangle\to\langle X,\le^e_X,\le^c_X,\ell_{X}\rangle\mid\langle Y,\le^e_Y,\le^c_Y,\ell_{Y}\rangle$ can be decomposed into $\langle Z,\le^e_Z,\le^c_Z,\ell_{Z}\rangle\to\langle X,\le^e_X,\le^c_X,\ell_{X}\rangle$ and $\langle Z,\le^e_Z,\le^c_Z,\ell_{Z}\rangle\to\langle Y,\le^e_Y,\le^c_Y,\ell_{Y}\rangle$ by use of projections $\iota_1$ and $\iota_2$, which can be expressed by the following doubled line rule working in two directions:

\[\doublefrac{\langle Z,\le^e_Z,\le^c_Z,\ell_{Z}\rangle\to\langle X,\le^e_X,\le^c_X,\ell_{X}\rangle\quad \langle Z,\le^e_Z,\le^c_Z,\ell_{Z}\rangle\to\langle Y,\le^e_Y,\le^c_Y,\ell_{Y}\rangle}{\langle Z,\le^e_Z,\le^c_Z,\ell_{Z}\rangle\to\langle X,\le^e_X,\le^c_X,\ell_{X}\rangle\mid\langle Y,\le^e_Y,\le^c_Y,\ell_{Y}\rangle}\] 

The $n$-fold merge is defined as $\langle X,\le^e_X,\le^c_X,\ell_{X}\rangle^{n,\mid}=\langle X,\le^e_X,\le^c_X,\ell_{X}\rangle\mid\cdots\mid\langle X,\le^e_X,\le^c_X,\ell_{X}\rangle$ for a given $\langle X,\le^e_X,\le^c_X,\ell_{X}\rangle$ and $n\in\mathbb{N}$. And the empty merge $\langle X,\le^e_X,\le^c_X,\ell_{X}\rangle^{0,\mid}$ is called a final or terminal object.

\begin{definition}[Final object]
A final object in a category $\mathbb{C}$ is an object $1\in\mathbb{C}$, such that for each object $X\in\mathbb{C}$ there is a unique morphism $!_{X}:X\to 1$ in $\mathbb{C}$.
\end{definition} 

\begin{proposition}[Final object of $\mathbf{OCs}$]
$\mathbf{OCs}$ has a final object.
\end{proposition}

\begin{proof}
Any singleton labelled occurrence multiset is final.
\end{proof}

$\mathbf{OCs}$ has finite merges because it has binary merges and a final object $1$. Finite merges can be generalized to set-indexed merges: given an index set $I$, a collection $(\langle X,\le^e_X,\le^c_X,\ell_{X}\rangle)_{i\in I}$ of $I$-indexed objects has $I$-indexed merges $\langle X,\le^e_X,\le^c_X,\ell_{X}\rangle=\prod_{i\in I}^{\mid}\langle X_{i},\le^e_{X_{i}},\le^c_{X_{i}},\ell_{X_{i}}\rangle$ with projections $\iota_{i}:\langle X,\le^e_X,\le^c_X,\ell_{X}\rangle\to\langle X_{i},\le^e_{X_{i}},\le^c_{X_{i}},\ell_{X_{i}}\rangle$ for $i\in I$, which are universal as in \cref{definitionofmerge}: for an arbitrary object $\langle Y,\le^e_Y,\le^c_Y,\ell_{Y}\rangle$ and an $I$-indexed collection $f_{i}:\langle Y,\le^e_Y,\le^c_Y,\ell_{Y}\rangle\to\langle X_{i},\le^e_{X_{i}},\le^c_{X_{i}},\ell_{X_{i}}\rangle$ of morphisms there is a unique morphism $f=\{f_{i}\}: \langle Y,\le^e_Y,\le^c_Y,\ell_{Y}\rangle\to\langle X,\le^e_X,\le^c_X,\ell_{X}\rangle$ with $\iota_{i}\circ f=f_{i}$ for $i\in I$. In the category $\mathbf{OCs}$, such merges exist and may be described as:

\[\prod_{i\in I}^{\mid}\langle X_{i},\le^e_{X_{i}},\le^c_{X_{i}},\ell_{X_{i}}\rangle=\{t:I\to\bigcup_{i\in I}\langle X_{i},\le^e_{X_{i}},\le^c_{X_{i}},\ell_{X_{i}}\rangle|\forall i\in I.t(i)\in\langle X_{i},\le^e_{X_{i}},\le^c_{X_{i}},\ell_{X_{i}}\rangle\}\] 

\runhead{Powersets}

For an lmset $\langle X,\le^e_X,\le^c_X,\ell_{X}\rangle$ we write the powerset of $\langle X,\le^e_X,\le^c_X,\ell_{X}\rangle$ as $\mathcal{P}(\langle X,\le^e_X,\le^c_X,\ell_{X}\rangle)=\{\langle U,\le^e_U,\le^c_U,\ell_{U}\rangle|\langle U,\le^e_U,\le^c_U,\ell_{U}\rangle\subseteq \langle X,\le^e_X,\le^c_X,\ell_{X}\rangle\}$ for the set of all subsets of $\langle X,\le^e_X,\le^c_X,\ell_{X}\rangle$. The powerset $\mathcal{P}(\langle X,\le^e_X,\le^c_X,\ell_{X}\rangle)$ is naturally ordered by set inclusion $\subseteq$ and yields a poset $\langle \mathcal{P}(\langle X,\le^e_X,\le^c_X,\ell_{X}\rangle),\subseteq\rangle$. By defining meets by intersection $\bigcap_{i\in I}\langle U_{i},\le^e_{U_{i}},\le^c_{U_{i}},\ell_{U_{i}}\rangle=\{x\in\langle X,\le^e_X,\le^c_X,\ell_{X}\rangle|\forall i\in I.x\in\langle U_{i},\le^e_{U_{i}},\le^c_{U_{i}},\ell_{U_{i}}\rangle\}$, joins by unions $\bigcup_{i\in I}\langle U_{i},\le^e_{U_{i}},\le^c_{U_{i}},\ell_{U_{i}}\rangle=\{x\in\langle X,\le^e_X,\le^c_X,\ell_{X}\rangle|\exists i\in I.x\in\langle U_{i},\le^e_{U_{i}},\le^c_{U_{i}},\ell_{U_{i}}\rangle\}$ and negation by complement $\neg\langle U,\le^e_U,\le^c_U,\ell_{U}\rangle=\{x\in\langle X,\le^e_X,\le^c_X,\ell_{X}\rangle|x\notin\langle U,\le^e_U,\le^c_U,\ell_{U}\rangle\}$, and truth predicate $\top_{\langle X,\le^e_X,\le^c_X,\ell_{X}\rangle}=(\langle X,\le^e_X,\le^c_X,\ell_{X}\rangle\subseteq\langle X,\le^e_X,\le^c_X,\ell_{X}\rangle)$ and falsity predicate $\bot_{\langle X,\le^e_X,\le^c_X,\ell_{X}\rangle}=(\emptyset\not\subseteq\langle X,\le^e_X,\le^c_X,\ell_{X}\rangle)$, the poset $\langle \mathcal{P}(\langle X,\le^e_X,\le^c_X,\ell_{X}\rangle),\subseteq\rangle$ forms a complete Boolean algebra, and thus $\{\top,\bot\}$ of Booleans is the powerset $\mathcal{P}(1)$ of the final object $1$.

A relation $R$ on lmsets $\langle X,\le^e_X,\le^c_X,\ell_{X}\rangle$ and $\langle Y,\le^e_Y,\le^c_Y,\ell_{Y}\rangle$ is a subset $R\subseteq\langle X,\le^e_X,\le^c_X,\ell_{X}\rangle\times\langle Y,\le^e_Y,\le^c_Y,\ell_{Y}\rangle$, i.e, $R\in\mathcal{P}(\langle X,\le^e_X,\le^c_X,\ell_{X}\rangle\times\langle Y,\le^e_Y,\le^c_Y,\ell_{Y}\rangle)$ is a special case of predicates. Like predicates, $\langle \mathcal{P}(\langle X,\le^e_X,\le^c_X,\ell_{X}\rangle\times\langle Y,\le^e_Y,\le^c_Y,\ell_{Y}\rangle),\subseteq\rangle$ forms again a complete Boolean algebra. Ordinarily, relations are equipped with reversal $R^{\dagger}$ for $R$ and composition $\circ$.

For endo-relations $R\subseteq\langle X,\le^e_X,\le^c_X,\ell_{X}\rangle\times\langle X,\le^e_X,\le^c_X,\ell_{X}\rangle$ on some lmset $\langle X,\le^e_X,\le^c_X,\ell_{X}\rangle$, the equality relation $\mathrm{Eq}(\langle X,\le^e_X,\le^c_X,\ell_{X}\rangle)=\{(x,y)\in \langle X,\le^e_X,\le^c_X,\ell_{X}\rangle\times\langle X,\le^e_X,\le^c_X,\ell_{X}\rangle|x=y\}=\{(x,x)|x\in \langle X,\le^e_X,\le^c_X,\ell_{X}\rangle\}$. The category $\mathbf{LMSetsRel}$ contains lmsets as objects, relations as morphisms and equality relations as identity morphisms.

The powerset operation $\langle X,\le^e_X,\le^c_X,\ell_{X}\rangle\mapsto\mathcal{P}({\langle X,\le^e_X,\le^c_X,\ell_{X}\rangle})$ is also functorial. For a function $f:\langle X,\le^e_X,\le^c_X,\ell_{X}\rangle\to \langle Y,\le^e_Y,\le^c_Y,\ell_{Y}\rangle$, there is a function $\mathcal{P}(f):\mathcal{P}(\langle X,\le^e_X,\le^c_X,\ell_{X}\rangle)\to\mathcal{P}(\langle Y,\le^e_Y,\le^c_Y,\ell_{Y}\rangle)$, there are several ways to give this functor:

\begin{enumerate}
  \item The direct image way: for $\langle U,\le^e_U,\le^c_U,\ell_{U}\rangle\subseteq \langle X,\le^e_X,\le^c_X,\ell_{X}\rangle$,
  \[\begin{aligned}
  \mathcal{P}(f)(\langle U,\le^e_U,\le^c_U,\ell_{U}\rangle)&=\{f(x)|x\in\langle U,\le^e_U,\le^c_U,\ell_{U}\rangle\}\\
  &=\{y\in\langle Y,\le^e_Y,\le^c_Y,\ell_{Y}\rangle|\exists x\in\langle X,\le^e_X,\le^c_X,\ell_{X}\rangle.f(x)=y\wedge x\in \langle U,\le^e_U,\le^c_U,\ell_{U}\rangle\}
  \end{aligned}\]
  The direct image can also be described by $f[\langle U,\le^e_U,\le^c_U,\ell_{U}\rangle]$ or $\coprod_{f}(\langle U,\le^e_U,\le^c_U,\ell_{U}\rangle)$. The direct image way describe the powerset as a functor $\mathcal{P}:\mathbf{OCs}\to\mathbf{OCs}$. For finite powerset $\mathcal{P}_{\mathrm{fin}}(\langle X,\le^e_X,\le^c_X,\ell_{X}\rangle)=\{\langle U,\le^e_U,\le^c_U,\ell_{U}\rangle\in\mathcal{P}(\langle X,\le^e_X,\le^c_X,\ell_{X}\rangle)|\langle U,\le^e_U,\le^c_U,\ell_{U}\rangle\mbox{ is finite}\}$, which also forms a functor $\mathbf{OCs}\to\mathbf{OCs}$.
  \item The inverse image way: the inverse image way describes the powerset as a functor $\mathcal{P}:\mathbf{OCs}^{\mathrm{op}}\to\mathbf{OCs}$. For a function $f:\langle X,\le^e_X,\le^c_X,\ell_{X}\rangle\to \langle Y,\le^e_Y,\le^c_Y,\ell_{Y}\rangle$, there is a function $f^{-1}:\mathcal{P}(\langle Y,\le^e_Y,\le^c_Y,\ell_{Y}\rangle)\to\mathcal{P}(\langle X,\le^e_X,\le^c_X,\ell_{X}\rangle)$, for $\langle U,\le^e_U,\le^c_U,\ell_{U}\rangle\subseteq \langle X,\le^e_X,\le^c_X,\ell_{X}\rangle$,
  \[f^{-1}(\langle U,\le^e_U,\le^c_U,\ell_{U}\rangle)=\{x|f(x)\in\langle U,\le^e_U,\le^c_U,\ell_{U}\rangle\}\] 
\end{enumerate}

The correspondence of the direct image and inverse image can be reflected by the following rule:

\[\doublefrac{\coprod_{f}(\langle U,\le^e_U,\le^c_U,\ell_{U}\rangle)\subseteq\langle V,\le^e_V,\le^c_V,\ell_{V}\rangle}{\langle U,\le^e_U,\le^c_U,\ell_{U}\rangle\subseteq f^{-1}(\langle V,\le^e_V,\le^c_V,\ell_{V}\rangle)}\]

And there is a correspondence for powerset, which can be expressed by the following rule:

\[\doublefrac{\langle X,\le^e_X,\le^c_X,\ell_{X}\rangle\to\mathcal{P}(\langle Y,\le^e_Y,\le^c_Y,\ell_{Y}\rangle)}{\mbox{relations}\subseteq \langle Y,\le^e_Y,\le^c_Y,\ell_{Y}\rangle\times\langle X,\le^e_X,\le^c_X,\ell_{X}\rangle}\]

\runhead{Injections and Surjections}

For lmsets $\langle X,\le^e_X,\le^c_X,\ell_{X}\rangle$ and $\langle Y,\le^e_Y,\le^c_Y,\ell_{Y}\rangle$, an injection (a monomorphism (mono) or a monic map) is a map such that if $f(x)=f(x')$ then $x=x'$ for $x,x'\in\langle X,\le^e_X,\le^c_X,\ell_{X}\rangle$. For an injection, in case $\langle X,\le^e_X,\le^c_X,\ell_{X}\rangle\subseteq\langle Y,\le^e_Y,\le^c_Y,\ell_{Y}\rangle$, we write $f:\langle X,\le^e_X,\le^c_X,\ell_{X}\rangle\hookrightarrow\langle Y,\le^e_Y,\le^c_Y,\ell_{Y}\rangle$ or $f:\langle X,\le^e_X,\le^c_X,\ell_{X}\rangle\rightarrowtail\langle Y,\le^e_Y,\le^c_Y,\ell_{Y}\rangle$. A surjection (an epimorphism (epi) or epic map) is a map written as $f:\langle X,\le^e_X,\le^c_X,\ell_{X}\rangle\twoheadrightarrow\langle Y,\le^e_Y,\le^c_Y,\ell_{Y}\rangle$ such that for each $y\in\langle Y,\le^e_Y,\le^c_Y,\ell_{Y}\rangle$ there is an $x\in \langle X,\le^e_X,\le^c_X,\ell_{X}\rangle$ with $f(x)=y$.

Every surjection $f:\langle X,\le^e_X,\le^c_X,\ell_{X}\rangle\twoheadrightarrow\langle Y,\le^e_Y,\le^c_Y,\ell_{Y}\rangle$ has a section (splitting), a function $s:\langle Y,\le^e_Y,\le^c_Y,\ell_{Y}\rangle\to\langle X,\le^e_X,\le^c_X,\ell_{X}\rangle$ in the reverse direction with $f\circ s=\mathrm{id}$. This $s$ thus chooses an element $s(y)$ in the non-empty set $f^{-1}(y)=\{x\in \langle X,\le^e_X,\le^c_X,\ell_{X}\rangle|f(x)=y\}$, and such a section is an injection.

Also every injection $f:\langle X,\le^e_X,\le^c_X,\ell_{X}\rangle\rightarrowtail\langle Y,\le^e_Y,\le^c_Y,\ell_{Y}\rangle$ with $\langle X,\le^e_X,\le^c_X,\ell_{X}\rangle\neq\emptyset$, there is a function $g:\langle Y,\le^e_Y,\le^c_Y,\ell_{Y}\rangle\to\langle X,\le^e_X,\le^c_X,\ell_{X}\rangle$ with $g\circ f=\mathrm{id}$, where $x_0\in \langle X,\le^e_X,\le^c_X,\ell_{X}\rangle$:

\[g(y)=\begin{cases}
         x, & \mbox{if there is an element }x \mbox{ with } f(x)=y;\\
         x_0, & \mbox{otherwise}.
       \end{cases}\]

\begin{lemma}[Injection and surjection]
Let $F:\mathbf{OCs}\to\mathbf{OCs}$ be an arbitrary functor.

\begin{enumerate}
  \item If $f:\langle X,\le^e_X,\le^c_X,\ell_{X}\rangle\to\langle Y,\le^e_Y,\le^c_Y,\ell_{Y}\rangle$ is surjective, then so is $F(f):F(\langle X,\le^e_X,\le^c_X,\ell_{X}\rangle)\to F(\langle Y,\le^e_Y,\le^c_Y,\ell_{Y}\rangle)$.
  \item If $f:\langle X,\le^e_X,\le^c_X,\ell_{X}\rangle\to\langle Y,\le^e_Y,\le^c_Y,\ell_{Y}\rangle$ is injective with $\langle X,\le^e_X,\le^c_X,\ell_{X}\rangle\neq\emptyset$, then so is $F(f):F(\langle X,\le^e_X,\le^c_X,\ell_{X}\rangle)\to F(\langle Y,\le^e_Y,\le^c_Y,\ell_{Y}\rangle)$.
\end{enumerate}
\end{lemma}

\begin{definition}[Collections of polynomial functors]
We define three collections of functors $\mathbf{OCs}\to\mathbf{OCs}$, $\mathbf{SPF}$, $\mathbf{EPF}$ and $\mathbf{KPF}$, for simple, exponent and Kripke polynomial functors:

\begin{enumerate}
  \item $\mathbf{SPF}$ is the least class of functors $\mathbf{OCs}\to\mathbf{OCs}$ satisfying:
  \begin{enumerate}
    \item The identity functor $\mathbf{OCs}\to\mathbf{OCs}$ is in $\mathbf{SPF}$.
    \item For each lmset $\langle A,\le^e_A,\le^c_A,\ell_{A}\rangle$, the constant functor $A:\mathbf{OCs}\to\mathbf{OCs}$ is in $\mathbf{SPF}$, which maps every lmset $\langle X,\le^e_X,\le^c_X,\ell_{X}\rangle$ to $\langle A,\le^e_A,\le^c_A,\ell_{A}\rangle$ and every function $f$ to the identity $\mathrm{id}_{\langle A,\le^e_A,\le^c_A,\ell_{A}\rangle}$ on $\langle A,\le^e_A,\le^c_A,\ell_{A}\rangle$.
    \item If functors $F,G\in\mathbf{SPF}$, then $F\times G\in\mathbf{SPF}$, as $\langle X,\le^e_X,\le^c_X,\ell_{X}\rangle\mapsto F(\langle X,\le^e_X,\le^c_X,\ell_{X}\rangle)\times F(\langle X,\le^e_X,\le^c_X,\ell_{X}\rangle)$ and for functions $f\mapsto F(f)\times G(f)$.
    \item For non-empty index set $I$, an $I$-indexed collection of functors $F_{i}\in\mathbf{SPF}$ for $i\in I$, then $\coprod_{i\in I}F_{i}\in\mathbf{SPF}$, as $\langle X,\le^e_X,\le^c_X,\ell_{X}\rangle\mapsto\coprod_{i\in I}F_{i}(\langle X,\le^e_X,\le^c_X,\ell_{X}\rangle)$ with $\coprod_{i\in I}F_{i}(\langle X,\le^e_X,\le^c_X,\ell_{X}\rangle)=\{(i,u)|i\in I\wedge u\in F_{i}(\langle X,\le^e_X,\le^c_X,\ell_{X}\rangle)\}$ and for a function $f:\langle X,\le^e_X,\le^c_X,\ell_{X}\rangle\to\langle Y,\le^e_Y,\le^c_Y,\ell_{Y}\rangle$ to the mapping $(i,u)\mapsto(i,F_{i}(f)(u))$.
    \item If functors $F,G\in\mathbf{SPF}$, then $F\mid G\in\mathbf{SPF}$, as $\langle X,\le^e_X,\le^c_X,\ell_{X}\rangle\mapsto F(\langle X,\le^e_X,\le^c_X,\ell_{X}\rangle)\mid F(\langle X,\le^e_X,\le^c_X,\ell_{X}\rangle)$ and for functions $f\mapsto F(f)\mid G(f)$.
  \end{enumerate}
  \item $\mathbf{EPF}\supset\mathbf{SPF}$, and additionally, for each lmset $\langle A,\le^e_A,\le^c_A,\ell_{A}\rangle$, if $F\in\mathbf{EPF}$, then so is the constant exponent $F^{\langle A,\le^e_A,\le^c_A,\ell_{A}\rangle}$ as $\langle X,\le^e_X,\le^c_X,\ell_{X}\rangle\mapsto F(\langle X,\le^e_X,\le^c_X,\ell_{X}\rangle)^{\langle A,\le^e_A,\le^c_A,\ell_{A}\rangle}$. It sends a function $f:\langle X,\le^e_X,\le^c_X,\ell_{X}\rangle\to\langle Y,\le^e_Y,\le^c_Y,\ell_{Y}\rangle$ to the function $F(f)^{\langle A,\le^e_A,\le^c_A,\ell_{A}\rangle}=F(f)^{\mathrm{id}_{\langle A,\le^e_A,\le^c_A,\ell_{A}\rangle}}$, which maps $h:\langle A,\le^e_A,\le^c_A,\ell_{A}\rangle\to F(\langle X,\le^e_X,\le^c_X,\ell_{X}\rangle)$ to $F(f)\circ h:\langle A,\le^e_A,\le^c_A,\ell_{A}\rangle\to F(\langle Y,\le^e_Y,\le^c_Y,\ell_{Y}\rangle)$.
  \item $\mathbf{KPF}\supset\mathbf{EPF}$, and additionally, 
  \begin{enumerate}
    \item If $F\in\mathbf{KPF}$, then so is the powerset $\mathcal{P}(F)\in\mathbf{KPF}$, as $\langle X,\le^e_X,\le^c_X,\ell_{X}\rangle\mapsto\mathcal{P}(F(\langle X,\le^e_X,\le^c_X,\ell_{X}\rangle))$ and functions $f\mapsto\mathcal{P}(F(f))$. 
    \item for each lmset $\langle A,\le^e_A,\le^c_A,\ell_{A}\rangle$, if $F\in\mathbf{KPF}$, then so is the powerset exponent $F^{\mathcal{P}(\langle A,\le^e_A,\le^c_A,\ell_{A}\rangle)}$ as $\langle X,\le^e_X,\le^c_X,\ell_{X}\rangle\mapsto F(\langle X,\le^e_X,\le^c_X,\ell_{X}\rangle)^{\mathcal{P}(\langle A,\le^e_A,\le^c_A,\ell_{A}\rangle})$. It sends a function $f:\langle X,\le^e_X,\le^c_X,\ell_{X}\rangle\to\langle Y,\le^e_Y,\le^c_Y,\ell_{Y}\rangle$ to the function $F(f)^{\mathcal{P}(\langle A,\le^e_A,\le^c_A,\ell_{A}\rangle)}=F(f)^{\mathcal{P}(\mathrm{id}_{\langle A,\le^e_A,\le^c_A,\ell_{A}\rangle})}$, which maps $h:\mathcal{P}(\langle A,\le^e_A,\le^c_A,\ell_{A}\rangle)\to F(\langle X,\le^e_X,\le^c_X,\ell_{X}\rangle)$ to $F(f)\circ h:\mathcal{P}(\langle A,\le^e_A,\le^c_A,\ell_{A}\rangle)\to F(\langle Y,\le^e_Y,\le^c_Y,\ell_{Y}\rangle)$.
  \end{enumerate}
   If the powersets $\mathcal{P}$ in $F$ are $\mathcal{P}_{\mathrm{fin}}$, then $F$ is called a finite $\mathbf{KPF}$.
\end{enumerate}
\end{definition}

The list functor $F^{\star}$ is given as infinite coproduct of products:

\[F^{\star}=\coprod_{n\in\mathbb{N}}F^{n}\mbox{ where }F^{n}=F\times F^{n-1}\]

If $F$ is the identity functor, then $F^{\star}$ maps an lmset $\langle X,\le^e_X,\le^c_X,\ell_{X}\rangle$ to the set of lists:

$\langle X,\le^e_X,\le^c_X,\ell_{X}\rangle^{\star}=1+\langle X,\le^e_X,\le^c_X,\ell_{X}\rangle+(\langle X,\le^e_X,\le^c_X,\ell_{X}\rangle\times\langle X,\le^e_X,\le^c_X,\ell_{X}\rangle)+(\langle X,\le^e_X,\le^c_X,\ell_{X}\rangle\times\langle X,\le^e_X,\le^c_X,\ell_{X}\rangle\times\langle X,\le^e_X,\le^c_X,\ell_{X}\rangle)+\cdots$

\begin{theorem}
Each finite Kripke polynomail functor $\mathbf{OCs}\to\mathbf{OCs}$ has a final coalgebra.
\end{theorem}

\begin{proposition}
An arbitrary initial algebra $\alpha:F(\langle A,\le^e_A,\le^c_A,\ell_{A}\rangle)\xrightarrow{\cong}\langle A,\le^e_A,\le^c_A,\ell_{A}\rangle$ satisfying the following recursion property: for each map $h:F(\langle X,\le^e_X,\le^c_X,\ell_{X}\rangle\times\langle A,\le^e_A,\le^c_A,\ell_{A}\rangle)\to \langle X,\le^e_X,\le^c_X,\ell_{X}\rangle$, there is a unique map $f:\langle A,\le^e_A,\le^c_A,\ell_{A}\rangle\to \langle X,\le^e_X,\le^c_X,\ell_{X}\rangle$ making the following diagram commute:

\begin{center}
\begin{tikzcd}[row sep=huge, column sep=huge]
F(\langle A,\le^e_A,\le^c_A,\ell_{A}\rangle) \arrow[dashed,r,"\ensuremath{F(\langle f,\mathrm{id}\rangle)}"] \arrow[d,"\ensuremath{\alpha}"',"\cong"] & F(\langle X,\le^e_X,\le^c_X,\ell_{X}\rangle\times\langle A,\le^e_A,\le^c_A,\ell_{A}\rangle) \arrow[d,"\ensuremath{h}"] \\
\langle A,\le^e_A,\le^c_A,\ell_{A}\rangle \arrow[dashed,r,"\ensuremath{f}"']  & \langle X,\le^e_X,\le^c_X,\ell_{X}\rangle 
\end{tikzcd}
\end{center}
\end{proposition}

\subsection{Algebra for True Concurrency $APTC$}

\subsubsection{Step Bisimulation}

\begin{definition}[Relation lifting]
Let $F:\mathbf{OCs}\to\mathbf{OCs}$ be a polynomial functor, and let $\langle X,\le^e_X,\le^c_X,\ell_{X}\rangle$, $\langle Y,\le^e_Y,\le^c_Y,\ell_{Y}\rangle$ be arbitrary lmsets. The mapping $\mathrm{Rel}(F)$ sends a relation $R\subseteq\langle X,\le^e_X,\le^c_X,\ell_{X}\rangle\times\langle Y,\le^e_Y,\le^c_Y,\ell_{Y}\rangle$ to a lifted relation $\mathrm{Rel}(F)(R)\subseteq F(\langle X,\le^e_X,\le^c_X,\ell_{X}\rangle)\times F(\langle Y,\le^e_Y,\le^c_Y,\ell_{Y}\rangle)$, which is defined by induction on the structure of the functor $F$ as follows:

\begin{enumerate}
  \item If $F$ is the identity functor, then 
  \[\mathrm{Rel}(F)(R)=R\]
  \item If $F$ is a constant functor $\langle X,\le^e_X,\le^c_X,\ell_{X}\rangle\mapsto\mathcal{P}(\langle A,\le^e_A,\le^c_A,\ell_{A}\rangle)$, then
  \[\mathrm{Rel}(F)(R)=\{(U,U)|U\in\mathcal{P}(\langle A,\le^e_A,\le^c_A,\ell_{A}\rangle)\}\]
  \item If $F$ is a product $F_{1}\times F_{2}$, then
  \[\mathrm{Rel}(F)(R)=\{((u_{1},u_{2}),(v_{1},v_{2}))|\mathrm{Rel}(F_{1})(R)(u_{1},v_{1})\wedge\mathrm{Rel}(F_{2})(R)(u_{2},v_{2})\}\]
  \item If $F$ is a set-indexed coproduct $\coprod_{i\in I}F_{i}$, then
  \[\mathrm{Rel}(F)(R)=\bigcup_{i\in I}\{(\kappa_{i}(u),\kappa_{i}(v))|\mathrm{Rel}(F_{i})(R)(u,v)\}\]
  \item If $F$ is an exponent $G^{\mathcal{P}(\langle A,\le^e_A,\le^c_A,\ell_{A}\rangle)}$, then
  \[\mathrm{Rel}(F)(R)=\{(f,g)|\forall U\in\mathcal{P}(\langle A,\le^e_A,\le^c_A,\ell_{A}\rangle).\mathrm{Rel}(G)(R)(f(U),g(U))\}\]
  \item If $F$ is a merge $F_{1}\mid F_{2}$, then
  \[\mathrm{Rel}(F)(R)=\{((u_{1},u_{2}),(v_{1},v_{2}))|\mathrm{Rel}(F_{1})(R)(u_{1},v_{1})\wedge\mathrm{Rel}(F_{2})(R)(u_{2},v_{2})\}\]
  \item If $F$ is a powerset $\mathcal{P}(G)$, then
  \[\mathrm{Rel}(F)(R)=\{(U,V)|\forall u\in U.\exists v\in V.\mathrm{Rel}(G)(R)(u,v)\wedge \forall v\in V.\exists u\in U.\mathrm{Rel}(G)(R)(u,v)\}\]
\end{enumerate}
\end{definition}

\begin{definition}[Step bisimulation]
Let $F:\mathbf{OCs}\to\mathbf{OCs}$ be a polynomial functor. A step bisimulation for coalgebras $c:\langle X,\le^e_X,\le^c_X,\ell_{X}\rangle\to F(\langle X,\le^e_X,\le^c_X,\ell_{X}\rangle)$ and $d:\langle Y,\le^e_Y,\le^c_Y,\ell_{Y}\rangle\to F(\langle Y,\le^e_Y,\le^c_Y,\ell_{Y}\rangle)$ is a relation $R\subseteq\langle X,\le^e_X,\le^c_X,\ell_{X}\rangle\times\langle Y,\le^e_Y,\le^c_Y,\ell_{Y}\rangle$ satisfying:

\[(x,y)\in R\Rightarrow(c(x),d(y))\in\mathrm{Rel}(F)(R)\]

A step bisimulation equivalence is a step bisimulation on a single coalgebra which is an equivalence relation.
\end{definition}

\begin{definition}[Step bisimilarity]
Let $c:\langle X,\le^e_X,\le^c_X,\ell_{X}\rangle\to F(\langle X,\le^e_X,\le^c_X,\ell_{X}\rangle)$ and $d:\langle Y,\le^e_Y,\le^c_Y,\ell_{Y}\rangle\to F(\langle Y,\le^e_Y,\le^c_Y,\ell_{Y}\rangle)$ be two coalgebras of a polynomial functor $F$. The step bisimilarity relation $\underline{\leftrightarrow}$ is the union of all step bisimulations:

\[x\underline{\leftrightarrow}y\Leftrightarrow\exists R\subseteq \langle X,\le^e_X,\le^c_X,\ell_{X}\rangle\times\langle Y,\le^e_Y,\le^c_Y,\ell_{Y}\rangle.R\mbox{ is a step bisimulation for }c\mbox{ and }d,\mbox{ and }R(x,y)\] 
\end{definition}

\begin{theorem}\label{tbisimulation}
Let $F:\mathbf{OCs}\to\mathbf{OCs}$ be a finite polynomial functor which has a final coalgebra $\zeta:\langle Z,\le^e_Z,\le^c_Z,\ell_{Z}\rangle\xrightarrow{\cong}F(\langle Z,\le^e_Z,\le^c_Z,\ell_{Z}\rangle)$. Let $c:\langle X,\le^e_X,\le^c_X,\ell_{X}\rangle\to F(\langle X,\le^e_X,\le^c_X,\ell_{X}\rangle)$ and $d:\langle Y,\le^e_Y,\le^c_Y,\ell_{Y}\rangle\to F(\langle Y,\le^e_Y,\le^c_Y,\ell_{Y}\rangle)$ be arbitrary coalgebras, with associated behaviour homomorphism $\mathrm{beh}_{c}:\langle X,\le^e_X,\le^c_X,\ell_{X}\rangle\to\langle Z,\le^e_Z,\le^c_Z,\ell_{Z}\rangle$ and $\mathrm{beh}_{d}:\langle Y,\le^e_Y,\le^c_Y,\ell_{Y}\rangle\to\langle Z,\le^e_Z,\le^c_Z,\ell_{Z}\rangle$. Two states $x\in \langle X,\le^e_X,\le^c_X,\ell_{X}\rangle$ and $y\in\langle Y,\le^e_Y,\le^c_Y,\ell_{Y}\rangle$ are then step bisimilar if and only if they have the same truly concurrent behaviour:

\[x\prescript{}{c}{\underline{\leftrightarrow}}_{d}y\Leftrightarrow\mathrm{beh}_{c}(x)=\mathrm{beh}_{d}(y)\]
\end{theorem}

\begin{definition}[Congruence]
Let $F:\mathbf{OCs}\to\mathbf{OCs}$ be a polynomial functor. A congruence for algebras $a:F(\langle X,\le^e_X,\le^c_X,\ell_{X}\rangle)\to\langle X,\le^e_X,\le^c_X,\ell_{X}\rangle$ and $b:F(\langle Y,\le^e_Y,\le^c_Y,\ell_{Y}\rangle)\to\langle Y,\le^e_Y,\le^c_Y,\ell_{Y}\rangle$ is a relation $R\subseteq\langle X,\le^e_X,\le^c_X,\ell_{X}\rangle\times\langle Y,\le^e_Y,\le^c_Y,\ell_{Y}\rangle$ satisfying:

\[(u,v)\in\mathrm{Rel}(F)(R)\Rightarrow(a(u),b(v))\in R\]

A congruence equivalence is a congruence which is an equivalence relation.
\end{definition}

\begin{theorem}\label{tcongruence}
Assume two algebras $a:F(\langle X,\le^e_X,\le^c_X,\ell_{X}\rangle)\to\langle X,\le^e_X,\le^c_X,\ell_{X}\rangle$ and $b:F(\langle Y,\le^e_Y,\le^c_Y,\ell_{Y}\rangle)\to\langle Y,\le^e_Y,\le^c_Y,\ell_{Y}\rangle$ of a Kripke polynomial functor $F$. A relation $R\subseteq \langle X,\le^e_X,\le^c_X,\ell_{X}\rangle\times\langle Y,\le^e_Y,\le^c_Y,\ell_{Y}\rangle$ is a congruence if and only if it carries a unique algebra $F(R)\to R$ itself, making $\langle r_1,r_2\rangle:R\rightarrowtail \langle X,\le^e_X,\le^c_X,\ell_{X}\rangle\times\langle Y,\le^e_Y,\le^c_Y,\ell_{Y}\rangle$ of the inclusion homomorphisms of algebras as:

\[\begin{tikzcd}[row sep=huge, column sep=huge]
F(\langle X,\le^e_X,\le^c_X,\ell_{X}\rangle)  \arrow[d,"a"']& F(R) \arrow[l, "\ensuremath{F(r_1)}"']\arrow[r, "\ensuremath{F(r_2)}"] \arrow[d,"\ensuremath{\langle r_1,r_2\rangle}"']& F(\langle Y,\le^e_Y,\le^c_Y,\ell_{Y}\rangle)\arrow[d,"b"]\\
\langle X,\le^e_X,\le^c_X,\ell_{X}\rangle & R \arrow[l, "\ensuremath{r_1}"]\arrow[r, "\ensuremath{r_2}"'] &\langle Y,\le^e_Y,\le^c_Y,\ell_{Y}\rangle
\end{tikzcd}\]
\end{theorem} 

\subsubsection{Algebra for True Concurrency $APTC$}

Since the functor $\langle X,\le^e_X,\le^c_X,\ell_{X}\rangle\mapsto\mathcal{P}_{\mathrm{fin}}(\langle X,\le^e_X,\le^c_X,\ell_{X}\rangle)^{\mathcal{P}_{\mathrm{fin}}(\langle A,\le^e_A,\le^c_A,\ell_{A}\rangle)}$ is a finite Kripke polynomial functor which has a final coalgebra:

\[Z_{\mathcal{P}_{\mathrm{fin}}(\langle A,\le^e_A,\le^c_A,\ell_{A}\rangle)}\xlongrightarrow[\cong]{\zeta_{\mathcal{P}_{\mathrm{fin}}(\langle A,\le^e_A,\le^c_A,\ell_{A}\rangle)}}\mathcal{P}_{\mathrm{fin}}(Z_{\mathcal{P}_{\mathrm{fin}}(\langle A,\le^e_A,\le^c_A,\ell_{A}\rangle)})^{\mathcal{P}_{\mathrm{fin}}(\langle A,\le^e_A,\le^c_A,\ell_{A}\rangle)}\]

The elements of $Z_{\mathcal{P}_{\mathrm{fin}}(\langle A,\le^e_A,\le^c_A,\ell_{A}\rangle)}$ of the final coalgebra is called truly concurrent processes.

The null process is a trivial process which does nothing:

\[\mathbf{0}\triangleq\zeta_{\mathcal{P}_{\mathrm{fin}}(\langle A,\le^e_A,\le^c_A,\ell_{A}\rangle)}^{-1}(\lambda U\in\mathcal{P}_{\mathrm{fin}}(\langle A,\le^e_A,\le^c_A,\ell_{A}\rangle).\emptyset)\]

Given two processes $z_1,z_2\in Z_{\mathcal{P}_{\mathrm{fin}}(\langle A,\le^e_A,\le^c_A,\ell_{A}\rangle)}$, the sum process $z_1+z_2\in Z_{\mathcal{P}_{\mathrm{fin}}(\langle A,\le^e_A,\le^c_A,\ell_{A}\rangle)}$ is defined as follows:

\[z_1+z_2\triangleq\zeta_{\mathcal{P}_{\mathrm{fin}}(\langle A,\le^e_A,\le^c_A,\ell_{A}\rangle)}^{-1}(\lambda U\in \mathcal{P}_{\mathrm{fin}}(\langle A,\le^e_A,\le^c_A,\ell_{A}\rangle).\zeta_{\mathcal{P}_{\mathrm{fin}}(\langle A,\le^e_A,\le^c_A,\ell_{A}\rangle)}(z_1)(U)\cup\zeta_{\mathcal{P}_{\mathrm{fin}}(\langle A,\le^e_A,\le^c_A,\ell_{A}\rangle)}(z_2)(U))\]

Then, we have:

\[\begin{aligned}
(z_1+z_2)\xrightarrow{U}w &\Leftrightarrow w\in\zeta_{\mathcal{P}_{\mathrm{fin}}(\langle A,\le^e_A,\le^c_A,\ell_{A}\rangle)}(z_1+z_2)(U)\\
&\Leftrightarrow w\in\zeta_{\mathcal{P}_{\mathrm{fin}}(\langle A,\le^e_A,\le^c_A,\ell_{A}\rangle)}(z_1)(U)\cup\zeta_{\mathcal{P}_{\mathrm{fin}}(\langle A,\le^e_A,\le^c_A,\ell_{A}\rangle)}(z_2)(U)\\
&\Leftrightarrow w\in\zeta_{\mathcal{P}_{\mathrm{fin}}(\langle A,\le^e_A,\le^c_A,\ell_{A}\rangle)}(z_1)(U)\mbox{ or }w\in\zeta_{\mathcal{P}_{\mathrm{fin}}(\langle A,\le^e_A,\le^c_A,\ell_{A}\rangle)}(z_2)(U)\\
&\Leftrightarrow z_1\xrightarrow{U}w\mbox{ or }z_2\xrightarrow{U}w
\end{aligned}\]

which can be described by the following two rules:

\[\frac{z_1\xrightarrow{U}w}{z_1+z_2\xrightarrow{U}w}\qquad\frac{z_2\xrightarrow{U}w}{z_1+z_2\xrightarrow{U}w}\]

Given a process $z\in Z_{\mathcal{P}_{\mathrm{fin}}(\langle A,\le^e_A,\le^c_A,\ell_{A}\rangle)}$ and a step $V\in\mathcal{P}_{\mathrm{fin}}(\langle A,\le^e_A,\le^c_A,\ell_{A}\rangle)$, the prefixing process $V.z$ is defined as follows:

\[V.z\triangleq\zeta_{\mathcal{P}_{\mathrm{fin}}(\langle A,\le^e_A,\le^c_A,\ell_{A}\rangle)}^{-1}(\lambda U\in \mathcal{P}_{\mathrm{fin}}(\langle A,\le^e_A,\le^c_A,\ell_{A}\rangle)).\{z|U\cong V\}\]

Then, we have:

\[\begin{aligned}
(V.z)\xrightarrow{U}w &\Leftrightarrow w\in\zeta_{\mathcal{P}_{\mathrm{fin}}(\langle A,\le^e_A,\le^c_A,\ell_{A}\rangle)}(V.z)(U)\\
&\Leftrightarrow w\in\{z|U\cong V\}\\
&\Leftrightarrow U\cong V\wedge w=z
\end{aligned}\]

which can be described by the following rule:

\[\frac{}{V.z\xrightarrow{V}z}\]

Given two processes $z_1,z_2\in Z_{\mathcal{P}_{\mathrm{fin}}(\langle A,\le^e_A,\le^c_A,\ell_{A}\rangle)}$, the concurrent process $z_1\between z_2\in Z_{\mathcal{P}_{\mathrm{fin}}(\langle A,\le^e_A,\le^c_A,\ell_{A}\rangle)}$ is defined as follows:

\[z_1\between z_2\triangleq\zeta_{\mathcal{P}_{\mathrm{fin}}(\langle A,\le^e_A,\le^c_A,\ell_{A}\rangle)}^{-1}(\lambda U,V\in \mathcal{P}_{\mathrm{fin}}(\langle A,\le^e_A,\le^c_A,\ell_{A}\rangle).\zeta_{\mathcal{P}_{\mathrm{fin}}(\langle A,\le^e_A,\le^c_A,\ell_{A}\rangle)}(z_1)(U)\cup\zeta_{\mathcal{P}_{\mathrm{fin}}(\langle A,\le^e_A,\le^c_A,\ell_{A}\rangle)}(z_2)(V))\]

Then, we have:

\[\begin{aligned}
(z_1\between z_2)\xrightarrow{U\between V}z_1'\between z_2' &\Leftrightarrow z_1'\between z_2'\in\zeta_{\mathcal{P}_{\mathrm{fin}}(\langle A,\le^e_A,\le^c_A,\ell_{A}\rangle)}(z_1\between z_2)(U\between V)\\
&\Leftrightarrow z_1'\between z_2'\in\zeta_{\mathcal{P}_{\mathrm{fin}}(\langle A,\le^e_A,\le^c_A,\ell_{A}\rangle)}(z_1)(U)\cup\zeta_{\mathcal{P}_{\mathrm{fin}}(\langle A,\le^e_A,\le^c_A,\ell_{A}\rangle)}(z_2)(V)\\
&\Leftrightarrow z_1'\in\zeta_{\mathcal{P}_{\mathrm{fin}}(\langle A,\le^e_A,\le^c_A,\ell_{A}\rangle)}(z_1)(U)\mbox{ and }z_2'\in\zeta_{\mathcal{P}_{\mathrm{fin}}(\langle A,\le^e_A,\le^c_A,\ell_{A}\rangle)}(z_2)(V)\\
&\Leftrightarrow z_1\xrightarrow{U}z_1'\mbox{ and }z_2\xrightarrow{V}z_2'
\end{aligned}\]

which can be described by the following two rules:

\[\frac{z_1\xrightarrow{U}z_1'\quad z_2\xrightarrow{V}z_2'}{z_1\between z_2\xrightarrow{U\between V}z_1'\between z_2'}\]

Given two processes $z_1,z_2\in Z_{\mathcal{P}_{\mathrm{fin}}(\langle A,\le^e_A,\le^c_A,\ell_{A}\rangle)}$, the parallel process $z_1\parallel z_2\in Z_{\mathcal{P}_{\mathrm{fin}}(\langle A,\le^e_A,\le^c_A,\ell_{A}\rangle)}$ is defined as follows:

\[z_1\parallel z_2\triangleq\zeta_{\mathcal{P}_{\mathrm{fin}}(\langle A,\le^e_A,\le^c_A,\ell_{A}\rangle)}^{-1}(\lambda U,V\in \mathcal{P}_{\mathrm{fin}}(\langle A,\le^e_A,\le^c_A,\ell_{A}\rangle).\zeta_{\mathcal{P}_{\mathrm{fin}}(\langle A,\le^e_A,\le^c_A,\ell_{A}\rangle)}(z_1)(U)\cup\zeta_{\mathcal{P}_{\mathrm{fin}}(\langle A,\le^e_A,\le^c_A,\ell_{A}\rangle)}(z_2)(V))\]

Then, we have:

\[\begin{aligned}
(z_1\parallel z_2)\xrightarrow{U\parallel V}z_1'\between z_2' &\Leftrightarrow z_1'\between z_2'\in\zeta_{\mathcal{P}_{\mathrm{fin}}(\langle A,\le^e_A,\le^c_A,\ell_{A}\rangle)}(z_1\parallel z_2)(U\parallel V)\\
&\Leftrightarrow z_1'\between z_2'\in\zeta_{\mathcal{P}_{\mathrm{fin}}(\langle A,\le^e_A,\le^c_A,\ell_{A}\rangle)}(z_1)(U)\cup\zeta_{\mathcal{P}_{\mathrm{fin}}(\langle A,\le^e_A,\le^c_A,\ell_{A}\rangle)}(z_2)(V)\\
&\Leftrightarrow z_1'\in\zeta_{\mathcal{P}_{\mathrm{fin}}(\langle A,\le^e_A,\le^c_A,\ell_{A}\rangle)}(z_1)(U)\mbox{ and }z_2'\in\zeta_{\mathcal{P}_{\mathrm{fin}}(\langle A,\le^e_A,\le^c_A,\ell_{A}\rangle)}(z_2)(V)\\
&\Leftrightarrow z_1\xrightarrow{U}z_1'\mbox{ and }z_2\xrightarrow{V}z_2'
\end{aligned}\]

which can be described by the following two rules:

\[\frac{z_1\xrightarrow{U}z_1'\quad z_2\xrightarrow{V}z_2'}{z_1\parallel z_2\xrightarrow{U\parallel V}z_1'\between z_2'}\]

Given two processes $z_1,z_2\in Z_{\mathcal{P}_{\mathrm{fin}}(\langle A,\le^e_A,\le^c_A,\ell_{A}\rangle)}$, the communication process $z_1\mid z_2\in Z_{\mathcal{P}_{\mathrm{fin}}(\langle A,\le^e_A,\le^c_A,\ell_{A}\rangle)}$ is defined as follows:

\[z_1\mid z_2\triangleq\zeta_{\mathcal{P}_{\mathrm{fin}}(\langle A,\le^e_A,\le^c_A,\ell_{A}\rangle)}^{-1}(\lambda U,V\in \mathcal{P}_{\mathrm{fin}}(\langle A,\le^e_A,\le^c_A,\ell_{A}\rangle).\zeta_{\mathcal{P}_{\mathrm{fin}}(\langle A,\le^e_A,\le^c_A,\ell_{A}\rangle)}(z_1)(U)\cup\zeta_{\mathcal{P}_{\mathrm{fin}}(\langle A,\le^e_A,\le^c_A,\ell_{A}\rangle)}(z_2)(V))\]

Then, we have:

\[\begin{aligned}
(z_1\mid z_2)\xrightarrow{U\mid V}z_1'\between z_2' &\Leftrightarrow z_1'\between z_2'\in\zeta_{\mathcal{P}_{\mathrm{fin}}(\langle A,\le^e_A,\le^c_A,\ell_{A}\rangle)}(z_1\mid z_2)(U\mid V)\\
&\Leftrightarrow z_1'\between z_2'\in\zeta_{\mathcal{P}_{\mathrm{fin}}(\langle A,\le^e_A,\le^c_A,\ell_{A}\rangle)}(z_1)(U)\cup\zeta_{\mathcal{P}_{\mathrm{fin}}(\langle A,\le^e_A,\le^c_A,\ell_{A}\rangle)}(z_2)(V)\\
&\Leftrightarrow z_1'\in\zeta_{\mathcal{P}_{\mathrm{fin}}(\langle A,\le^e_A,\le^c_A,\ell_{A}\rangle)}(z_1)(U)\mbox{ and }z_2'\in\zeta_{\mathcal{P}_{\mathrm{fin}}(\langle A,\le^e_A,\le^c_A,\ell_{A}\rangle)}(z_2)(V)\\
&\Leftrightarrow z_1\xrightarrow{U}z_1'\mbox{ and }z_2\xrightarrow{V}z_2'
\end{aligned}\]

which can be described by the following two rules:

\[\frac{z_1\xrightarrow{U}z_1'\quad z_2\xrightarrow{V}z_2'}{z_1\mid z_2\xrightarrow{U\mid V}z_1'\between z_2'}\]

For an arbitrary powerset $\mathcal{P}_{\mathrm{fin}}(\langle A,\le^e_A,\le^c_A,\ell_{A}\rangle)$ of steps, let polynomial functor $\Sigma_{\mathcal{P}_{\mathrm{fin}}(\langle A,\le^e_A,\le^c_A,\ell_{A}\rangle)}:\mathbf{OCs}\to\mathbf{OCs}$ given by

$\Sigma_{\mathcal{P}_{\mathrm{fin}}(\langle A,\le^e_A,\le^c_A,\ell_{A}\rangle)}(\langle X,\le^e_X,\le^c_X,\ell_{X}\rangle)
=1+(\mathcal{P}_{\mathrm{fin}}(\langle A,\le^e_A,\le^c_A,\ell_{A}\rangle)\times\langle X,\le^e_X,\le^c_X,\ell_{X}\rangle)+(\langle X,\le^e_X,\le^c_X,\ell_{X}\rangle\times\langle X,\le^e_X,\le^c_X,\ell_{X}\rangle)+(\langle X,\le^e_X,\le^c_X,\ell_{X}\rangle\times\langle X,\le^e_X,\le^c_X,\ell_{X}\rangle)+(\langle X,\le^e_X,\le^c_X,\ell_{X}\rangle\mid\langle X,\le^e_X,\le^c_X,\ell_{X}\rangle)$

An algebra $\Sigma_{\mathcal{P}_{\mathrm{fin}}(\langle A,\le^e_A,\le^c_A,\ell_{A}\rangle)}(\langle X,\le^e_X,\le^c_X,\ell_{X}\rangle)\to\langle X,\le^e_X,\le^c_X,\ell_{X}\rangle$ for this functor consists of five operations: $\mathbf{0}:1\to\langle X,\le^e_X,\le^c_X,\ell_{X}\rangle$ for null process, $.:\mathcal{P}_{\mathrm{fin}}(\langle A,\le^e_A,\le^c_A,\ell_{A}\rangle)\times \langle X,\le^e_X,\le^c_X,\ell_{X}\rangle\to\langle X,\le^e_X,\le^c_X,\ell_{X}\rangle$ for prefixing process, $+:\langle X,\le^e_X,\le^c_X,\ell_{X}\rangle\times\langle X,\le^e_X,\le^c_X,\ell_{X}\rangle\to\langle X,\le^e_X,\le^c_X,\ell_{X}\rangle$, $\parallel:\langle X,\le^e_X,\le^c_X,\ell_{X}\rangle\times\langle X,\le^e_X,\le^c_X,\ell_{X}\rangle\to\langle X,\le^e_X,\le^c_X,\ell_{X}\rangle$ for parallel process and $\mid:\langle X,\le^e_X,\le^c_X,\ell_{X}\rangle\mid\langle X,\le^e_X,\le^c_X,\ell_{X}\rangle\to\langle X,\le^e_X,\le^c_X,\ell_{X}\rangle$ for communication process. The final coalgebra $Z_{\mathcal{P}_{\mathrm{fin}}(\langle A,\le^e_A,\le^c_A,\ell_{A}\rangle)}\xrightarrow{\cong}\mathcal{P}_{\mathrm{fin}}(Z_{\mathcal{P}_{\mathrm{fin}}(\langle A,\le^e_A,\le^c_A,\ell_{A}\rangle)})^{\mathcal{P}_{\mathrm{fin}}(\langle A,\le^e_A,\le^c_A,\ell_{A}\rangle)}$ carries a $\Sigma_{\mathcal{P}_{\mathrm{fin}}(\langle A,\le^e_A,\le^c_A,\ell_{A}\rangle)}$ algebra written as $\xi_{\mathcal{P}_{\mathrm{fin}}(\langle A,\le^e_A,\le^c_A,\ell_{A}\rangle)}:\Sigma_{\mathcal{P}_{\mathrm{fin}}(\langle A,\le^e_A,\le^c_A,\ell_{A}\rangle)}(Z_{\mathcal{P}_{\mathrm{fin}}(\langle A,\le^e_A,\le^c_A,\ell_{A}\rangle)})\to Z_{\mathcal{P}_{\mathrm{fin}}(\langle A,\le^e_A,\le^c_A,\ell_{A}\rangle)}$. Thus, a bialgebra is obtained:

\[\begin{tikzcd}[row sep=normal, column sep=normal]
\Sigma_{\mathcal{P}_{\mathrm{fin}}(\langle A,\le^e_A,\le^c_A,\ell_{A}\rangle)}(Z_{\mathcal{P}_{\mathrm{fin}}(\langle A,\le^e_A,\le^c_A,\ell_{A}\rangle)}) \arrow[r, "\ensuremath{\xi_{\mathcal{P}_{\mathrm{fin}}(\langle A,\le^e_A,\le^c_A,\ell_{A}\rangle)}}"] & Z_{\mathcal{P}_{\mathrm{fin}}(\langle A,\le^e_A,\le^c_A,\ell_{A}\rangle)}\arrow[r,"\ensuremath{\zeta_{\mathcal{P}_{\mathrm{fin}}(\langle A,\le^e_A,\le^c_A,\ell_{A}\rangle)}}","\ensuremath{\cong}"'] & \mathcal{P}_{\mathrm{fin}}(Z_{\mathcal{P}_{\mathrm{fin}}(\langle A,\le^e_A,\le^c_A,\ell_{A}\rangle)})^{\mathcal{P}_{\mathrm{fin}}(\langle A,\le^e_A,\le^c_A,\ell_{A}\rangle)} 
\end{tikzcd}\] 

The initial $\Sigma_{\mathcal{P}_{\mathrm{fin}}(\langle A,\le^e_A,\le^c_A,\ell_{A}\rangle)}$-algebra denoted $P_{\mathcal{P}_{\mathrm{fin}}(\langle A,\le^e_A,\le^c_A,\ell_{A}\rangle)}$ as $\alpha:\Sigma_{\mathcal{P}_{\mathrm{fin}}(\langle A,\le^e_A,\le^c_A,\ell_{A}\rangle)}(P_{\mathcal{P}_{\mathrm{fin}}(\langle A,\le^e_A,\le^c_A,\ell_{A}\rangle)})\xrightarrow{\cong}P_{\mathcal{P}_{\mathrm{fin}}(\langle A,\le^e_A,\le^c_A,\ell_{A}\rangle)}$, contains the closed process terms, which are built up from $\mathbf{0}$ and $U\in \mathcal{P}_{\mathrm{fin}}(\langle A,\le^e_A,\le^c_A,\ell_{A}\rangle)$. $P_{\mathcal{P}_{\mathrm{fin}}(\langle A,\le^e_A,\le^c_A,\ell_{A}\rangle)}$ also carries a bialgebra structure:

\[\begin{tikzcd}[row sep=normal, column sep=normal]
\Sigma_{\mathcal{P}_{\mathrm{fin}}(\langle A,\le^e_A,\le^c_A,\ell_{A}\rangle)}(P_{\mathcal{P}_{\mathrm{fin}}(\langle A,\le^e_A,\le^c_A,\ell_{A}\rangle)}) \arrow[r, "\ensuremath{\alpha}","\ensuremath{\cong}"'] & P_{\mathcal{P}_{\mathrm{fin}}(\langle A,\le^e_A,\le^c_A,\ell_{A}\rangle)}\arrow[r,"\ensuremath{\beta}"] & \mathcal{P}_{\mathrm{fin}}(P_{\mathcal{P}_{\mathrm{fin}}(\langle A,\le^e_A,\le^c_A,\ell_{A}\rangle)})^{\mathcal{P}_{\mathrm{fin}}(\langle A,\le^e_A,\le^c_A,\ell_{A}\rangle)} 
\end{tikzcd}\]

The coalgebra $\beta$ is defined by induction:

\[\beta(\mathbf{0})(U)=\emptyset\]
\[\beta(V.s)(U)=\{s|V\cong U\}\]
\[\beta(s+t)(U)=\beta(s)(U)\cup\beta(t)(U)\]
\[\beta(s\between t)(U\between V)=\beta(s)(U)\cup\beta(t)(V)\]
\[\beta(s\parallel t)(U\parallel V)=\beta(s)(U)\cup\beta(t)(V)\]
\[\beta(s\mid t)(U\mid V)=\beta(s)(U)\cup\beta(t)(V)\]

The denotational semantics given by initiality and operational semantics given by finality for process terms coincide.

\begin{proposition}
In the above situation, we obtain two maps $P_{\mathcal{P}_{\mathrm{fin}}(\langle A,\le^e_A,\le^c_A,\ell_{A}\rangle)}\to Z_{\mathcal{P}_{\mathrm{fin}}(\langle A,\le^e_A,\le^c_A,\ell_{A}\rangle)}$, one by initiality making the following diagram commute:

\begin{center}
\begin{tikzcd}[row sep=huge, column sep=huge]
\Sigma_{\mathcal{P}_{\mathrm{fin}}(\langle A,\le^e_A,\le^c_A,\ell_{A}\rangle)}(P_{\mathcal{P}_{\mathrm{fin}}(\langle A,\le^e_A,\le^c_A,\ell_{A}\rangle)}) \arrow[dashed,r,"\ensuremath{\Sigma_{\mathcal{P}_{\mathrm{fin}}(\langle A,\le^e_A,\le^c_A,\ell_{A}\rangle)}(\mathrm{int}_{\alpha})}"] \arrow[d,"\ensuremath{\alpha}"',"\cong"] & \Sigma_{\mathcal{P}_{\mathrm{fin}}(\langle A,\le^e_A,\le^c_A,\ell_{A}\rangle)}(Z_{\mathcal{P}_{\mathrm{fin}}(\langle A,\le^e_A,\le^c_A,\ell_{A}\rangle)})  \arrow[d,"\ensuremath{\xi}"]\\
P_{\mathcal{P}_{\mathrm{fin}}(\langle A,\le^e_A,\le^c_A,\ell_{A}\rangle)} \arrow[dashed,r,"\ensuremath{\mathrm{int}_{\alpha}}"'] & Z_{\mathcal{P}_{\mathrm{fin}}(\langle A,\le^e_A,\le^c_A,\ell_{A}\rangle)}
\end{tikzcd}
\end{center}

the other by finality making the following diagram commute:

\begin{center}
\begin{tikzcd}[row sep=huge, column sep=huge]
\mathcal{P}_{\mathrm{fin}}(P_{\mathcal{P}_{\mathrm{fin}}(\langle A,\le^e_A,\le^c_A,\ell_{A}\rangle)})^{\mathcal{P}_{\mathrm{fin}}(\langle A,\le^e_A,\le^c_A,\ell_{A}\rangle)} \arrow[dashed,r,"\ensuremath{\mathcal{P}_{\mathrm{fin}}(\mathrm{beh}_{\beta})^{\mathcal{P}_{\mathrm{fin}}(\langle A,\le^e_A,\le^c_A,\ell_{A}\rangle)}}"] & \mathcal{P}_{\mathrm{fin}}(Z_{\mathcal{P}_{\mathrm{fin}}(\langle A,\le^e_A,\le^c_A,\ell_{A}\rangle)})^{\mathcal{P}_{\mathrm{fin}}(\langle A,\le^e_A,\le^c_A,\ell_{A}\rangle)}\\
P_{\mathcal{P}_{\mathrm{fin}}(\langle A,\le^e_A,\le^c_A,\ell_{A}\rangle)} \arrow[dashed,r,"\ensuremath{\mathrm{beh}_{\beta}}"'] \arrow[u,"\ensuremath{\beta}"]& Z_{\mathcal{P}_{\mathrm{fin}}(\langle A,\le^e_A,\le^c_A,\ell_{A}\rangle)}\arrow[u,"\ensuremath{\cong}","\ensuremath{\zeta}"']
\end{tikzcd}
\end{center}
\end{proposition}

\begin{proof}
By induction on the structure of a term $s\in P_{\mathcal{P}_{\mathrm{fin}}(\langle A,\le^e_A,\le^c_A,\ell_{A}\rangle)}$, we prove that $\mathrm{beh}_{\beta}(s)=\mathrm{int}_{\alpha}(s)$, or equivalently, $\zeta(\mathrm{beh}_{\beta}(s))(U)=\zeta(\mathrm{int}_{\alpha}(s))(U)$:

\[\begin{aligned}
\zeta(\mathrm{beh}_{\beta}(\mathbf{0}))(U)&=\mathcal{P}_{\mathrm{fin}}(\mathrm{beh}_{\beta})^{\mathcal{P}_{\mathrm{fin}}(\langle A,\le^e_A,\le^c_A,\ell_{A}\rangle)}(\beta(\mathbf{0}))(U)\\
&=\mathcal{P}_{\mathrm{fin}}(\mathrm{beh}_{\beta})(\beta(\mathbf{0})(U))\\
&=\mathcal{P}_{\mathrm{fin}}(\mathrm{beh}_{\beta}(\emptyset))\\
&=\emptyset\\
&=\zeta(\mathbf{0})(U)\\
&=\zeta(\mathrm{int}_{\alpha}(\mathbf{0}))(U)
\end{aligned}\]

\[\begin{aligned}
\zeta(\mathrm{beh}_{\beta}(V.s))(U)&=\mathcal{P}_{\mathrm{fin}}(\mathrm{beh}_{\beta})(\beta(V.s))(U)\\
&=\mathcal{P}_{\mathrm{fin}}(\mathrm{beh}_{\beta})(\{s|V\cong U\})\\
&=\{\mathrm{beh}_{\beta}(s)|V\cong U\}\\
&=\{\mathrm{int}_{\alpha}(s)|V\cong U\}\\
&=\zeta(V.\mathrm{int}_{\alpha}(s))(U)\\
&=\zeta(\mathrm{int}_{\alpha}(V.s))(U)
\end{aligned}\]

\[\begin{aligned}
\zeta(\mathrm{beh}_{\beta}(s+t))(U)&=\mathcal{P}_{\mathrm{fin}}(\mathrm{beh}_{\beta})(\beta(s+t))(U)\\
&=\mathcal{P}_{\mathrm{fin}}(\mathrm{beh}_{\beta})(\beta(s)(U)\cup\beta(t)(U))\\
&=\mathcal{P}_{\mathrm{fin}}(\mathrm{beh}_{\beta})(\beta(s)(U))\cup \mathcal{P}_{\mathrm{fin}}(\mathrm{beh}_{\beta})(\beta(t)(U))\\
&=\zeta(\mathrm{beh}_{\beta}(s))(U)\cup \zeta(\mathrm{beh}_{\beta}(t))(U)\\
&=\zeta(\mathrm{int}_{\alpha}(s))(U)\cup \zeta(\mathrm{int}_{\alpha}(t))(U)\\
&=\zeta(\mathrm{int}_{\alpha}(s)+ \mathrm{int}_{\alpha}(t))(U)\\
&=\zeta(\mathrm{int}_{\alpha}(s+t))(U)
\end{aligned}\]

\[\begin{aligned}
\zeta(\mathrm{beh}_{\beta}(s\between t))(U\between V)&=\mathcal{P}_{\mathrm{fin}}(\mathrm{beh}_{\beta})(\beta(s\between t))(U\between V)\\
&=\mathcal{P}_{\mathrm{fin}}(\mathrm{beh}_{\beta})(\beta(s)(U)\cup\beta(t)(V))\\
&=\mathcal{P}_{\mathrm{fin}}(\mathrm{beh}_{\beta})(\beta(s)(U))\cup \mathcal{P}_{\mathrm{fin}}(\mathrm{beh}_{\beta})(\beta(t)(V))\\
&=\zeta(\mathrm{beh}_{\beta}(s))(U)\cup \zeta(\mathrm{beh}_{\beta}(t))(V)\\
&=\zeta(\mathrm{int}_{\alpha}(s))(U)\cup \zeta(\mathrm{int}_{\alpha}(t))(V)\\
&=\zeta(\mathrm{int}_{\alpha}(s)\between \mathrm{int}_{\alpha}(t))(U\between V)\\
&=\zeta(\mathrm{int}_{\alpha}(s\between t))(U\between V)
\end{aligned}\]

\[\begin{aligned}
\zeta(\mathrm{beh}_{\beta}(s\parallel t))(U\parallel V)&=\mathcal{P}_{\mathrm{fin}}(\mathrm{beh}_{\beta})(\beta(s\parallel t))(U\parallel V)\\
&=\mathcal{P}_{\mathrm{fin}}(\mathrm{beh}_{\beta})(\beta(s)(U)\cup\beta(t)(V))\\
&=\mathcal{P}_{\mathrm{fin}}(\mathrm{beh}_{\beta})(\beta(s)(U))\cup \mathcal{P}_{\mathrm{fin}}(\mathrm{beh}_{\beta})(\beta(t)(V))\\
&=\zeta(\mathrm{beh}_{\beta}(s))(U)\cup \zeta(\mathrm{beh}_{\beta}(t))(V)\\
&=\zeta(\mathrm{int}_{\alpha}(s))(U)\cup \zeta(\mathrm{int}_{\alpha}(t))(V)\\
&=\zeta(\mathrm{int}_{\alpha}(s)\parallel \mathrm{int}_{\alpha}(t))(U\parallel V)\\
&=\zeta(\mathrm{int}_{\alpha}(s\parallel t))(U\parallel V)
\end{aligned}\]

\[\begin{aligned}
\zeta(\mathrm{beh}_{\beta}(s\mid t))(U\mid V)&=\mathcal{P}_{\mathrm{fin}}(\mathrm{beh}_{\beta})(\beta(s\mid t))(U\mid V)\\
&=\mathcal{P}_{\mathrm{fin}}(\mathrm{beh}_{\beta})(\beta(s)(U)\cup\beta(t)(V))\\
&=\mathcal{P}_{\mathrm{fin}}(\mathrm{beh}_{\beta})(\beta(s)(U))\cup \mathcal{P}_{\mathrm{fin}}(\mathrm{beh}_{\beta})(\beta(t)(V))\\
&=\zeta(\mathrm{beh}_{\beta}(s))(U)\cup \zeta(\mathrm{beh}_{\beta}(t))(V)\\
&=\zeta(\mathrm{int}_{\alpha}(s))(U)\cup \zeta(\mathrm{int}_{\alpha}(t))(V)\\
&=\zeta(\mathrm{int}_{\alpha}(s)\mid \mathrm{int}_{\alpha}(t))(U\mid V)\\
&=\zeta(\mathrm{int}_{\alpha}(s\mid t))(U\mid V)
\end{aligned}\]
\end{proof}

\begin{proposition}
Still in the above situation, the step bisimilarity relation $\underline{\leftrightarrow}$ on the set $P_{\mathcal{P}_{\mathrm{fin}}(\langle A,\le^e_A,\le^c_A,\ell_{A}\rangle)}$ of process terms is a congruence.
\end{proposition}

\begin{proof}
In the following diagram, $f=\mathrm{beh}_{\beta}=\mathrm{int}_{\alpha}$.

\begin{center}
\adjustbox{rotate=90}{
\begin{tikzcd}[row sep=huge, column sep=huge]
\Sigma_{\mathcal{P}_{\mathrm{fin}}(\langle A,\le^e_A,\le^c_A,\ell_{A}\rangle)}(\underline{\leftrightarrow})\arrow[r,"d"] \arrow[dashed,d]&\Sigma_{\mathcal{P}_{\mathrm{fin}}(\langle A,\le^e_A,\le^c_A,\ell_{A}\rangle)}(P_{\mathcal{P}_{\mathrm{fin}}(\langle A,\le^e_A,\le^c_A,\ell_{A}\rangle)})\times\Sigma_{\mathcal{P}_{\mathrm{fin}}(\langle A,\le^e_A,\le^c_A,\ell_{A}\rangle)}(P_{\mathcal{P}_{\mathrm{fin}}(\langle A,\le^e_A,\le^c_A,\ell_{A}\rangle)}) \arrow[r,shift left,"\ensuremath{\Sigma_{\mathcal{P}_{\mathrm{fin}}(\langle A,\le^e_A,\le^c_A,\ell_{A}\rangle)}(f\circ\pi_1)}"]\arrow[r,shift right,"\ensuremath{\Sigma_{\mathcal{P}_{\mathrm{fin}}(\langle A,\le^e_A,\le^c_A,\ell_{A}\rangle)}(f\circ\pi_2)}"'] \arrow[d,"\ensuremath{\alpha\times\alpha}"] & \Sigma_{\mathcal{P}_{\mathrm{fin}}(\langle A,\le^e_A,\le^c_A,\ell_{A}\rangle)}(Z_{\mathcal{P}_{\mathrm{fin}}(\langle A,\le^e_A,\le^c_A,\ell_{A}\rangle)})  \arrow[d,"\ensuremath{\xi_{\mathcal{P}_{\mathrm{fin}}(\langle A,\le^e_A,\le^c_A,\ell_{A}\rangle)}}"]\\
\underline{\leftrightarrow} \arrow[r,tail,"e"'] &P_{\mathcal{P}_{\mathrm{fin}}(\langle A,\le^e_A,\le^c_A,\ell_{A}\rangle)}\times P_{\mathcal{P}_{\mathrm{fin}}(\langle A,\le^e_A,\le^c_A,\ell_{A}\rangle)} \arrow[r,shift left,"\ensuremath{f\circ \pi_1}"]\arrow[r,shift right,"\ensuremath{f\circ \pi_2}"'] & Z_{\mathcal{P}_{\mathrm{fin}}(\langle A,\le^e_A,\le^c_A,\ell_{A}\rangle)}
\end{tikzcd}}
\end{center}

The map $e$ is the equalizer of $f\circ\pi_1$ and $f\circ \pi_2$ by \cref{tbisimulation}, and $d$ is the pair $\langle\Sigma_{\mathcal{P}_{\mathrm{fin}}(\langle A,\le^e_A,\le^c_A,\ell_{A}\rangle)}(\pi_1\circ e),\Sigma_{\mathcal{P}_{\mathrm{fin}}(\langle A,\le^e_A,\le^c_A,\ell_{A}\rangle)}(\pi_2\circ e)\rangle$. We show that $(\alpha\times\alpha)\circ d$ equalizes $f\circ\pi_1$ and $f\circ\pi_2$. The dashed arrow then exists by the universal property of the equalizer $e$, make $\underline{\leftrightarrow}$ a congruence by \cref{tcongruence}.

\[\begin{aligned}
f\circ\pi_1\circ(\alpha\times\alpha)\circ d &=\mathrm{int}_{\alpha}\circ\alpha\circ\pi_1\circ d\\
&=\xi\circ\Sigma_{\mathcal{P}_{\mathrm{fin}}(\langle A,\le^e_A,\le^c_A,\ell_{A}\rangle)}(\mathrm{int}_{\alpha})\circ\Sigma_{\mathcal{P}_{\mathrm{fin}}(\langle A,\le^e_A,\le^c_A,\ell_{A}\rangle)}(\pi_1\circ e)\\
&=\xi\circ\Sigma_{\mathcal{P}_{\mathrm{fin}}(\langle A,\le^e_A,\le^c_A,\ell_{A}\rangle)}(\mathrm{int}_{\alpha})\circ\Sigma_{\mathcal{P}_{\mathrm{fin}}(\langle A,\le^e_A,\le^c_A,\ell_{A}\rangle)}(\pi_2\circ e)\\
&=\mathrm{int}_{\alpha}\circ\alpha\circ\pi_2\circ d\\
&=f\circ\pi_2\circ(\alpha\times\alpha)\circ d
\end{aligned}\]
\end{proof}

\begin{proposition}
The so-called exchange laws hold:

\[\begin{aligned}
(a.s)\between(b.t)&\underline{\leftrightarrow}(a.s)\parallel(b.t)+(a.s)\mid(b.t)\\
&\underline{\leftrightarrow}(a\parallel b).(s\between t)+\gamma(a,b).(s\between t)
\end{aligned}\]
\end{proposition} 

\bibliographystyle{elsarticle-num}
\newpage\bibliography{Refs-AIA}

\begin{thebibliography}{10}
\expandafter\ifx\csname url\endcsname\relax
  \def\url#1{\texttt{#1}}\fi
\expandafter\ifx\csname urlprefix\endcsname\relax\def\urlprefix{URL }\fi
\expandafter\ifx\csname href\endcsname\relax
  \def\href#1#2{#2} \def\path#1{#1}\fi

\bibitem{MCP}
Anthropic,
  \href{https://modelcontextprotocol.io/specification/2025-11-25}{Model context
  protocol specification}, version 2025-11-25 (2025).
\newline\urlprefix\url{https://modelcontextprotocol.io/specification/2025-11-25}

\bibitem{ReAct}
S.~Yao, J.~Zhao, D.~Yu, N.~Du, I.~Shafran, K.~Narasimhan, Y.~Cao,
  \href{https://arxiv.org/abs/2210.03629}{React: Synergizing reasoning and
  acting in language models}, in: International Conference on Learning
  Representations, 2023.
\newline\urlprefix\url{https://arxiv.org/abs/2210.03629}

\bibitem{Toolformer}
T.~Schick, J.~Dwivedi-Yu, R.~Dessì, R.~Raileanu, M.~Lomeli, E.~Hambro,
  L.~Zettlemoyer, N.~Cancedda, T.~Scialom,
  \href{https://arxiv.org/abs/2302.04761}{Toolformer: Language models can teach
  themselves to use tools}, in: Advances in Neural Information Processing
  Systems, Vol.~36, 2023.
\newline\urlprefix\url{https://arxiv.org/abs/2302.04761}

\bibitem{Toolllm}
Y.~Qin, S.~Liang, Y.~Ye, K.~Zhu, L.~Yan, Y.~Lu, Y.~Lin, X.~Cong, X.~Tang,
  B.~Qian, S.~Zhao, L.~Hong, R.~Tian, R.~Xie, J.~Zhou, M.~Gerstein, D.~Li,
  Z.~Liu, M.~Sun, \href{https://arxiv.org/abs/2307.16789}{Toolllm: Facilitating
  large language models to master 16000+ real-world apis}, in: International
  Conference on Learning Representations, 2024.
\newline\urlprefix\url{https://arxiv.org/abs/2307.16789}

\bibitem{Gorilla}
S.~G. Patil, T.~Zhang, X.~Wang, J.~E. Gonzalez,
  \href{https://arxiv.org/abs/2305.15334}{Gorilla: Large language model
  connected with massive apis}, in: Advances in Neural Information Processing
  Systems, Vol.~37, 2024.
\newline\urlprefix\url{https://arxiv.org/abs/2305.15334}

\bibitem{HuggingGPT}
Y.~Shen, K.~Song, X.~Tan, D.~Li, W.~Lu, Y.~Zhuang,
  \href{https://arxiv.org/abs/2303.17580}{Hugginggpt: Solving ai tasks with
  chatgpt and its friends in hugging face}, in: Advances in Neural Information
  Processing Systems, Vol.~36, 2023.
\newline\urlprefix\url{https://arxiv.org/abs/2303.17580}

\bibitem{Art}
B.~Paranjape, S.~Lundberg, S.~Singh, H.~Hajishirzi, L.~Zettlemoyer, M.~T.
  Ribeiro, \href{https://arxiv.org/abs/2303.09014}{Art: Automatic multi-step
  reasoning and tool-use for large language models} (2023).
\newblock \href {http://arxiv.org/abs/2303.09014} {\path{arXiv:2303.09014}}.
\newline\urlprefix\url{https://arxiv.org/abs/2303.09014}

\bibitem{LangChain}
H.~Chase, L.~Team, \href{https://github.com/langchain-ai/langchain}{Langchain:
  Building applications with llms through composability} (2022).
\newline\urlprefix\url{https://github.com/langchain-ai/langchain}

\bibitem{AutoGen}
Q.~Wu, G.~Bansal, J.~Zhang, Y.~Wu, S.~Zhang, E.~Zhu, B.~Li, L.~Jiang, X.~Zhang,
  J.~Liu, A.~H. Awadallah, R.~W. White, D.~Burger, C.~Wang,
  \href{https://arxiv.org/abs/2308.08155}{Autogen: Enabling next-gen llm
  applications via multi-agent conversation} (2023).
\newblock \href {http://arxiv.org/abs/2308.08155} {\path{arXiv:2308.08155}}.
\newline\urlprefix\url{https://arxiv.org/abs/2308.08155}

\bibitem{MetaGPT}
S.~Hong, M.~Zhuge, J.~Chen, X.~Zheng, Y.~Cheng, C.~Zhang, J.~Wang, Z.~Wang,
  S.~K.~S. Yau, Z.~Lin, L.~Zhou, C.~Ran, L.~Xiao, C.~Wu, J.~Schmidhuber,
  \href{https://arxiv.org/abs/2308.00352}{Metagpt: Meta programming for a
  multi-agent collaborative framework}, in: International Conference on
  Learning Representations, 2024.
\newline\urlprefix\url{https://arxiv.org/abs/2308.00352}

\bibitem{ChatDev}
C.~Qian, W.~Liu, H.~Liu, N.~Chen, Y.~Dang, J.~Li, C.~Yang, W.~Chen, Y.~Su,
  X.~Cong, J.~Xu, D.~Li, Z.~Liu, M.~Sun,
  \href{https://arxiv.org/abs/2307.07924}{Chatdev: Communicative agents for
  software development}, in: Proceedings of the 62nd Annual Meeting of the
  Association for Computational Linguistics (ACL), Vol.~1, 2024, pp.
  15174--15186.
\newblock \href {https://doi.org/10.18653/v1/2024.acl-long.810}
  {\path{doi:10.18653/v1/2024.acl-long.810}}.
\newline\urlprefix\url{https://arxiv.org/abs/2307.07924}

\bibitem{TaskWeaver}
B.~Qiao, L.~Li, X.~Zhang, S.~He, Y.~Kang, C.~Zhang, F.~Yang, H.~Dong, J.~Zhang,
  L.~Wang, M.~Ma, P.~Zhao, S.~Qin, X.~Qin, C.~Du, Y.~Xu, Q.~Lin, S.~Rajmohan,
  D.~Zhang, \href{https://arxiv.org/abs/2311.17541}{Taskweaver: A code-first
  agent framework} (2023).
\newblock \href {http://arxiv.org/abs/2311.17541} {\path{arXiv:2311.17541}}.
\newline\urlprefix\url{https://arxiv.org/abs/2311.17541}

\bibitem{Separating}
A.~S. Parmar, \href{https://arxiv.org/abs/2605.00827}{Separating intelligence
  from execution: A workflow engine for the model context protocol} (2026).
\newblock \href {http://arxiv.org/abs/2605.00827} {\path{arXiv:2605.00827}}.
\newline\urlprefix\url{https://arxiv.org/abs/2605.00827}

\bibitem{CKA7}
T.~Kapp\'{e}, Concurrent kleene algebra: completeness and decidability
  (doctoral dissertation), Ph.D. thesis, UCL (University College London)
  (2020).

\bibitem{AIAYN}
A.~Vaswani, N.~Shazeer, N.~Parmar, J.~Uszkoreit, L.~Jones, A.~N. Gomez,
  L.~Kaiser, I.~Polosukhin, \href{https://arxiv.org/abs/1706.03762}{Attention
  is all you need} (2023).
\newblock \href {http://arxiv.org/abs/1706.03762} {\path{arXiv:1706.03762}}.
\newline\urlprefix\url{https://arxiv.org/abs/1706.03762}

\bibitem{TTCB}
A.~Yang, C.~Watson, A.~Xue, S.~Bhattamishra, J.~Llarena, W.~Merrill, E.~D.~S.
  Ferreira, A.~S., D.~Chiang, \href{https://arxiv.org/abs/2510.00368}{The
  transformer cookbook} (2025).
\newblock \href {http://arxiv.org/abs/2510.00368} {\path{arXiv:2510.00368}}.
\newline\urlprefix\url{https://arxiv.org/abs/2510.00368}

\bibitem{AT}
Y.~Wang, Z.~Wang, H.~Kang, \href{https://arxiv.org/abs/2610.36795}{An algebra
  of transformers} (2026).
\newblock \href {http://arxiv.org/abs/2610.36795} {\path{arXiv:2610.36795}}.
\newline\urlprefix\url{https://arxiv.org/abs/2610.36795}

\bibitem{APTC}
Y.~Wang, Algebraic theory for true concurrency, Elsevier AP, 2023.

\bibitem{APTC2}
Y.~Wang, Handbook of truly concurrent process algebra, Elsevier MK, 2023.

\end{thebibliography}

\end{document}